\documentclass[preprint,3p,times, 12pt,authoryear]{elsarticle}
\usepackage{lineno,hyperref}
\usepackage{graphicx} 
\date{November 22, 2022}
\usepackage[font=small,labelfont=bf]{caption}
\usepackage{subcaption}
\usepackage{algorithm}
\usepackage{algpseudocode} 
\usepackage{amsmath}
\usepackage{caption}
\usepackage[table, dvipsnames]{xcolor}
\usepackage{soul}
\usepackage[nameinlink,noabbrev]{cleveref} 
\usepackage{bigdelim} 
\usepackage{dirtytalk} 
\usepackage{booktabs} 
\usepackage{arydshln} 
\usepackage{adjustbox} 
\usepackage{tikz}

\algrenewcommand\algorithmiccomment[2][\footnotesize]{{#1\hfill\(\triangleright\) #2}} 
\newcounter{parentalgorithm}

\makeatother
\begin{document}

\begin{frontmatter}
\title{Feature construction using explanations of individual predictions}

\author{Boštjan Vouk}\corref{mycorrespondingauthor}
\ead{bostjan.vouk@scng.si}
\cortext[mycorrespondingauthor]{Corresponding author}

\author{Matej Guid} \ead{matej.guid@fri.uni-lj.si}
\author{Marko Robnik-Šikonja} \ead{marko.robnik@fri.uni-lj.si}
\address{University of Ljubljana, Faculty of Computer and Information Science, \\ Večna pot 113, 1000 Ljubljana, Slovenia}

\journal{Engineering Applications of Artificial Intelligence}

\begin{abstract}
Feature construction can contribute to comprehensibility and performance of machine learning models. Unfortunately, it usually requires exhaustive search in the attribute space or time-consuming human involvement to generate meaningful features. We propose a novel heuristic approach for reducing the search space based on aggregation of instance-based explanations of predictive models. The proposed Explainable Feature Construction (EFC) methodology identifies groups of co-occurring attributes exposed by popular explanation methods, such as IME and SHAP. We empirically show that reducing the search to these groups significantly reduces the time of feature construction using logical, relational, Cartesian, numerical, and threshold (\textit{num-of-N} and \textit{X-of-N}) constructive operators. An analysis on 10 transparent synthetic datasets shows that EFC effectively identifies informative groups of attributes and constructs relevant features. Using 30 real-world classification datasets, we show significant improvements in classification accuracy for several classifiers and  demonstrate the feasibility of the proposed feature construction even for large datasets. Finally, EFC generated interpretable features on a real-world problem from the financial industry, which were confirmed by a domain expert.
\end{abstract}


\end{frontmatter}

\section{Introduction}
Explainable artificial intelligence (XAI) is an important area of machine learning (ML) \citep{tjoa2019survey} and is one of crucial elements for deploying artificial intelligence (AI) models in practice  due to its ability to assure trust of users \citep{arrieta2020XAI}. The main goals of XAI are trustworthiness, causality, transferability, informativeness, confidence, fairness, accessibility, interactivity, and privacy awareness \citep{arrieta2020XAI}. Many of these goals can be achieved by transparent prediction models. The need for powerful explainable models is exacerbated by the lack of interpretability of currently dominant complex prediction models, such as deep neural networks and boosting.

The motivation for this work stems from the observation that practically no current learning system incorporates constructive induction (CI). This might be the consequence of computational complexity of the existing CI approaches. We address these issues by proposing a novel approach to constructive induction with a reduced search space and a lower computational time, by implementing more constructive operators than existing methods, and by making the system freely accessible.

The inherently interpretable predictive models, such as decision trees, na\"ive Bayes and linear models, lack predictive power due to inadequate representation of raw data. While this can be amended by construction of composite features, a generation of new features is computationally problematic due to a huge search space. To overcome this issue, we propose a novel approach to constructive induction that uses popular explanation methods to identify groups of potentially interacting attributes that could provide useful constructs. Limiting the search for new features to the identified groups results in much smaller sets of high-quality candidates.

The research in perturbation-based explanation methods has produced several useful methods, such as IME \citep{Strumbelj-2010}, LIME \citep{Ribeiro-2016-LIME}, and SHAP \citep{Lundberg-2017-SHAP-modelInterpreting}. These methods provide explanation for individual predictions of black-box models as contributions of attributes to the prediction of a given class label. We hypothesise that groups of attributes frequently appearing together in these explanations identify potentially useful attribute interactions -- the possibility that has not been exploited in the existing research.

Feature engineering (FE) transforms an input problem space by constructing new features to be used in ML models. A better representation of the underlying problem often leads to improvements in the learning performance \citep{ozdemir2018feature}. FE is usually time consuming and may lead to overfitting. The most typical FE methods are feature transformation, feature extraction, feature construction, feature selection, and feature analysis. This research focuses on feature construction. We aim for human-interpretable features that could be used in intrinsically human-comprehensible models, such as decision trees. Using better features, such models could offer better performance and better interpretation.

In the proposed EFC (Explainable Feature Construction) system, we construct various types of features: operator-based features (using logical, relational, Cartesian, and numerical operators), features from rule learning \citep{Huhn-2009-Furia}, and features based on a threshold for the presence of several features (frequently called \textit{num-of-N} and \textit{X-of-N}) \citep{zheng2000-XofN}. We evaluate the constructed features on a set of synthetic and real-world datasets. To show practical utility of the EFC approach, we demonstrate its use on a real-world problem from the financial domain with human evaluation.

The main contributions of the paper are as follows.
\begin{enumerate}
    \item \textit{A novel heuristic for reducing the search space of constructive induction. The proposed EFC constructive induction system searches in a small space of co-occurring explanations generated by the perturbation-based explanation methods. This space consists of groups of potentially interacting attributes.} By applying this novel idea, we reduce the exponential search space of constructive induction to the linear space of co-occuring features.
    \item \textit{A range of constructs.} Using an efficient search in the reduced space, the EFC system can generate more types of constructs (logical, relational, Cartesian, numerical, and threshold features) as other existing systems. EFC is freely available\footnote{\url{https://github.com/bostjanv76/featConstr}}.
    \item \textit{The use case from the financial industry shows that generated features are similar or identical to human-created features in this domain.} This hints at applicability of the EFC method as a standalone approach or to support humans in the knowledge discovery process.
    \item \textit{An extensive empirical evaluation using synthetic and real-world UCI datasets shows the effectiveness of the proposed approach.} The gains are visible in terms of computational speed, prediction performance, and comprehensibility.
\end{enumerate}

\sloppy{The remainder of the paper is structured into five sections. In \Cref{sec:relatedWork}, we present a systematic literature review. \Cref{sec:methodology} describes the proposed methodology, where we first provide an outline, which is followed by the formal definition of the problem and description of the components. \Cref{lab:evaluation} presents the datasets used in the study (the synthetic, the UCI, and the real-world dataset from the financial domain) as well as the evaluation scenarios. \Cref{lab:results} reports the results. We present conclusions, strengths and limitations of the approach, as well as the ideas for further work in \Cref{sec:conclusions}.}

\section{Background and related work}
\label{sec:relatedWork}
We review the related work in two subsections. In Section \ref{sec:FC}, we overview the research on feature construction, roughly divided into standalone feature construction, embedded feature construction, and work on feature interactions. Section \ref{sec:relatedExplanations} outlines applications of explanation methods related to our work. Each subsection is accompanied by a table, summarising the listed references.

\subsection{Feature construction}
\label{sec:FC}
Feature construction (FC) is a process in which new features are constructed from raw data or previously constructed features to improve model robustness, interpretability, and/or generalization \citep{he2019automl}. In the process of FC, we use constructive operators and existing attributes/features. As a result, we get features that may better describe the target concept \citep{matheus1989constructive}.

Machine learning research has produced several feature construction approaches, such as FICUS \citep{markovitch2002feature}, FCTree \citep{fan2010generalized}, FEADIS \citep{dor2012strengthening}, and, more recently, automated feature engineering methods, e.g., ExploreKit \citep{katz2016explorekit}, Data Science Machine \citep{kanter2015deep}, One Button Machine \citep{lam2017one}, and Auto-Sklearn \citep{feurer2015auto-sklearn}. These techniques construct new features that better describe target concepts. Consequently, new features can improve classifiers' performance and comprehensibility. The above-mentioned methods are summarised in Table \ref{tab:reviewFC} and described below,  divided into three categories: standalone (operator-based) feature construction, methods embedded into automated ML, and methods focused on feature interactions.

\subsubsection{Standalone feature construction}
Feature construction consists of the following four steps: 1) generating a set of candidate features, 2) ranking candidate features, 3) evaluating and selecting high-ranked features, and 4) adding promising features to the dataset \citep{katz2016explorekit}. Process of FC is mostly guided by human experts and is not as efficient as automatic feature construction (AFC) \citep{fang2019automatic}, which can be used in multiple domains and applications, while (manual) FC is problem- or domain-specific \citep{dong2018feature}.

\sloppy{\cite{markovitch2002feature} presented the FICUS algorithm for feature construction. The input of the framework is a set of original attributes and a set of construction functions. This method extends past methods (CITRE \citep{matheus1989constructive}, FRINGE \citep{bagallo1990boolean}, IB3-CI \citep{aha1991incremental}, LFC \citep{ragavan1993complex}) that use only a minimal set of logical operators ($\neg, \land$) to construct Boolean features. FICUS uses mathematical operators ($+,-,/,*,=, avg, max, min, absDi\!f\!f$), standard logical operators ($\land, \vee$), special logical operators (occurrence counting), and interval operators (range test). In FICUS, the feature space is reduced by a heuristic measure based on information gain in a decision tree, and other surrogate measures of performance. There are two potential drawbacks of this method. First, feature subset selection does not take into account feature interactions, and second, it is difficult to supply a set of operators suitable for a given problem beforehand.}

FCTree \citep{fan2010generalized} uses decision trees to partition the data with original attributes or constructed features as splitting criteria. FCTree divides the training data recursively based on the information gain criterion, and constructs new features randomly. 

The FEADIS \citep{dor2012strengthening} algorithm relies on a combination of random feature generation and feature selection. It greedily adds constructed features to the training set, and therefore requires many computationally expensive evaluations. 

\cite{duan2018automated} leverage Fourier analysis of Boolean functions. From the input attributes, they first generate features in the disjunctive normal form (DNF) using the RIPPER decision rule learner \citep{cohen1995fast} and then extract the resulting Boolean features. Afterwards, they apply Fourier analysis to extracted features in order to generate new parity features, which are ranked in descending order according to the absolute values of Fourier coefficients. The authors suggest selecting up to 30 top-ranked features for each domain.

\sloppy{\cite{mozina-phd-thesis} proposed an argument-based machine learning algorithm ABCN2 for feature construction (requiring human experts). Argumentation is used to construct new features, e.g., from the argument \textit{Credit = no} because  \textit{money\_bank} $<$ \textit{monthly\_payment} $\cdot$ \textit{num-of\_payments} ... a new feature \textit{enough\_money} can be created by an expert, covering the explanatory part. ABCN2 does not automatically generate suggestions for new features. ABCN2 was used in an intelligent tutoring system \citep{woolf2010building} to teach students arguing strategies  \citep{guid2019learning}. The intelligent system by \cite{xing2022-selfmatch} uses feature extraction in semi-supervised learning and self-distillation; a similar feature extractor was used in the federated distillation learning system \citep{xing2022-EFDLS}. The last two methods are specific to time-series classification and do not use explicit constructive operators.}

\cite{zupan1998feature} proposed a data-driven constructive induction method called HINT (Hierarchy INduction Tool). The method hierarchically constructs new features. In each step, the method tests all possible combinations of variables (attributes) up to the prespecified length and selects the one from which the function that minimises the search space can be constructed. It replaces these variables with a new, derived variable and iteratively continues. The method can discover a hierarchy of concepts by iteratively transforming attributes into new concepts. The method does not need  constructive operators, therefore it can be termed as data-driven operator-free constructive inducer. However, the lack of operators reduces the comprehensibility of the induced representation. Unfortunately, this complex method is no longer available in the authors' Orange data mining suite \citep{JMLR:demsar13a}.

\subsubsection{Embedded feature construction}
To reduce the exploration time and to improve the performance, the process of feature construction can be automated \citep{katz2016explorekit, kanter2015deep, lam2017one, feurer2015auto-sklearn}. The frameworks reviewed below try to automatically construct useful and comprehensible features from the original set of attributes.

There are two approaches to constructing features automatically; the expand-reduce approach and the wrapper approach. ExploreKit \citep{katz2016explorekit}, Deep Feature Synthesis \citep{kanter2015deep}, One button machine \citep{lam2017one}, and Auto-Sklearn \citep{feurer2015auto-sklearn} use the expand-reduce method, which first generates a vast number of features, then evaluates them, and after that reduces their number. The wrapper approach uses a prediction method wrapped within a greedy heuristic search (e.g., forward selection, backward elimination, hill-climbing, best-first search) to select the optimal feature subset.

\par ExploreKit generates new feature candidates by combining the original attributes and retaining the most promising ones. From a single attribute, it generates unary features with normalisation and discrimination operators, e.g., discretisation of attributes into segments. New features can be combinations of two features, e.g., arithmetic operations $(+,-,*,/)$, or combinations of several features, e.g., operators  GroupByThenMax, GroupByThenMin, GroupByThenAvg, GroupByThenStdev, and GroupByThenCount. The aim of the method is to maximise the predictive performance. First, the candidate features are ranked by a classifier, and then the method iteratively selects the features with the highest rank. The features that reduce the classification error for more than a user-defined threshold are added to the feature set. The ranking classifiers require additional training data to avoid overfitting and may be computationally expensive to train.

\sloppy{Data Science Machine (DSM) automatically synthesises features from relational normalised datasets. Features are built by stacking predefined basic aggregation functions (e.g., max, min, avg) across relationships in the datasets. Each feature has a certain depth $d$ (e.g., $d=1$: tableX.attributeA; $d=2$: sum(tableX.attributeA); $d=3$: avg(tableY.sum(tableX.attributeA))). The system performs feature selection and model optimisation on the enlarged dataset. The weakness of the framework is that it does not support learning from unstructured data such as texts, i.e., it cannot handle complex data structures and features. Besides, the use of exhaustive feature enumeration and selection is time and memory consuming and may lead to overfitting due to generation of all possible features.}

\par An extension of DSM is One Button Machine (OneBM), which supports structured and unstructured data. It expands the types of feature quantizations from DSM and uses embeddings. OneBM uses propositionalization \citep{Lachiche2010}, which is a two-step process. In the first step, relational data are transformed into a single table, and in the second step, a propositional learning algorithm is applied to the transformed data. Another extension of DSM was proposed by \cite{lam2018RNN}, who used Recurrent Neural Networks as the aggregation functions.

In Auto-Sklearn feature construction is embedded in meta learning phase where Bayesian optimisation is used to determine both the optimal combination of relevant features and model hyper-parameters to minimise the classification error. All datasets are  represented by 38 meta-features which are extracted from preprocessed data. Meta-features are based on simple dataset characteristics (e.g., number of classes, number of attributes, number of instances), information-theoretic properties (e.g., class entropy), and statistical characteristics (e.g., data skewness). 

Our approach uses some of the above-listed  constructive operators, e.g., decision rules, logical, relational, numerical, and threshold operators.  Decision rules also cover basic aggregation operators and interval operators. We do not use all of the above operators, as some of them (e.g., parity operators) are specific for certain types of datasets and problems.

The feature construction process is usually not fully automated. First, an exhaustive search of the entire feature space for a given problem is practically infeasible as the number of possible feature subsets increases exponentially with respect to the number of original attributes. Further, the potential feature space can be even larger if new features can be constructed iteratively or with many operators \citep{zhao2009effects}. One way to reduce the number of potential subsets in feature construction is to consider interactions between attributes.

\begin{table*}[!htbp]
  \centering
  \caption{A summary of covered feature construction (FC) methods and their strengths and weaknesses. We use the following  category labels: OB (operator-based FC), eML (FC embedded into auto ML), TS (FC for time-series classification), and INT (interactions). The abbreviations used in the table are as follows: DT (decision trees), FSS (forward sequential selection), BSS (backward sequential selection), and IG (information gain).}
    \label{tab:reviewFC}
    \scriptsize
    \resizebox{\linewidth}{!}{
    \begin{tabular}{@{}p{.2\textwidth}p{0.05\textwidth}p{.40\textwidth}p{.35\textwidth}@{}}
    \toprule
    \textbf{FC Method} & \multicolumn{1}{c}{\textbf{Category}} & \multicolumn{1}{c}{\textbf{Weakness}} & \multicolumn{1}{c}{\textbf{Strength}} \\
    \midrule
    FICUS \citep{markovitch2002feature} & 
    \multicolumn{1}{c}{OB}
        &Must be supplied a set of suitable operators for a given problem.
        Feature evaluation does not take into account feature interactions.
        &Generality and flexibility. Construction of comprehensible features.
    \\
    CITRE \citep{matheus1989constructive} & 
    \multicolumn{1}{c}{OB}
        &Supports only a minimal set of logical operators. 
        Only applicable for specific learning algorithms such as DT.    
        Limited to binary classification problems.
        Uses rigid domain-knowledge filtering.
        &Comprehensibility.
    \\
    FRINGE \citep{bagallo1990boolean} & 
    \multicolumn{1}{c}{OB}
        &Supports only a minimal set of logical operators.      
        Only applicable for specific learning algorithms such as DT.
        Limited to binary classification problems.
        Tested only on synthetic datasets.
        & Comprehensibility (reducing a tree size).
        Increased predictive accuracy.
    \\
    IB3-CI \citep{aha1991incremental}  & 
    \multicolumn{1}{c}{OB}
        & Supports only a minimal set of logical operators.
        Sensitive to noise.
        & Well adapted to board game problems. 
        Uses domain knowledge to reach high generalization rate.
    \\
    LFC \citep{ragavan1993complex} & 
    \multicolumn{1}{c}{OB}
        &Supports only a minimal set of logical operators.
        Generates a large set of attributes.
        & Construction is heuristically guided by IG measure.
    \\
    FCTree \citep{fan2010generalized} & 
    \multicolumn{1}{c}{OB}
        & Not capable of composing transformations, i.e. it searches in a small space.
        &Uses information-theoretic criteria to guide the search.
        Creates new features during the tree construction.
        The number of selected features is bounded.
    \\
    FEADIS \citep{dor2012strengthening} & 
    \multicolumn{1}{c}{OB}
        &Requires many computationally expensive evaluations.
        Greedily adds constructed features.
        Dimensionality before feature selection can become huge.
        & Flexibility, extendability, and simplicity.
        Requires no tuning parameters.
    \\
    Duan \citep{duan2018automated} & 
    \multicolumn{1}{c}{OB}
        &Uses only parity features.
        & Naive heuristic for selecting top-ranked features.
    \\
    ABCN2 \citep{mozina-phd-thesis} & 
    \multicolumn{1}{c}{OB}
        &Suggestions for new features are not generated automatically and require a domain expert.
        &
        Experts' arguments reduce the search space.
        System automatically detects important examples and counter-examples.
    \\
    HINT \citep{zupan1998feature} & 
    \multicolumn{1}{c}{OB}
        & A lack of constructive operators and comprehensibility.
        & Identifies and eliminates redundant subsets of attributes.
    \\
    \hdashline
    ExploreKit \citep{katz2016explorekit} & 
    \multicolumn{1}{c}{eML}
        & Time and memory consuming. Prone to overfitting. Unable to infer domain knowledge. Issues of trust. Iterative algorithm is difficult to parallelise.
        &Outperforms the IG-based methods both in ranking-only and ranking-and-evaluation settings. 
        Robustness.
        More scalable than grid search.
    \\
    DSM \citep{kanter2015deep} &
    \multicolumn{1}{c}{eML}
        &Time and memory consuming. Prone to overfitting. Unable to infer domain knowledge. Issues of trust. Does not support learning from unstructured data.
        & Exploits structural relationships inherent in database design.
    \\
    OneBM \citep{lam2017one} & 
    \multicolumn{1}{c}{eML}
        &Time and memory consuming. Prone to overfitting. Unable to infer domain knowledge. Issues of trust.
        Uses a relatively restrictive bias.
        Suffers from performance and scalability bottleneck.
        &Able to extract distinguishing features from relational databases.
        Scalability.
    \\
    Auto-Sklearn \citep{feurer2015auto-sklearn} & 
    \multicolumn{1}{c}{eML}
        &Time and memory consuming. Prone to overfitting. Inability to infer domain knowledge. Issues of trust. 
        Acting as a black box.
        Weak models can remain in the ensemble.
        & Optimizes hyperparameters.
        Meta-learning and ensemble building.
    \\
    \hdashline
    ResNet–LSTMaN \citep{xing2022-selfmatch} & 
    \multicolumn{1}{c}{TS}
        &Works only on time series.
        &Robustness.
    \\
    EFDLS \citep{xing2022-EFDLS} & 
    \multicolumn{1}{c}{TS}
        & Works only on time series.
        & Multi-task time-series classification.
    \\
        \midrule
    SNB \citep{kononenko1991semiNB} & 
    \multicolumn{1}{c}{INT}
        &Requires multiple iterations.
        & Joins dependent attributes based on the Chebyshev theorem.
    \\
    Pazzani \citep{pazzani1996searching} & 
    \multicolumn{1}{c}{INT}
        &Requires multiple iterations.
        & Finds dependencies between the attributes with FSS, BSS, elimination, and joining.
    \\
    Jakulin \citep{jakulin2005MLattrInteractionsPhD} &  
    \multicolumn{1}{c}{INT}
        & Uses greedy attribute selection. 
        Only tested on synthetic domains.
        &Heuristically detects attribute interactions.
    \\
    IME \citep{Strumbelj-2009} & 
    \multicolumn{1}{c}{INT}
        & Slow for large datasets.
        Interactions are not explicitly identified. 
        Dependent on a prediction model.
        & Graphical visualisation. 
        Linear time complexity.
    \\
    ASTRID \citep{henelius2017interpreting} & 
    \multicolumn{1}{c}{INT}
        & Computationally demanding. 
        Only suitable for classification problems.
        & Identifies interacting attributes. 
        Can be applied to any model.
    \\
    \bottomrule
     \end{tabular}%
 	}
\end{table*}%

\subsubsection{Feature interactions}
\label{sec:featureInteraction}
Feature interaction refers to a situation where certain features are not individually related to the target concept, unless they are in some way combined with other features. To avoid construction of meaningless features, methods search for possibly co-occurring attributes before the feature construction process.

\cite{jakulin2003quantifying} defined feature interaction as the amount of information that is common to all the random variables but not present in any of them. According to this definition, the interaction information can be negative, because the dependency among a set of variables can increase or decrease with the knowledge of a new variable. \cite{Strumbelj-2009} proposed interactions-based explanation method and defined feature-interaction contributions as the contribution of interactions between subsets of attributes to the model’s decision. \cite{henelius2017interpreting} proposed the ASTRID method, which interprets classifiers by detecting attribute interactions. The main idea of the method is to examine attribute interactions for better interpretation of the model's predictions. The method searches for such partitions of attributes into subsets that do not significantly decrease the classifier's accuracy (using certain subsets). This process may identify interacting attributes. A drawback of the method is that it is computationally demanding. 

To improve prediction performance of naive Bayesian algorithm, \cite{pazzani1996searching} and \cite{kononenko1991semiNB} used attribute interaction information to guide feature construction. Both authors considered joining more than two attributes, but this required multiple iterations.
 
\cite{jakulin2005MLattrInteractionsPhD} used interaction information to construct new features. He proposed the following exhaustive methodology: 1) estimate interaction information from all pairs of attributes, 2) select attribute pairs with the highest interaction information, 3) use each previously selected pair and construct new features.

Our approach differs from the above mentioned methods in the way we identify possible interactions between attributes. We count attribute co-occurrence in explanations of model predictions. We assume that used complex black-box models (e.g., boosting) have captured some interactions between attributes and that this information is reflected in explanations. The groups of co-occurring attributes in explanations therefore potentially identify attribute interactions. The reduction of search to these groups significantly reduces the search space of constructive induction. 

\subsection{Applications of explanation methods}
\label{sec:relatedExplanations}
While simple predictive models are self-explanatory, complex models (including decision trees or linear models) are beyond human comprehension. However, simple models are often less accurate than complex ones. In order to trust predictions of complex models, it is important to verify them with explanation methods. These methods often use interpretable (local) approximations of the original complex model. If a user can unravel inner workings of a black-box model, its decisions are more trustworthy \citep{Bohanec2017SalesPredictions,molnar2019-Interpretable-ML}.

Explanation methods can be classified into two groups. The first group includes model-agnostic methods like SRV (Shapley Regression Values) \citep{lipovetsky2001analysis}, EXPLAIN \citep{robnik2008explain}, IME \citep{Strumbelj-2009}, LIME \citep{Ribeiro-2016-LIME}, QII \citep{datta2016algorithmic}, and SHAP (SHapley Additive exPlanations) \citep{Lundberg-2017-SHAP-modelInterpreting}. The second group consists of model-specific methods, such as interpreting random forests \citep{Saabas2014}, Layer-wise Relevance Propagation (LRP) \citep{bach2015RelevancePropagation}, DeepLIFT \citep{DeepLIFT2017}, and GAM \citep{ibrahim2019global}. Many model-agnostic methods are based on game theory (SRV, QII, IME, LIME, and SHAP). \cite{bodria2021benchmarking} present a recent survey of explanation methods. 

Although the proposed EFC method is independent of a particular explanation method we use IME \citep{Strumbelj-2014} and SHAP \citep{Lundberg-2017-SHAP-modelInterpreting} in this study to identify groups of possible interacting attributes. This is a novel use of explanation methods. Below we outline other applications of explanation methods and summarise them in Table \ref{tab:reviewEXPL}.

\begin{table*}[!htb]
  \centering
    \caption{Applications of explanation methods. The type label denotes either a model-specific (specific) or a model-agnostic (agnostic) approach.}
    \label{tab:reviewEXPL}
    \small
    \begin{tabular}{@{}p{.5\textwidth} @{} p{0.1\textwidth} @{} p{.4\textwidth}@{}}
    \toprule
    \textbf{Explanation Method} & \textbf{Type} & \textbf{Applications} \\
    \midrule
    SRV \citep{lipovetsky2001analysis} & agnostic & telecommunications industry \\ 
    EXPLAIN \citep{robnik2008explain} & agnostic & medicine, economics, electrical industry, airline industry, text categorisation problems \\
    Lemaire's method \citep{lemaire2008contact} & agnostic & telecommunications industry\\
    IME \citep{Strumbelj-2009} & agnostic & medicine, economics, electrical industry, airline industry, text categorisation problems \\
    LIME \citep{Ribeiro-2016-LIME} & agnostic & text, image and tabular classification problems\\
    QII \citep{datta2016algorithmic} & agnostic & socioeconomics\\
    SHAP \citep{Lundberg-2017-SHAP-modelInterpreting} & agnostic & medicine, economics,  sports, aviation industry, software industry \\
    \hdashline
    Saabas's method \citep{Saabas2014} & specific & real estate industry \\
    LRP \citep{bach2015RelevancePropagation} & specific & image classification problems\\
    DeepLIFT \citep{DeepLIFT2017} & specific & biological problems\\
    Arras's method \citep{Arras_2017} & specific & text categorisation problems\\
    GAM \citep{ibrahim2019global} & specific & economics \\
    RUDDER \citep{arjonamedina2018rudder} & specific & game industry\\
    \bottomrule
    \end{tabular}
\end{table*}

We first focus on perturbation-based methods. The SHAP method has been used in the system called Prescience that deals with anesthesia safety \citep{Lundberg-2018-SHAP-hypoxaemia} and to calculate mortality risk factors in the general US population. \cite{lundberg2020local} report several applications of the SHAP method. It has been used to optimise performance in sports (NBA team Portland Trail Blazers); Microsoft uses the method in its Cloud Pipeline; Cleveland clinic uses SHAP to identify different subtypes of cancers; it has been used for optimising manufacturing process for a jet engine at Rolls Royce where they build complicated models and try to figure out what is breaking and why. Bank of England uses SHAP values to improve collaboration between humans and machines for decision-making and for financial risk prediction. Recently, SHAP has been used in the Driverless AI commercial package \citep{H2ODriverlessAI}. The package also contains other explanatory methods, such as K-LIME.

The IME method has been initially applied to real-world medical problems \citep{vstrumbelj2010explanation}. We can find applications of IME in economics, more precisely, in the context of B2B sales forecasting, where explanations are used to validate and test user hypotheses. Besides the IME method, \cite{Bohanec2017SalesPredictions}  also used the EXPLAIN  method. \cite{lemaire2008contact} applied explanations in the telecommunications industry. Authors used instance level explanations for immediate decisions in a customer relationship management system.

Applicability of explanation methods is not limited to interpretability of black-box models, but can be used for other tasks (e.g., feature selection and stream mining). \cite{Strumbelj-2009} proposed to use the IME explanation method as a filter method for feature selection. Explanation method proposed by \cite{Lundberg-2017-SHAP-modelInterpreting} can also be used to calculate feature importance. The EXPLAIN and IME explanation methods were used to explain a concept drift in an incremental model for predicting flight delays and for predicting changes of electricity prices \citep{demvsar2018detecting}. \cite{arjonamedina2018rudder} used explanation in reinforcement learning, while \cite{Arras_2017} used explanations of word features to build a representation of a document.

To the best of our knowledge, no previous work has used explanation methodology to identify groups of co-occurring attributes that could constitute feature interactions. Additionally, we are not aware of any application in the financial industry.

\section{Methodology}
\label{sec:methodology}
We propose a methodology for efficient feature construction based on explanation of individual predictions. Our aim is to construct human-comprehensible features that will improve interpretability of ML models and their prediction performance.

\begin{figure}[h!tb]
    \centering
    \caption{Illustration of the proposed EFC methodology.}
    \includegraphics[keepaspectratio,width=\linewidth,height=\dimexpr\textheight-3\baselineskip]{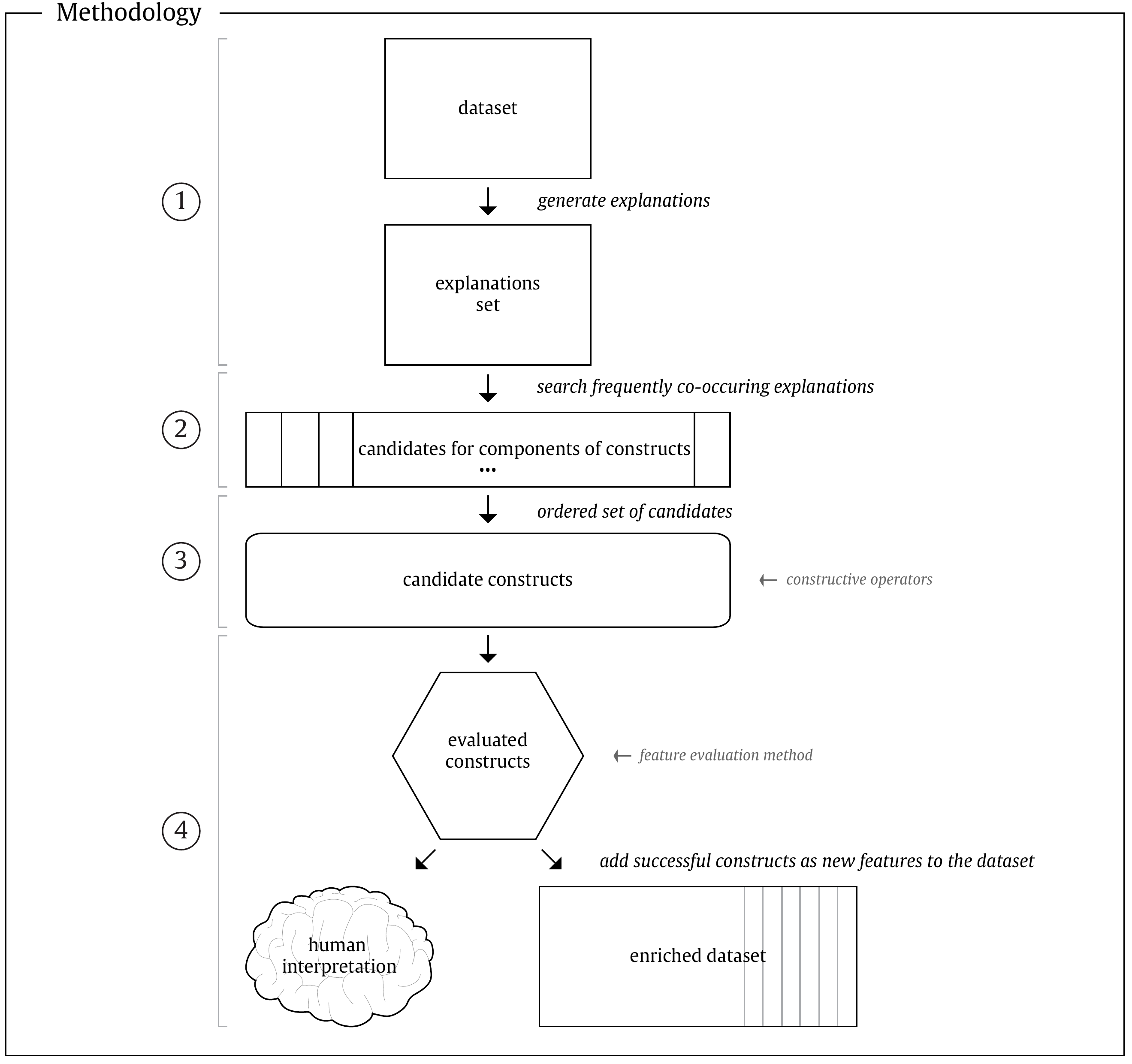}
  \label{fig:methodology}
\end{figure}

The proposed methodology consist of the following four steps illustrated in \Cref{fig:methodology}:
\begin{enumerate}
    \item Explanation of model predictions for individual instances, which is described in \Cref{sec:firstStepMethodology}.
    \item Identification of groups of attributes that commonly appear together in explanations, which is presented in \Cref{sec:secondStepMethodology}.
    \item Efficient creation of constructs from the identified groups, which is described in \Cref{sec:thirdStepMethodology}.
    \item Evaluation of constructs and selection of the best as new features, which is presented in \Cref{sec:fourthStepMethodology}.
\end{enumerate}
Below, we first present the notation. After we outline the methodology in Sections \ref{sec:firstStepMethodology} - \ref{sec:fourthStepMethodology}, we analyse its computational complexity in \Cref{sec:complexity} and give details of the implementation in \ref{sec:implementation}. A worked example of the methodology is demonstrated in \Cref{sec:example}.
The evaluation scenario is described in \Cref{lab:evaluation} and the results are presented in \Cref{lab:results}.

We use the following notation. We assume two sets of instances, training and explanation. 
Let $I^t=(X^{n \times m},Y^{1 \times n})$ be the training data, where $X^{n \times m}$ is an instance space of $n$ training instances and $m$ attributes. $Y^{1 \times n}$ is a label space (vector of class label). We denote the prediction model as $M$ and the class to explain as $c$ where $c\in \{1\dots C\}$. In this work, if not stated otherwise, we explain the minority class, which is frequently of more interest, e.g., in financial fraud \citep{albashrawi2016detecting}. Let $I^e=(X^{e \times m},Y^{1 \times e})$ be the data we use for explanations (typically, this would be the training set or a part of the training set consistent with $e$ instances). Let $E^{e \times m}$ be the matrix of explanations obtained with the given explanation method from prediction model $M$ using the explanation  instances $I^e$. $E_{i}$ is an explanation of individual instance $i$, and $E_{ij}$ is an estimated contribution of attribute $j$ to the prediction of instance $i$ into class $c$.

\subsection{Generation of instance explanations}
\label{sec:firstStepMethodology}
In Step 1 of EFC methodology (see \Cref{fig:methodology}), we generate explanations $E$ based on the prediction model $M$ and  explanation instances $I^e$. While we can use any prediction model, we recommend using robust well-performing models such as random forests or XGBoost. Namely, if the prediction model is wrong, the produced explanations and constructs that exploit its information will suffer and produce suboptimal results. In the next step, these explanations are used to find the groups of attributes that often co-occur in explanations, i.e., the candidate interactions. To produce explanations, we use any of the before-mentioned instance explanation techniques. In this work, we work with two state-of-the-art-methods, SHAP \citep{Lundberg-2017-SHAP-modelInterpreting}, and IME \citep{Strumbelj-2009}. The explanation methods return a vector of explanations $E_i$ for each of the instances $i$. The dimension of explanation vectors is equal to the number of attributes $m$. Each dimension of the explanation vector represents an estimated contribution of one attribute to the prediction of the instance into the selected class $c$. We illustrate these explanations later, in \Cref{fig:ExplanationConcept1}. The value of $E_{ij}$ represents the estimated contribution of attribute $j$ to the prediction of instance $i$.

\subsection{Identification of co-occurring attributes}
\label{sec:secondStepMethodology}
In Step 2 (see \Cref{fig:methodology}), we take the matrix of instance explanations $E$, identify commonly co-occurring groups of attributes, and treat them as possible interactions which could be expressed with constructs. The absolute value of $E_{ij}$ represents the amplitude of the impact of attribute $j$ to the classification of instance $i$ with model $M$. Each instance $i$ typically has only a few large explanation values (either positive or negative), others being close to zero. If the prediction model detects any interactions between the attributes, the components of the interactions are expressed as large explanation values in those instances where the interactions affect the predictions. Large co-occurring explanation values might also be the result of strongly correlated independent (non-interacting) attributes, so frequent co-occurrence of attributes serves only as a heuristic guideline which attributes to probe together in constructive induction. This intuition is the key to our approach. 

\begin{figure}[h!bt]
    \centering
    \caption{Search of frequently co-occurring explanations.}
    \includegraphics[keepaspectratio,width=\linewidth,height=\dimexpr\textheight-3\baselineskip]{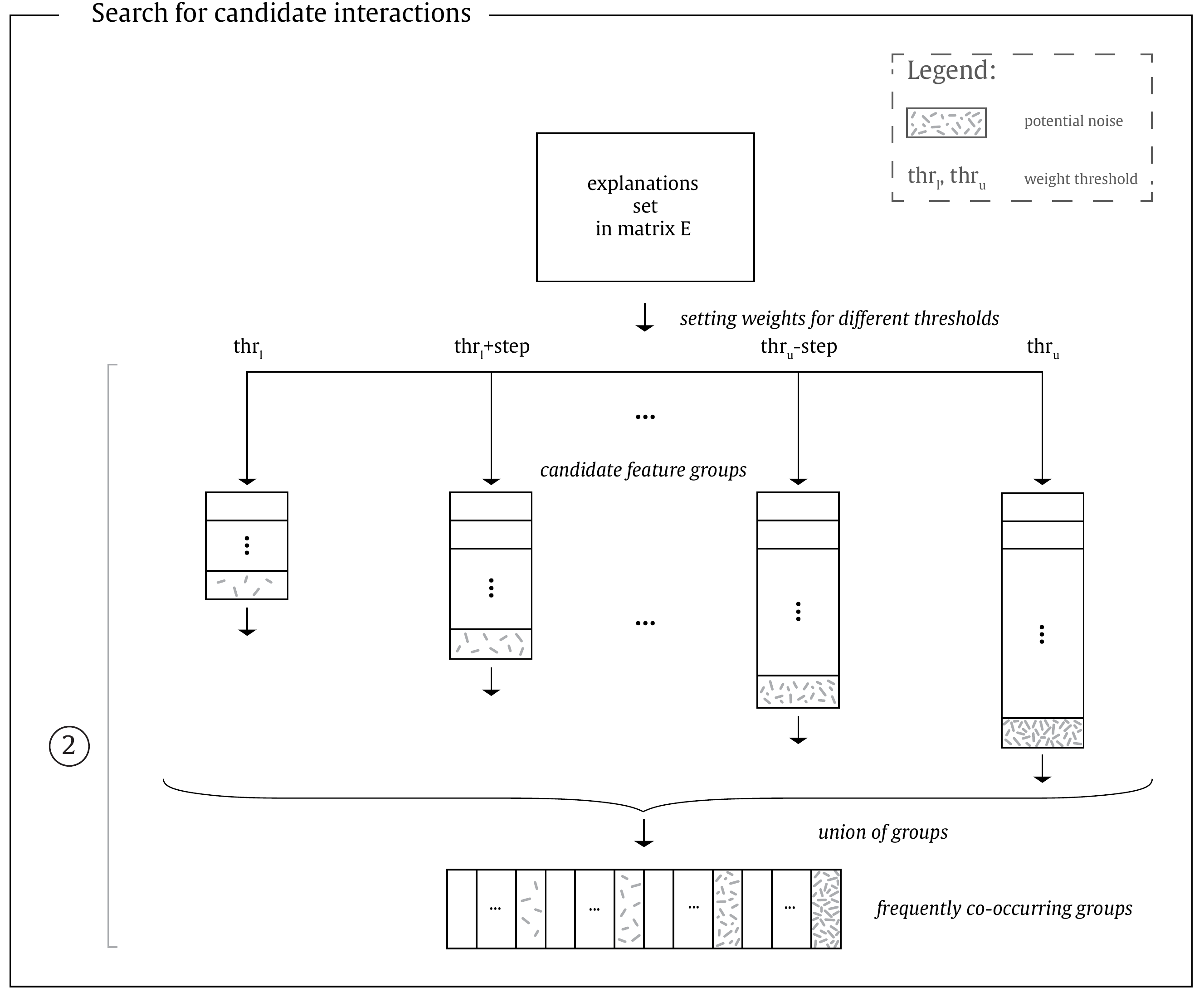}
    \label{fig:Second step}
\end{figure}

In the rows (representing instances) of matrix $E$, we search for commonly co-occurring groups of large values (indicating possibly interacting attributes) and take them as candidates for components of new features. In this way, instead of using the huge space of the original attributes, the search for useful constructs is conducted in the much smaller and more informative space of co-occurring explanations. 
The search procedure is summarised in \Cref{fig:Second step}. The absolute values of explanations $E_i$ (for each instance) are normalised to 1 and  sorted in descending order (i.e., by decreasing impact on the prediction). They are summed in that order and when they exceed a threshold (parameter $thr$), we mark those that are part of the sum as important and the rest as unimportant. For each threshold (from the lower bound $thr_{l}$ to the  upper bound $thr_{u}$ by the step $step$), we generate candidate feature subsets and concatenate them. Due to potential noise in the prediction models and explanation procedures, we might also collect a few false candidates.

The output of Step 2 is a collection of frequently co-occurring groups of attributes stored in such a way that duplicates are avoided.

\subsection{Efficient feature construction}
\label{sec:thirdStepMethodology} 
In Step 3 of the EFC methodology (illustrated in \Cref{fig:methodology}),  we construct several types of features, using individual groups of frequently co-occurring attributes to reduce search space, thereby significantly reducing the computational complexity of constructive induction in practical and theoretical terms (see \Cref{sec:complexity}). The actual number of constructed features depends on the prediction model used in  explanations, the number and size of the co-occurring groups of attributes, the depth of the construction, and the number of operators. For logical features, generated from numerical attributes, the number of discretisation intervals must be also taken into account. We generate several types of features, e.g., operator-based features (e.g., logical, relational, Cartesian, and numerical operators), features from decision rule algorithms (we use the FURIA decision rule learner \citep{Huhn-2009-Furia}), and threshold features (\textit{X-of-N}, \textit{all-of-N}, \textit{M-of-N}, and \textit{num-of-N} \citep{zheng2000-XofN}). \textit{X-of-N} construct is true if $X$ of $N$ specified conditions are true, and \textit{all-of-N} is true if all the argument conditions are true. \textit{M-of-N} is true if at least $M$ out of $N$ specified conditions are true. Finally, \textit{num-of-N} counts the number of true conditions and returns their number.

In this way, we build an explicit and potentially more comprehensible representation in the space of original attributes. The output of this step is the set of constructed features, which is evaluated and reduced in the next step.

\subsection{Evaluation and selection of new features}
\label{sec:fourthStepMethodology} 
In Step 4 (see the \Cref{fig:methodology}), we evaluate the features constructed in the previous step and select the most relevant ones. We evaluate features in the original feature space using the MDL feature evaluation measure \citep{kononenko-1995-biases}, which has a number of favourable properties compared to other impurity-based measures (such as better-known gain ratio \citep{quinlan1986induction}), e.g., better behaviour in multi-class scenario and fairer treatment of multi-valued attributes. The reason to use impurity-based measure instead of ReliefF measure \citep{robnik2003experiments}, which can detect conditionally interacting features is that successful feature construction should generate features that are directly related to the class and detectable with impurity-based measures. 

Note that \Cref{fig:methodology} shows explanations and creation of new features based on binary-class predictions focusing on the positive class. For multi-class problems, it might be sensible to repeat the process for all class values and harvest more constructs. We leave this exploration for further work. 

\subsection{Computational complexity}
\label{sec:complexity}
The computational complexity of EFC methodology consists of four parts: generation of instance explanations, identification of groups of co-occurring attributes in explanations, feature construction, and feature evaluation.

The time to generate instance explanations consists of two parts: the construction of the prediction model and generation of explanations with the explanation method (i.e., with IME or SHAP). We skip the analysis of the prediction model complexity, which depends on the machine learning algorithm used, e.g., for support vector machines it is between $\mathcal{O}(n^2)$ and $\mathcal{O}(n^3)$ \citep{chapelle2007training}, for random forest $\mathcal{O}(m \cdot n \cdot \log{(n)} \cdot k)$ \citep{hassine2019important}, for XGBoost $\mathcal{O}(k \cdot d_{t} \cdot n' \cdot \log{(n)})$ \citep{chen2016xgboost}; here $k$ is number of trees, $n'$ number of non-missing entries in the training data, and $d_{t}$ the maximum depth of the tree.

In other analyses, we use the following notation. Let $n_{e}$ denotes the number of explaining instances, $n_{t}$  the number of instances in the training dataset, $g_{n}$ the number of groups of attributes for feature construction, $g_{m}$ the number of attributes in the groups, $d_{f}$ the depth of the construction, $op_{n}$ the number of operators, and $l_{max}$ the maximum number of leaves in any tree. 

The computational complexity of explanation methods (e.g., SHAP and IME) that generate instance explanations is linear in the number of explained instances. The computational complexity of computing SHAP values for $n_{e}$ instances with boosting trees is $\mathcal{O}(n_{e} \cdot k \cdot l_{max} \cdot d_{t}^2)$ \citep{lundberg2017consistent}. In practice, $k$, $d_{t}$ and $l_{max}$ ($l_{max}=2^{d_{t}}$) are constants; in this case we use $k=100$ and $d_{t}=3$, which results in $\mathcal{O}(n_{e})$. For IME, the computational complexity of generating explanations is likewise linear in the number of explained instances and attributes  $\mathcal{O}(n_{e} \cdot m)$ \citep{Strumbelj-2010}.

The computational complexity for identifying groups of co-occurring attributes in explanations is $\mathcal{O}(n_{e} \cdot m)$, and for constructing features is $\mathcal{O}(g_{n} \cdot g_{m}^{d_{f}} \cdot op_{n})$, where we use $max(d_{f})=3$. The evaluation part is bounded by $\mathcal{O}(n_{t} \cdot g_{n} \cdot g_{m}^{d_f} \cdot op_{n}$). 
If we omit the computational complexity of building the prediction model for explanations, the total computational complexity of the EFC methodology is $\mathcal{O}(n_{e} \cdot m+ n_{e} \cdot m + g_{n} \cdot g_{m}^{d_f} \cdot op_n + n_{t} \cdot g_{n} \cdot g_{m}^{d_f} \cdot op_n)$. For all practical purposes $g_{n} \cdot g_{m}^{d_f} \cdot op_{n}$ is constant (see distributions of these quantities for a large collection of UCI datasets in   \Cref{fig:numOfGeneratedGroups}). The computational complexity is therefore linear in the number of explanation instances and attributes $\mathcal{O}(n_{e} \cdot m+ n_{t})$.

\subsection{Implementation of the proposed approach}
\label{sec:implementation}
In this section, we elaborate on the proposed methodology outlined above by presenting the pseudo code of individual steps.

The main flow of the proposed approach is shown in \textbf{\Cref{alg:MainAlgorithm}}. The input parameters are the explanation instances $I^e$, the training instances $I^t$, the prediction model $M$, the class to explain $c$, the lower and upper weight thresholds ($thr_{l}$, $thr_{u}$) with the step $step$, and the noise threshold $noiseThr$. For each explanation threshold $q$, the algorithm generates a group of candidate attributes $G_{temp}$ from the matrix of explanations $E$ and then constructs a set of features $F$ from the set of candidates $G$. The output of the algorithm is the set of constructed features $F$.

The  components of the EFC method are located as follows. \Cref{alg:ExplanationsA} shows generation of instance explanations. \Cref{alg:SetWeights} and \Cref{alg:getMostFqSubset} describe the detection of the most frequently co-occurring groups of attributes in the explanations. A set of features is generated from each group with Algorithms \ref{alg:genFeatures} and \ref{alg:ConstructFeatures} and then evaluated.

\begin{algorithm}[hbt]
\caption{EFC: Explanation-based Feature Construction}
\label{alg:MainAlgorithm}
\footnotesize
\renewcommand{\algorithmicrequire}{\textbf{Input:}}
\renewcommand{\algorithmicensure}{\textbf{Output:}}
\begin{algorithmic}
\Require {Explanation instances $I^e=(X^{e \times m},Y^{1 \times e})$}\\ 
\hphantom{Input:}Training instances  $I^t=(X^{n \times m},Y^{1 \times n})$\\
\hphantom{Input:}Prediction model $M$ \\ 
\hphantom{Input:}Class value to explain $c\in \{1..C\}$ \\
\hphantom{Input:}Explanation method $P$ \\
\hphantom{Input:}Weight threshold $thr_{l}$, $thr_{u}$, $step$ \Comment{$0 < thr_{l} < thr_{u} \leq 1$, $0 < step < 1$}\\
\hphantom{Input:}Noise threshold $noiseThr$\Comment{Expected percentage of noise in the target class.}
\Ensure{Constructed features $F$}
	\State $G \gets \{\}$\Comment{set of candidate groups}
	\State $E \gets$ getExplanations$(I^e,M,c,P)$\Comment{using Alg. \ref{alg:ExplanationsA}}
	\For{$q \gets thr_{l}$ \textbf{to} $thr_{u}$ \textbf{by} $step$}\Comment {defaults: $thr_{l}=0.4$, $thr_{u}=0.8$, $step=0.1$}
		\State $W^{e \times m} \gets$ setWeights$(E,q)$\Comment{using Alg. \ref{alg:SetWeights}}
		\State $G_{temp} \gets$ getMostFqSubsets$(W,noiseThr)$\Comment{using Alg. \ref{alg:getMostFqSubset}}
		\State $G \gets G \cup G_{temp} $\Comment{Using LinkedHashSet to avoid duplicates}
	\EndFor
	\State $F \gets$ constructFeatures$(X^{n \times m},G,c) $\Comment{using Alg. \ref{alg:genFeatures}}
	\State evaluate$(F,X^{n \times m})$\Comment{Evaluate features and keep only useful ones} 
	\State \Return $F$
\end{algorithmic}
\end{algorithm}

\textbf{\Cref{alg:ExplanationsA}} explains each individual instance with an explanation method (e.g., SHAP or IME) and returns the matrix of explanations $E$. The algorithm inputs are the explanation instances $I^e$, the pre-trained model $M$, the explanation method P, and the class to explain $c$. The output is the matrix of instance explanations $E$.

\begin{algorithm}[htb]
    \caption{Get explanations for one class}
    \label{alg:ExplanationsA}
    \footnotesize
    \begin{algorithmic}
    \renewcommand{\algorithmicrequire}{\textbf{Input:}}
    \renewcommand{\algorithmicensure}{\textbf{Output:}}
    \Require {Explanation instances $I^e=(X^{e \times m},Y^{1 \times e})$}, Model $M$\\
    \hphantom{Input:}Class to explain $c$, Explanation method $P$
    \Ensure{Explanations $E$}
	    \ForAll{$x_{i}\in X$}
		    \State $E_{i} \gets$ explain$(x_{i},M,c,P)$
    	\EndFor
    \end{algorithmic}
\end{algorithm}

\textbf{\Cref{alg:SetWeights}} selects groups of candidate interactions by setting their weights to 1 and zeroing all the others. The input parameters are the matrix of explanations $E$ and the explanation threshold $q$. The output of the algorithm is the matrix of weights $W$ initialised to all zeros. The sum of normalised explanations is:
\begin{equation}\label{eq:SumOfNormalizedValues}
    s_{j}=\sum_{j=1 \atop a[o[j]]<q}^m a[o[j]]
\end{equation}
where $a$ is the vector of normalised explanations, $o$ the vector of indexes and $q$ the weight threshold. The algorithm first normalises the absolute values of explanations for each instance and then orders them in descending order. The normalised values are summed (the highest values are taken first) in $s_{j}$, see \Cref{eq:SumOfNormalizedValues}, until the threshold value $q$ is exceeded. When the normalised value of attribute is added to $s_{j}$, the weight of that attribute is set to 1 in the weight vector $w_{i}$.

\begin{algorithm}[hbt]
\caption{Select largest explanations}
\label{alg:SetWeights}
\footnotesize
\renewcommand{\algorithmicrequire}{\textbf{Input:}}
\renewcommand{\algorithmicensure}{\textbf{Output:}}
\newcommand{\pluseq}{\mathrel{+}=}
\begin{algorithmic}
\Require {Explanations $E^{e \times m}$}, Importance threshold $q$
\hphantom{Input:}
\Ensure{$W^{e \times m}$} \Comment{matrix of important explanations}
	\State {$E \gets |E|$} \Comment{get absolute values of explanation scores}
	\State $W \gets 0$ \Comment{initialise all values to zero}
	\ForAll{$e_{i}\in E$} \Comment{for all instances $E_{i}$; $i=1..e$}
	\State $a \gets ||e_{i}||$\Comment {normalise sum of explanations to 1}
		\State $o \gets$ order$(a)$\Comment {get descending order of explanations}
		\State $thrTmp \gets 0$
		\For{$j \gets 1$ \textbf{to} $m$}
			\If {$thrTmp < q$}
				\State {$thrTmp \pluseq a[o[j]]$}\Comment {add next biggest explanation}	
				\State $W_{i,o[j]} \gets 1$\Comment {this explanation is above threshold}
				\Else
				\State break \Comment{When the sum of the normalised explanations exceeds the threshold.}
			\EndIf
         \EndFor	
	\EndFor
	\State \Return $W$
\end{algorithmic}
\end{algorithm}

In \textbf{\Cref{alg:getMostFqSubset}}, we first generate sets of candidate groups $G^{e}$ from the matrix $W$, and calculate the frequency of subsets taking into account all possible subsets of given $G_{i}$. The subsets with frequency below the noise threshold $noiseThr$ and subsets with only one attribute are removed.
The input parameters are the matrix of important explanations $W$ and the noise threshold $noiseThr$ that determines the amount of required empirical support for candidate groups, i.e. the minimal required frequency to accept interaction as important. The output of the algorithm is a set of the most common co-occurring subsets of attributes $U$ from the matrix $W$.

\begin{algorithm}[!htb]
\caption{Get most frequent subsets}
\label{alg:getMostFqSubset}
\footnotesize
\renewcommand{\algorithmicrequire}{\textbf{Input:}}
\renewcommand{\algorithmicensure}{\textbf{Output:}}
\begin{algorithmic}
\Require {Matrix of important explanations $W^{e \times m}$}, Noise threshold $noiseThr$
\Ensure{Set $U$}\Comment{set of interaction candidates}
	\State $G^{e} \gets \{\}$ \Comment{set of candidate interactions, one for each instance}
	\ForAll{$w_{i}\in W$}\Comment{important explanation indicators for instance i}
		\For{$j \gets 1$ to $m$}\Comment{for all attributes}
			\If{$w_{i,j} = 1$} \Comment{explanation above threshold}
				\State $G_{i} \gets G_{i} \cup j$\Comment{add to the set of candidate interactions for explanation i}
			\EndIf
		\EndFor       	
	\EndFor
	\State $Q$ $\gets$ frequencyMap$(G)$ \Comment{get frequency of subsets}
	\State sortDesc$(Q)$\Comment{sort by frequency of subsets}
	\State $U \gets$ remove$(Q$, noiseThr) \Comment{remove subsets below the noise threshold and subsets with only one attribute}
	\State \Return $U$
\end{algorithmic}
\end{algorithm}

The \textbf{\Cref{alg:genFeatures}} describes feature construction. The input parameters are the training instances $I^t$, the collection of the most frequent candidate interactions $U$, the class to explain $c$, the percentage of covered instances $pci$, the threshold for certainty factor $cf$, the types of features $R$, and the set of operators $O$. The algorithm constructs operator-based features (using logical, relational, Cartesian, and numerical operators), features from the FURIA decision rule learning algorithm \citep{Huhn-2009-Furia}, and threshold features using a threshold for presence of several other features (e.g., \textit{num-of-N} and \textit{X-of-N} constructs). \Cref{alg:genFeatures} first generates operator-based features with \Cref{alg:ConstructFeatures} and merges them with features from FURIA. The merged set is used in the construction of threshold features. Each constructed feature is tested for sufficient coverage. When the number of covered instances reaches the threshold ($pci \times n_{c}$), feature construction is stopped. The output of the algorithm is the set of constructed features $S$.

\begin{algorithm}[!htb]
\caption{Generate features \Comment {all types of features}}
\footnotesize
\label{alg:genFeatures}
\renewcommand{\algorithmicrequire}{\textbf{Input:}}
\renewcommand{\algorithmicensure}{\textbf{Output:}}
\begin{algorithmic}
\Require {Instances $I^t=(X^{n \times m},Y^{1 \times n})$, Collection of the most frequent candidate interactions $U$, Class to explain $c$, Percentage of covered instances $pci$, Threshold for certainty factor $cf$}, Types of features $R$, Set of operators $O$
\Ensure{Set of features $S$}
	\State $S \gets \{\}$\\ $F \gets \{\}$ \Comment {temporary features}
	\ForAll{$r_{i}\in R$} \Comment{construct different types of features, using different operators}
	    \State $S \gets$ generateOperatorBasedFeatures($r_{i},op_{i}$) \Comment{using Alg. \ref{alg:ConstructFeatures}}
	\EndFor
	\For{$i \gets 1$ \textbf{to} $|U|$} \Comment {for all groups of frequently interacting features}
		\State $F \gets$ construct features from $U_{i}$ \Comment {using Furia and $cf$}
			\ForAll{$\phi_{i}\in F$}\Comment {interval and values using $\land, \lor$} 
				\If{$\phi_{i} \notin S $}
					\State $S \gets S \cup \phi_{i}$\Comment{add $\phi_{i}$ to $S$}
					\State remove instances covered by $\phi_{i}$ from $X$ where $Y=c$
				\EndIf
				\If {$\phi_{i}$ contains $\land$}
				    \State $args \gets$ split($\phi_{i}, \land$) \Comment{array of arguments/conditions}
      		         \State $S  \gets S$ $\cup$ createThresholdFeature($args$) \Comment{num-of-N}
      		    \EndIf
      		\EndFor
      		\If{\# covered instances $\geq pci \times n_{c}$}	\Comment{sufficient amount of covered instances}
      			\State break
      		\EndIf
	\EndFor
	\State \Return $S$
\end{algorithmic}
\end{algorithm}

In \textbf{\Cref{alg:ConstructFeatures}}, we create different types of operator-based features using, e.g., logical, relational, Cartesian, and numerical operators. Each d-element subset of attributes is used for feature construction with operators $op$. For logical $\lor$ and $\land$ we use d=3, for other logical operators and other types of features we use d=2. Constructed features are added to $S$ if not already present. The input parameters are the training instances $I^t$, the collection of the most frequent candidate interactions $U$, the depth of feature $d$, and the set of the operators $O$ for the chosen type of features. The output of the algorithm is the set of operator-based features $S$.

\begin{algorithm}[!htbp]
\caption{Construct operator-based features}
\label{alg:ConstructFeatures}
\footnotesize
\renewcommand{\algorithmicrequire}{\textbf{Input:}}
\renewcommand{\algorithmicensure}{\textbf{Output:}}
\begin{algorithmic}
\Require {Instances $I^t=(X^{n \times m},Y^{1 \times n})$}\\
\hphantom{Input:}{Set $U$} \Comment {set of candidate groups} \\
\hphantom{Input:}Depth of feature $d$\Comment {e.g., d=2, $A_{1} \lor A_{2}$; d=3, $A_{1}\lor A_{2}\lor A_{3}$}\\
\hphantom{Input:}Type of feature $r$\Comment{e.g., logical operators, relational, Cartesian, numerical}\\
\hphantom{Input:}Set $O$\Comment{set of operators for the chosen type of features}
\Ensure{Set of features $S$}\\
$V \gets \{\}$\Comment {set of d-element subsets of attributes}
	\For{$i \gets 1$ to $|U|$}\Comment {for all groups of interacting attributes}
		\State $V \gets$ generate d-element subsets from $U_{i}$
		\ForAll{$v \in V$}
		    \ForAll{$op \in O$} 
			\State $\phi \gets$ generateFeature($v,op,r$)		
				\State $S \gets S \cup \phi$
			\EndFor	
		\EndFor
	\EndFor
	\State \Return $S$
\end{algorithmic}
\end{algorithm}

\subsection{A worked example}
\label{sec:example}
To illustrate the proposed methodology, we execute the steps in \Cref{fig:methodology} using a toy dataset that consists of 6 binary attributes $A_{1}, A_2\dots A_{6}$, where the class $C$ is determined with the expression:  if $A_{1}==0$ then $C=A_{2}\land A_{3}$ else $C=A_{4}\land A_{5}$. 
 $A_{6}$ is unrelated to the binary class $C$. The attribute $A_{1}$ divides the problem space into two subspaces. The dataset is imbalanced and consists   of 1511 instances of majority class 0 and 489 instances of minority class 1. We explain the minority class. A few instances from the dataset are shown in \Cref{simple-dataset}.
 
\begin{table}[!htb]
\caption{A few instances from the illustrative dataset.}
\label{simple-dataset}
\centering
\begin{tabular}{ccccccccc}
\hline
\multicolumn{1}{c}{\bfseries $i$}   &
\multicolumn{1}{c}{\bfseries $A_{1}$}   & 
\multicolumn{1}{c}{\bfseries $A_{2}$}   & 
\multicolumn{1}{c}{\bfseries $A_{3}$}   & 
\multicolumn{1}{c}{\bfseries $A_{4}$}   & 
\multicolumn{1}{c}{\bfseries $A_{5}$}   &
\multicolumn{1}{c}{\bfseries $A_{6}$}   &
\multicolumn{1}{c}{\bfseries $C$}
\\ 
\hline
\tiny{51}&1&0&0&0&0&0&0&\\
\tiny{52}&0&0&0&1&1&1&0&\\
\rowcolor[gray]{0.8} \tiny{53}\cellcolor{white}&0&1&1&1&0&1&1&\cellcolor{white}
\rdelim\}{-1.1}{-\tabcolsep}[\scriptsize {instance from the first subspace}]\\
\tiny{54}&1&0&0&1&1&1&1&\\
\tiny{55}&1&0&0&1&0&0&0&\\
\rowcolor[gray]{0.8} \tiny{56}\cellcolor{white}&1&0&1&1&1&1&1&\cellcolor{white} \rdelim\}{-1.1}{-\tabcolsep}[\scriptsize {instance from the second subspace}]\\
\tiny{57}&0&1&1&0&0&0&1&\\
\tiny{58}&1&0&1&0&0&1&0&\\
\tiny{59}&1&1&0&1&1&0&1&\\
\tiny{60}&0&1&0&1&0&1&0
\\
\hline
\end{tabular}
\end{table}

\begin{figure}[h!tb]
    \centering
    \caption{Explanation of instance 53 from the first subspace of the illustrative example.}
     \includegraphics[keepaspectratio,width=0.7 \linewidth]{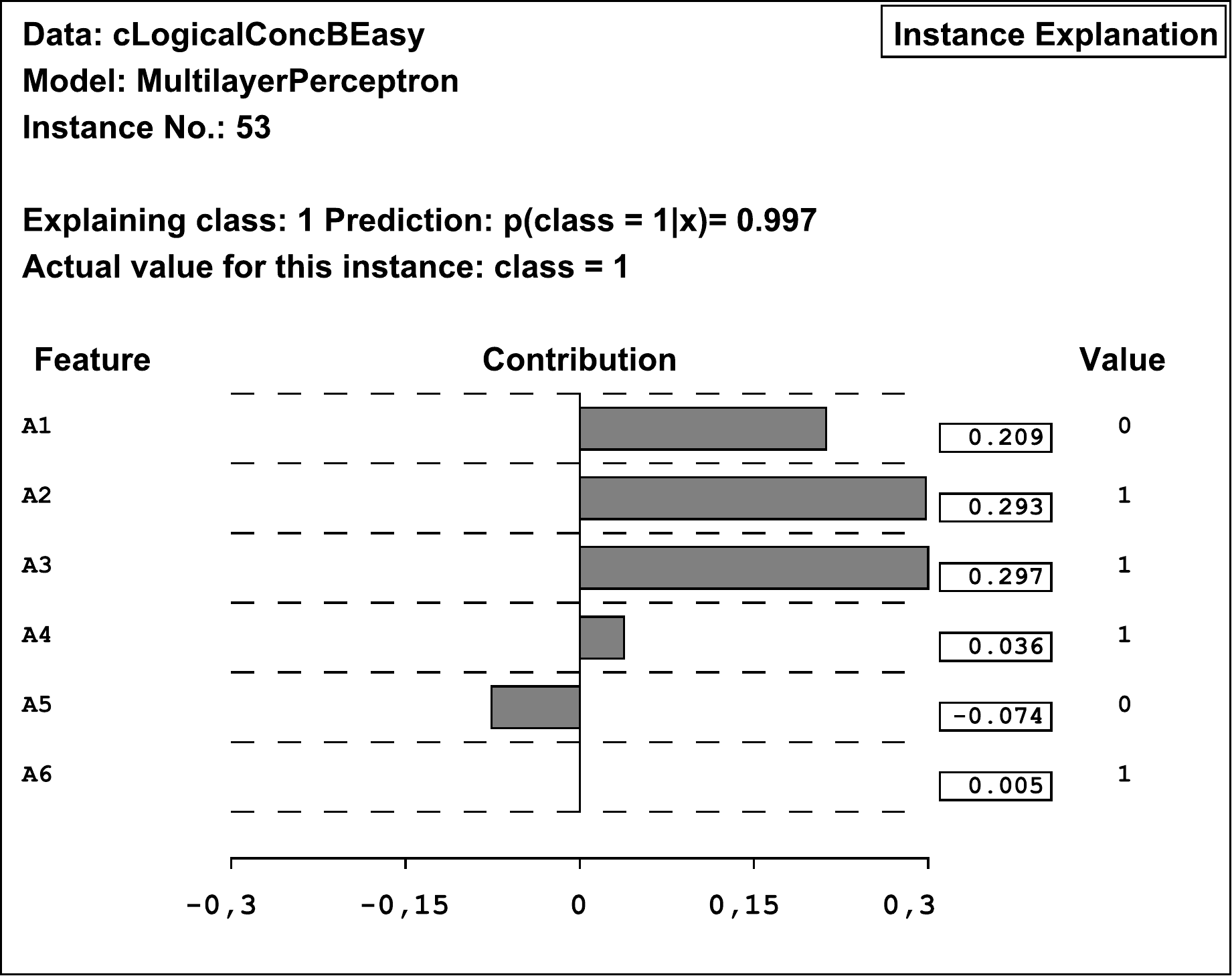}
    \label{fig:ExplanationConcept1}
\end{figure}

As described in \Cref{sec:firstStepMethodology}, we first train the classifier (in this example this is a multilayer perceptron) using the above-described dataset, and explain all instances from the minority class using the IME explanation method. For example, as the explanation of instance 53 ($A_{1}=0$, $A_{2}=1$, $A_{3}=1$, $A_{4}=1$, $A_{5}=0$, $A_{6}=1$) we get $E_{53}=$(0.209, 0.293, 0.297, 0.036, -0.074, 0.005). The explanation is illustrated in Figure \ref{fig:ExplanationConcept1}, which shows the contributions of attributes for classification of instance 53 into class 1. 

On the left-hand side of the vertical axis, the names of the attributes are listed, on the right-hand side (in boxes) the contributions of the attributes for this instance, and on the far right the attribute values. The bars correspond to the contributions of the attributes to the prediction. The first three attributes satisfy the condition $A_{1}=0$ and $A_{2}=1$ and $A_{3}=1$ and are in favour of class 1. The model correctly predicted class 1 with the probability of 0.997 and assigned a large positive contribution to the first three attributes. The contributions of $A_{4}$ and $A_{5}$ are small and the contribution of $A_{6}$ is almost zero.

The second step (see \Cref{sec:secondStepMethodology}) determines the largest explanation scores for each instance for different thresholds; we use lower threshold $thr_{l}=0.6$, upper threshold $thr_{u}=0.8$, and $step=0.1$. For each instance, we take the absolute values of the explanations, normalise them to 1, and sort them in descending order. The result for instance 53 is shown in \Cref{normalized-explanations}. 

\begin{figure}[htb]
\centering
\caption{Explanations normalised to 1 and ordered.}
\begin{tabular}{|c|c|c|c|c|c|}
\multicolumn{1}{c}{\bfseries $A_{3}$}   & 
\multicolumn{1}{c}{\bfseries $A_{2}$}   & 
\multicolumn{1}{c}{\bfseries $A_{1}$}   & 
\multicolumn{1}{c}{\bfseries $A_{5}$}   & 
\multicolumn{1}{c}{\bfseries $A_{4}$}   & 
\multicolumn{1}{c}{\bfseries $A_{6}$}
\\ 
\hline
0.325 & 0.321 & 0.229 & 0.081 & 0.039 & 0.005\\
\hline
\end{tabular}
\label{normalized-explanations}
\end{figure}

To determine groups of co-occurring attributes, we iteratively set weights of the largest contributions to 1 until the sum of the normalised explanations exceeds the threshold. For the instance 53 in \Cref{fig:ExplanationConcept1} and the threshold of 0.6, the sum of the first two explanations is $0.646$ and we mark $A_{2}$ and $A_{3}$ as a candidate group by setting their weights to 1, as shown in \Cref{fig:weights-of-an-instance}.

\begin{figure}[htb]
\centering
\caption{Instance weights after applying the threshold ($q=0.6$).}
\begin{tabular}{|c|c|c|c|c|c|}
\multicolumn{1}{c}{\bfseries $A_{1}$}   & 
\multicolumn{1}{c}{\bfseries $A_{2}$}   & 
\multicolumn{1}{c}{\bfseries $A_{3}$}   & 
\multicolumn{1}{c}{\bfseries $A_{4}$}   & 
\multicolumn{1}{c}{\bfseries $A_{5}$}   & 
\multicolumn{1}{c}{\bfseries $A_{6}$}
\\ \hline
0 & 1 & 1 & 0 & 0 & 0\\
\hline
\end{tabular}
\label{fig:weights-of-an-instance}
\end{figure}

With the threshold 0.7, we add also $A_{1}$ to the group and the vector of weights is $w_{53}=(1,1,1,0,0,0)$. The vector of weights now contains all relevant attributes for this instance. 

We select candidate groups of attributes for each threshold and each explained instance. The results are presented in \Cref{tab:concepts}, which also shows the frequencies of candidate groups. For example, for threshold $0.6$, $A_{2}$ and $A_{3}$ appear together in 249 (50.9\%) explanations. The result of candidate groups identification step is  as shown in \Cref{fig:groupsOfInterAttr}.
\begin{table}[htb]
\caption{Identification of frequently co-occurring groups of attributes for each threshold (sorted in descending order).}
\label{tab:concepts}
\centering
\begin{tabular}{lr|lr|lrl}
\hline
\multicolumn{2}{c}{\bfseries $thr=0.6$}     & 
\multicolumn{2}{c}{\bfseries $thr=0.7$}     & 
\multicolumn{2}{c}{\bfseries $thr=0.8$}    &
\\ \hline
$A_{2},A_{3}$&249&$A_{1},A_{2},A_{3}$&187&$A_{1},A_{2},A_{3}$&187&\\
$A_{4},A_{5}$&237&$A_{1},A_{4},A_{5}$&180&$A_{1},A_{4},A_{5}$&180&\\
$A_{1},A_{2},A_{3}$&3&$A_{2},A_{3}$&65&$A_{3},A_{4},A_{5}$&36&\\
&&$A_{4},A_{5}$&57&$A_{2},A_{3},A_{5}$&36&\\
&&&&$A_{2},A_{3},A_{4}$&29&\\
&&&&$A_{2},A_{4},A_{5}$&21&\\
\hline
\end{tabular}
\end{table}

\begin{figure}[htb]
\centering
\caption{The collection of candidate groups.}
\begin{tabular}{|c|c|c|c|c|c|c|}
\hline
$A_{2},A_{3}$&{$A_{4},A_{5}$}&$A_{1},A_{2},A_{3}$&$A_{1},A_{4},A_{5}$&$A_{3},A_{4},A_{5}$&$A_{2},A_{3},A_{5}$&{...}\\
\hline
\end{tabular}
\label{fig:groupsOfInterAttr}
\end{figure}

In the feature construction step (see  \Cref{sec:thirdStepMethodology}), we generate operator-based features (using logical operators \{$\equiv,\oplus,\Rightarrow$\}), features of decision rule learner FURIA, and threshold features (\textit{num-of-N}). We create 30 operator-based features using pairs of attributes from the candidate groups. 
For the FURIA algorithm, we set the threshold for the certainty factor to $cf=0.9$ and the threshold for the percentage of covered instances to $pci=90\%$. In our method, the FURIA algorithm does not work on all attributes, but searches for good rules using attributes from individual candidate groups.

FURIA iterates through all groups of attributes, starting with the first group, i.e., \{$A_{2},A_{3}$\}. The algorithm  does not generate any features from the first and second group. For the group \{$A_{1},A_{2},A_{3}$\}, the algorithm generates feature ($A_{2}=1) \land (A_{3}=1) \land (A_{1}=0$) which covers 225 out of 489 instances from class 1. The remaining 237 instances are covered with the group \{$A_{1},A_{4},A_{5}$\} (see \Cref{alg:genFeatures} in Section \ref{sec:implementation} for details).

The threshold features are generated from the features previously generated by FURIA. In this example, two threshold features are generated: \textit{num-of-N}$((A_{2}=1), (A_{3}=1), (A_{1}=0))$ and \textit{num-of-N}$((A_{4}=1), (A_{5}=1), (A_{1}=1))$. The \textit{num-of-N} construct counts the true conditions in the set of N specified conditions. In this case, possible values for these features are integers from 0 to 3. 

In the feature evaluation step (see \Cref{sec:fourthStepMethodology}), we use the MDL measure. The results are shown in \Cref{tab:featuresV3}. We can see that two directly expressed parts of the target concept are ranked first.

\begin{table}[tb]
\caption{The evaluation of generated features ordered by the decreasing MDL score.}
\label{tab:featuresV3}
\centering
\begin{tabular}{lr}
Constructs & MDL score \\
\hline
\textit{num-of-N}$((A_{2}=1), (A_{3}=1), (A_{1}=0))$&0.32\\
$(A_{2}=1)\land (A_{3}=1) \land(A_{1}=0)$ &0.30\\
\textit{num-of-N}$((A_{4}=1), (A_{5}=1), (A_{1}=1))$&0.29\\
$(A_{4}=1)\land (A_{5}=1) \land(A_{1}=1)$ &0.28\\
$A1 \Rightarrow A5 $&0.12\\
$A1 \Rightarrow A4 $&0.12\\
$A1 \equiv A3 $&0.08\\
$A1 \oplus A3 $&0.08\\
$A2 \equiv  A3 $&0.07\\
$\vdots$&\\
$A2 \Rightarrow A4$&0 \\
\hline
\end{tabular}
\end{table}

If we add some noise to the class value by reverting it in 5\% of the cases, we get the same results with the additional, irrelevant group $\{A_{3},A_{4},A_{5}\}$. The FURIA constructs cover correctly classified instances from class 1 and leave the noisy instances uncovered.

\section{Evaluation settings} 
\label{lab:evaluation}
We evaluate the proposed methodology in three settings. Using synthetic datasets, we analyse its properties. Using real-world UCI datasets, we compare it to other constructive induction approaches and establish its performance. The use case from the financial industry shows how our methodology can be valuable in practice.
We start by presenting the three groups  of datasets (synthetic, UCI, and real-world use case), followed by a short description of compared feature construction methods and the experimental settings. Finally, we present the settings of the methods used in the evaluation. 

\subsection{Datasets}
The empirical evaluation uses several synthetic and real-world classification datasets. The synthetic datasets are presented in \Cref{sec:syntheticData} and summarised in \Cref{tab:ArtDatasets-summary}. The UCI datasets are presented in \Cref{sec:UCIdatasets} and  summarised in \Cref{tab:UCIDatasets-summary}. Finally, the dataset describing the financial use case is presented in \ref{realDataset}.

\subsubsection{Synthetic data}
\label{sec:syntheticData}
With synthetic datasets we test the proposed methodology on different types of concepts (conjunction, disjunction, xor, equivalence) and with different types of dependencies between attributes (conditionally independent attributes, redundant attributes, random attributes, etc.). The summary of these datasets is given in \Cref{tab:ArtDatasets-summary}. 
We created 10 synthetic datasets with numerical, nominal and binary attributes. The numerical attributes contain values from [0,1] interval.  The values of nominal attributes can be 0, 1, or 2.  Some of the datasets were used in past studies related to explanation methodology \citep{robnik2008explain, Strumbelj-2010, Strumbelj-2014, Strumbelj-2009}. Each dataset consists of 2000 instances. A short description of datasets is presented below.

\begin{table}[!htbp]
\caption{The summary of synthetic datasets. The columns show the number of instances (\#I), number of attributes (\#A), number of unrelated attributes (\#U), number of nominal attributes (\#Nom), number of numeric attributes (\#Num), number of classes (\#C), and \% of noise.}
\label{tab:ArtDatasets-summary}
\centering
\begin{tabular}{lccccccc}
\hline
\multicolumn{1}{l}{\bfseries Name}  & 
\multicolumn{1}{c}{\bfseries \#I}   & 
\multicolumn{1}{c}{\bfseries \#A}   & 
\multicolumn{1}{c}{\bfseries \#U}   & 
\multicolumn{1}{c}{\bfseries \#Nom} & 
\multicolumn{1}{c}{\bfseries \#Num} &
\multicolumn{1}{c}{\bfseries \#C}   &
\multicolumn{1}{c}{\bfseries Noise}
\\ \hline
LogicalConcB &         2000 & 7 & 1 & 7 & 0 & 2 & 0\\
LogicalConcBNoisy &    2000 & 7 & 1 & 7 & 0 & 2 & 5\\
BinClassDisAttr&       2000 & 5 & 1 & 5 & 0 & 2 & 0\\
BinClassNumBinAttr &   2000 & 5 & 1 & 2 & 3 & 2 & 0\\
BinClassNumDisAttr &   2000 & 5 & 1 & 2 & 3 & 2 & 0\\
DisjunctN &            2000 & 5 & 2 & 0 & 5 & 2 & 0\\
MultiVClassDisAttr &   2000 & 5 & 1 & 5 & 0 & 3 & 0\\
Concept &              2000 & 5 & 1 & 5 & 0 & 2 & 0\\
ModGroups &            2000 & 4 & 2 & 0 & 4 & 3 & 10\\
CondInd &              2000 & 8 & 4 & 8 & 0 & 2 & 0 \\
\hline
\end{tabular}
\end{table}

\par\noindent \textbf{LogicalConcB}\quad{The dataset consists of 7 binary attributes, the last of which is unrelated to the binary class $C$. Attribute $A_{2}$ divides the problem space into two 2 subspaces: 
\\if $A_{2}==0$ then  $C=A_{1} \overline {A_{3}}$ else $C$=$A_{4}A_{5} \lor A_{6}$}
\par\noindent \textbf{LogicalConcBNoisy}\quad{Dataset has the same concept as LogicalConcB, but contains 5\% class noise.}
\par\noindent \textbf{BinClassDisAttr}\quad{All five attributes are nominal, $A_{5}$ is unrelated to the class, the class is binary. The attribute $A_{1}$ divides the problem space into 3 sub-problems}:
\quad{
\\if $A_{1}==2$ then $C=(A_{2}==0) \land (A_{3}==1)$
\\else if $(A_{1}==1)$ then $C=(A_{4}==2) \lor (A_{3}==2) \lor (A_{2}\neq A_{1})$ 
\\else $C=(A_{4}==0)$ }

\par\noindent \textbf{BinClassNumBinAttr}\quad{The dataset consists of two binary attributes ($A_{1}, A_{2}$) and three numerical attributes ($A_{3}, A_{4}, A_{5}$). Attribute $A_{5}$ is unrelated to the class. The class value is binary.
\\if $A_{1}==0$ then $C=(A_{3}>0.3) \land (A_{4}<0.1)$  
\\else $C=(A_{2}==0) \land (A_{3}>0.7)$ }

\par\noindent \textbf{BinClassNumDisAttr}\quad{Dataset is similar to dataset BinClassNumBinAttr. The difference is that the first two attributes ($A_{1}, A_{2}$) are nominal.
\\if $A_{2}==0$ then $C=(A_{3}<0.5) \land (A_{1}==0)$  
\\else if $(A_{2}==1)$ then $C=(A_{4}>0.15) \land (A_{1}==2)$  \\else $C=(A_{3}>0.5) \land (A_{4}>0.5) \land (A_{1}==1)$ }

\par\noindent \textbf{DisjunctN}\quad{(disjunction with numerical attributes) The dataset consists of 5 numerical attributes. The first three are important for classification, the last two are unrelated to the class. The class value is binary.
\\ $C = (A_{1}>0.5 \lor A_{2}>0.7 \lor A_{3}<0.4)$} 
\par\noindent \textbf{MultiVClassDisAttr}\quad{All attributes in the data set are nominal, with the fifth attribute ($A_{5}$) unrelated to the class. The dataset simulates a three-class problem.
\\if $A_{1}==2$ then \{ if $(A_{2}==0) \land (A_{3}==1)$ then $C = 1$ else $C = 0$ \} 
\\else if $(A_{1}==1)$ \{ if $(A_{4}==2) \lor (A_{3}==2) \lor (A_{2}\neq A_{1})$ then $C=2$ else $C=1$ \} 
\\else \{ if $(A_{4}==0)$ then $C=2$ else $C=0$ \} }

\par\noindent \textbf{Concept}\quad{The data set consists of 5 binary attributes, the last is unrelated to the class.  The class value is binary. The attribute $A_{2}$ divides the problem space into two 2 subspaces:
\\if $A_{2}==0$ then $C=$ \textit{all-of-N}$(A_{1}==1, A_{3}==0,A_{3}==A_{4})$
\\else $C=$ \textit{num-of-N}$(A_{3}==1, A_{1}==A_{4})>0$ }

\par\noindent \textbf{ModGroups} \citep{robnik2008explain} \quad{The dataset consists of 4  numerical attributes. The first two $I_{1}$ and $I_{2}$ are important and represent the x and y axes. The last two attributes are unrelated to the class. The unit square is divided into nine equivalent squares, with each square representing one of three possible class values. An instance is correctly classified only if we have knowledge of both important attributes. Dataset contains 10\% noise; the noise was imputed by reverting 10\% class values.}

\par\noindent \textbf{CondInd} \citep{robnik2008explain} \quad{The dataset consists of 8 binary attributes, the first four are important and the other four are unrelated to the class. The class value is binary and the probability of each class is 50\%. In 90, 80, 70 and 60\% of the cases, the four important attributes correspond to the class. This dataset contains no dependent attributes.}

\subsubsection{UCI datasets}
\label{sec:UCIdatasets}
The summary of UCI datasets can be found in \Cref{tab:UCIDatasets-summary}. We collected datasets from several papers related to constructive induction \citep{fan2010generalized, gama1999discriminant, hammami2020feature, katz2016explorekit, markovitch2002feature,   muharram2005evolutionary, nargesian2017learning, pechenizkiy2005impact, zupan1999function} and attribute interactions \citep{jakulin2003analyzing, murthy2018generation, perez1995using, st2017sparse, tang2019feature, yazdani2017mbcgp, zeng2015mixed}. The Biological response dataset (taken from the Kaggle repository) was chosen to show usability of the methodology on the dataset with large number of attributes (marked with asterisk).

\begin{table}[!ht]
\caption{The summary of UCI datasets. The columns show the number of instances (\#I), number of attributes (\#A), number of nominal attributes (\#Nom), number of numeric attributes (\#Num), and number of classes (\#C).}
\label{tab:UCIDatasets-summary}
\centering
\begin{tabular}{lrrrrr}
\hline
\multicolumn{1}{l}{\bfseries Name}  & 
\multicolumn{1}{c}{\bfseries \#I}   & 
\multicolumn{1}{c}{\bfseries \#A}   & 
\multicolumn{1}{c}{\bfseries \#Nom} & 
\multicolumn{1}{c}{\bfseries \#Num} &
\multicolumn{1}{c}{\bfseries \#C}
\\ \hline
Australian credit   &   690     &   14  &   8   &   6   &   2\\
Autos               &   205     &   25  &   10  &   15  &   7\\
Balance scale       &   625     &   4   &   0   &   4   &   3\\
Bank marketing      &   4521    &   16  &   9   &   7   &   2\\
Biological response*&   3751    &   1776&   0   &   1776&   2\\
Car                 &   1728    &   6   &   6   &   0   &   4\\
CNAE-9              &   1080    &   856 &   0   &   856 &   9\\
Glass               &   214     &   9   &   0   &   9   &   7\\
Heart               &   270     &   13  &   0   &   13  &   2\\
Ionosphere          &   351     &   34  &   0   &   34  &   2\\
Japanese credit     &   690     &   15  &   9   &   6   &   2\\
Leukemia            &   72      &   7129&   0   &   7129&   2\\
Liver disorders     &   345     &   6   &   0   &   6   &   2\\
LSVT voice          &   126     &   310 &   0   &   310 &   2\\
Lung cancer         &   32      &   56  &   56  &   0   &   3\\
Mammographic mass   &   961     &   5   &   0   &   5   &   2\\
Molecular promoters &   106     &   57  &   57  &   0   &   2\\
Monks1              &   432     &   6   &   6   &   0   &   2\\
Monks2              &   432     &   6   &   6   &   0   &   2\\
Nursery             &   12960   &   8   &   8   &   0   &   5\\
Parkinsons          &   195     &   22  &   0   &   22  &   2\\
QSAR biodegradation &   1055    &   41  &   0   &   41  &   2\\
Semeion             &   1593    &   256 &   256 &   0   &   10\\
Sonar               &   208     &   60  &   0   &   60  &   2\\
Spambase            &   4601    &   57  &   0   &   57  &   2\\
Spect heart         &   267     &   22  &   22  &   0   &   2\\
Spectf heart        &   349     &   44  &   0   &   44  &   2\\
Tic-tac-toe         &   958     &   9   &   9   &   0   &   2\\
Vehicle             &   846     &   18  &   0   &   18  &   4\\
Voting              &   435     &   16  &   16  &   0   &   2\\
\hline
\end{tabular}
\end{table}

\subsubsection{Use case from the financial industry} 
\label{realDataset}
Credit ratings and credit risk assessments plays an important role in ensuring the financial health of financial and non-financial organisations. Understanding the financial statements can be significantly improved by the credit score arguments of a particular company \citep{ganguin2004standard}.

We use a dataset of annual financial statements and credit scores for 223 Slovenian companies. The data were obtained from an institution that specialises in issuing credit scores. Original (five class) dataset was split into two classes (good and bad) to simplify the learning process of machine learning approaches. The dataset contains 115 companies with the ``bad`` score and 108 companies with the ``good`` score. The domain expert selected 22 attributes that describe each company; nine from the income statement (net sales, cost of goods and services, cost of labor, depreciation, financial expenses, interest, EBIT, EBITDA, net income), eleven from the balance sheet (assets, equity, debt, cash, long-term assets, short-term assets, total operating liabilities, short-term operating liabilities, long-term liabilities, short-term liabilities, inventories), and two from the cash flow statement (FFO - fund from operations, OCF - operating cash flow).

Beside these attributes, the financial expert introduced 9 new attributes during the argument-based machine learning (ABML) process \citep{guid2012abml}. The additional attributes are: Debt to Total Assets Ratio, Current Ratio, Long-Term Sales Growth Rate, Short-Term Sales Growth Rate, EBIT Margin Change, Net Debt To EBITDA Ratio, Equity Ratio, TIE - Times Interest Earned, and ROA - Return on Assets. Descriptions of these concepts can be found in financial accounting literature, e.g., \citep{holt2001financial}.

\begin{table}[!ht]
  \caption{Baseline financial indicators and expert attributes (in bold) from the credit scoring problem ordered by their decreasing MDL score.}
  \centering
    \begin{tabular}{l:c}
    \textbf{Original attributes} & \textbf{MDL score} \\
    \hline
    \textbf{(Debt - Cash) / EBITDA \textit{(Net Debt To EBITDA)}}	&	0.548	\\
    \textbf{EBIT / Interest \textit{(TIE)}}	                        &	0.402	\\
    Net Income	                                                    &	0.306	\\
    \textbf{EBIT / Assets \textit{(ROA)}}	                        &	0.294	\\
    FFO	                                                            &	0.212	\\
    EBIT	                                                        &	0.194	\\
    \textbf{Equity / Assets \textit{(Equity Ratio)}}	            &	0.183	\\
    \textbf{Short-Term Assets / Short-Term Liabilities \textit{(Current Ratio)}}	    &	0.166	\\
    EBITDA	                                                        &	0.147	\\
    \textbf{Short-Term Sales Growth}	                            &  	0.107	\\
    \textbf{Long-Term Sales Growth}	                                &	0.094	\\
    Equity	                                                        &	0.091	\\
    OCF	                                                            &	0.074	\\
    Cash	                                                        &	0.072	\\
    \textbf{EBIT Margin Change}	                                    &	0.066	\\
    Debt	                                                        &	0.052	\\
    Interest	                                                    &	0.048	\\
    Financial expenses	                                            &	0.043	\\
    \textbf{Total Operating Liabilities / Asset \textit{(Debt To Assets)}}	&	0.033	\\
    Short-Term Liabilities	                                        &	0.021	\\
    Long-Term Liabilities	                                        &	0.018	\\
    Depreciation	                                                &	0.017	\\
    Short-Term Assets	                                            &	0.017	\\
    Assets	                                                        &	0.017	\\
    Total Operating Liabilities	                                    &	0.013	\\
    Short-Term Operating Liabilities	                            &	0.013	\\
    Long-Term Assets	                                            &	0.010	\\
    Cost Of Labor	                                                &	0.007	\\
    Cost Of Goods And Services	                                    &	0.005	\\
    Net Sales	                                                    &	0.005	\\
    Inventories	                                                    &	0.004	\\
    \hline
    \end{tabular}
    \label{tab:creditScoreAttrMDL}
\end{table}

\Cref{tab:creditScoreAttrMDL} lists all existing attributes in this domain, i.e., both baseline financial indicators as well as expert-derived indicators (in bold). The attributes are ordered by their MDL score (i.e., direct impact on the credit score class value). The ordering shows that expert-derived attributes are stronger predictors of the credit score, as evidenced from their dominance in the upper part of the list.

\subsection{Compared methods}
We compare EFC methodology with baseline and other feature construction methods. The only other constructive induction method we could find was presented by \cite{jakulin2005MLattrInteractionsPhD}. This method was implemented in one of the earlier versions of Orange data mining suite (Orange 2.7)\footnote{\url{https://orangedatamining.com/}} but is no longer available. As the implementation worked for larger datasets, we reimplemented it. This method is based on the interaction information as described in \Cref{sec:featureInteraction}. As explained in \Cref{sec:FC}, FICUS \citep{markovitch2002feature}  and HINT \citep{zupan1998feature}  constructive induction methods are no longer available\footnote{We obtained this information from the authors, S. Markovitch and J. Demšar.}. These methods are complex and their reliable reimplementation is questionable, so we omitted them from evaluation. Other constructive induction methods (CITRE \citep{matheus1989constructive}, FRINGE \citep{bagallo1990boolean}, IB3-CI \citep{aha1991incremental}, and LFC \citep{ragavan1993complex}) can be simulated with methodology by excluding the search space reduction based on explanations and by reducing the set of constructive operators. We analyse these reductions in \Cref{lab:results}. 

As baselines we use i) comprehensible classifiers (decision trees and Naive Bayes) without constructive induction, and ii) comprehensible classifiers using  constructive induction with the exhaustive search, where we use constructive operators with all combinations of attributes. We also compare the predictive performance of EFC with state-of-the-art non-interpretable classifier (i.e., XGBoost). The results of XGBoost serve as an upper-bound of predictive performance and are an indication of the loss suffered if we settle for interpretable classifiers (with and without constructive induction). These experiments use datasets from the UCI repository, summarised in \Cref{sec:UCIdatasets}.

\subsection{Experimental settings}
\label{sec:expSettings}
Unless mentioned otherwise, we use the following settings. As the explanation method, we use SHAP in combination with the XGBoost prediction model using the parameters \textit{numOfRounds=100}, \textit{maxDepth=3}, \textit{eta=0.3}, and \textit{gamma=1}. We explain the minority class value $c$. The parameter $instThr$, which controls the minimum number of explained instances, is set to 10\% of instances in the dataset. If the number of instances in the explained class $c$ is below $instThr$, we take another class that has at least $instThr$ instances.  We set the maximal number of explained instances ($maxToExplain$) to 500. If the instances from class $c$ exceed $maxToExplain$, we randomly sample $maxToExplain$ instances from class $c$.

The minimum weight threshold ($thr_{l}$) is set to 0.1 and the maximum weight threshold ($thr_{u}$) to 0.8, with the step 0.1. The noise threshold ($noiseThr$) for subgroups is set to 1\%. When constructing features with the FURIA learner, we use the certainty factor ($cf$) 0.6 and omit the coverage parameter ($pci$) to get the maximal number of rules/features related to the explained class.

All experiments used  2.93 GHz Intel Xeon X5670 CPU server with 32 GB RAM. Algorithms are implemented in Java (v1.8.0). The classifiers are taken from the Weka \citep{weka} machine learning package using the default settings.

\section{Results} 
\label{lab:results}
The results are presented in three subsections. \Cref{sec:resultsSynthetic} addresses the synthetic data, \Cref{sec:resultsUCI} contains results on the UCI data, and \Cref{sec:resultsFinance} covers the financial industry use case.

\subsection{Synthetic data}
\label{sec:resultsSynthetic}
Experiments on synthetic datasets test components and properties of the proposed methodology. We check detection of relevant attribute groups, and the performance of resulting classifiers. 

\subsubsection{Detection of relevant attribute groups}
We first check if EFC finds the relevant groups of attributes that appear in the definitions of datasets (see \Cref{sec:syntheticData}). The identified groups are essential for the success of feature construction phase and reduction of search space. The results are presented in \Cref{tab:groupsOfAttributes} where complete correct expressions from definitions are marked with blue colour and groups containing irrelevant attributes with red colour (e.g., in MultiVClassDisAttr and ModGroups dataset). Detected parts of expressions are underlined. Groups in black contain attributes from different subspaces of the problem, e.g., the group \{$A_{1}, A_{3}, A_{5}$\} from LogicalConcB consists of attributes $A_{1}$ and $A_{3}$ from one subspace and $A_{5}$ from the other subspace. Obtaining groups from mixed subspaces is not surprising as the group members were detected as relevant by the prediction model.  

\begin{table}[!htb]
\caption{Detected groups of co-occurring attributes in synthetic datasets. Blue groups represent complete expressions from the dataset definitions, and underlined groups contain parts of expressions. Groups containing attributes unrelated to the class are marked with red. Groups of relevant attributes from mixed subspaces are in black. }
  \label{tab:groupsOfAttributes}
    \centering
    \begin{tabular}{p{0.3\linewidth}  p{0.65\linewidth}}
    \toprule
    \textbf{Dataset} & \textbf{Detected groups of attributes}\\
    \midrule
    
    \multirow{5}[0]{*}{LogicalConcB } & \underline{A1 A3}, \underline{A2 A6}, \underline{A2 A4}, \underline{A4 A5}, \underline{A2 A4 A5}, A1 A3 A6, \underline{A4 A6}, \underline{A5 A6}, \textcolor{blue}{A1 A2 A3}, A1 A3 A5, \underline{A4 A5 A6}, \underline{A2 A4 A6}, A3 A4 A6, A1 A3 A4, \textcolor{blue}{A2 A4 A5 A6}, A1 A4 A6, \underline{A2 A5 A6}, A1 A2 A3 A6, A1 A3 A4 A6, A2 A3 A4 A6, A1 A2 A3 A5, A1 A4 A5 A6, A1 A2 A4 A6 \\
    \midrule
    \multirow{5}[0]{10mm}{LogicalConcBNoisy } & \underline{A1 A3}, \underline{A2 A6}, A2 A5, \underline{A2 A4}, \underline{A4 A5}, \underline{A2 A4 A5}, \underline{A4 A6}, \underline{A5 A6}, \underline{A2 A4 A6}, \underline{A2 A5 A6}, \underline{A4 A5 A6}, A1 A3 A6, \textcolor{blue}{A1 A2 A3}, A1 A3 A5, A1 A2 A6, \textcolor{blue}{A2 A4 A5 A6}, A1 A2 A3 A6, A2 A3 A6, A1 A2 A3 A5, A1 A2 A4 A6, A1 A3 A5 A6, A1 A3 A4 A6, A1 A4 A5 A6 \\
    \midrule
    BinClassDisAttr  & \underline{A2 A3}, \underline{A2 A4}, \underline{A1 A2}, \underline{A1 A3}, \textcolor{blue}{A1 A2 A3}, \underline{A1 A2 A4} \\
    \midrule
    BinClassNumBinAttr  & \underline{A2 A3}, \textcolor{blue}{A1 A2 A3}, \underline{A3 A4}, A2 A3 A4, \underline{A1 A4} \\
    \midrule
    BinClassNumDisAttr  & \underline{A1 A3}, \underline{A3 A4}, \underline{A1 A2}, \underline{A1 A3 A4}, \textcolor{blue}{A1 A2 A3}  \\
    \midrule
    DisjunctN & \underline{A1 A3}, \textcolor{blue}{A1 A2 A3} \\
    \midrule
    MultiVClassDisAttr  & A1 \textcolor{red}{A5}, A2 \textcolor{red}{A5}, \underline{A2 A4}, A1 A2 \textcolor{red}{A5}, A1 A4 \textcolor{red}{A5}, A2 A4 \textcolor{red}{A5} \\
    \midrule
    Concept  & \underline{A1 A2}, \underline{A2 A4}, \underline{A3 A4}, \underline{A1 A3}, \underline{A1 A3 A4}, \underline{A2 A3 A4}, \underline{A1 A2 A3} \\
    \midrule
    ModGroups  & \textcolor{blue}{I1 I2}, I1 \textcolor{red}{R2}, I1 I2 \textcolor{red}{R2} \\
    \midrule
    CondInd  & \textcolor{blue}{I90}, I80 I90, I70 I80 I90, I60 I80 I90 \\
    \bottomrule
    \end{tabular}
\end{table}

The results show that EFC ignores irrelevant attributes present in the datasets in 8 out of 10 cases. The robustness of the methodology was tested on two noisy datasets (LogicalConcBNoisy and ModGroups). EFC extracts all correct concepts from both datasets, however, for the ModGroups dataset it also extracts two groups with an irrelevant attribute. An error at this stage is not crucial, as the later stages of EFC can ignore irrelevant attributes. However, inclusion of irrelevant attributes unnecessarily increases the search space of construction. 

The CondInd dataset contains no relevant attribute groups (all attributes are conditionally independent of class). EFC correctly detects the attribute I90 which has the strongest connection with the class. Other groups contain several attributes because their joint probabilities are high enough (e.g., the joint probability of I90 and I80 is $0.9 \cdot 0.8 = 0.72$, which is still strongly relevant for the class detection and was therefore used in the prediction model). The results of the CondInd dataset show that EFC may detect spurious attribute interactions, which will be removed later in the construction and evaluation step.

\subsubsection{Predictive performance on synthetic data}
We check if constructed features improve classification accuracy (CA). We test several interpretable classifiers: decision trees (DT), Naive Bayes (NB), and classification rules obtained with the FURIA rule learner (FU). Besides these, we include a black-box random forest (RF) classifier. We expect that constructive induction will improve the performance of interpretable classifiers (DT and NB). Here the constructs improve the representation and should also improve the interpretability of models. RF uses feature construction internally and we use it to estimate the upper bounds of classification performance. FURIA is used internally by some of the constructive operators, so we expect improvements only in datasets where dataset definitions contain constructive operators not  covered by FURIA rules.

Without using constructive induction as a baseline (Base), we compare six types of constructed features. The first setting includes only logical operators (Log); we use operators $O_{l}=\{\land, \vee, \equiv,\oplus,\Rightarrow\}$. The second includes only relational (Rel) operators;  we use operators $O_{r}=\{<, \neq \}$. The third setting tests only Cartesian product (Cart) features, where we form the Cartesian product of values for a pair of attributes. The fourth setting uses only decision rule and threshold (DrThr) features; the decision rule features are expressions produced by the FURIA learner, and the threshold features count the number of correct logical expressions entered as their arguments. The fifth setting includes all the previously mentioned feature types (All), and the last, sixth setting constructs all types of features but applies feature selection (FS) using a validation dataset to measure their performance. In the FS setting, the training dataset is internally split into the training and validation part, allowing the search for optimal parameters in the training part and testing them on the validation part using CA as the criterion. The proportion of training and validation instances is a parameter of the EFC; we use the ratio 75:25 which is commonly used by ML practitioners. The best parameters are applied to the whole training dataset and the resulting datasets and models are tested on the testing set. The method could use other feature selection algorithms (e.g.,  MCEC \citep{azadifar2022graph}, \citep{rostami2022-gene}), or any other method covered in a recent survey \citep{rostami2021-review}. In this experiment, we use the MDL feature evaluation score \citep{kononenko-1995-biases} with three different thresholds (0, 0.25, and 0.5). In all six settings, the MDL feature evaluation measure is used to calculate the importance of generated features at the end of methodology, in the sixth setting (FS), however, it is also used in the feature selection phase. We leave testing and analysis of other parameters for future work.

The results are reported in \Cref{tab:accArtDatasetsFinal}. The best result for each dataset and classifier is underlined and results better or equal than the baseline (without feature construction) are typeset in bold.

\begin{table}[!htbp]
\caption{Classification accuracy on synthetic datasets. We report success of different types of constructive operators: logical  (Log), relational (Rel), Cartesian product (Cart), decision rules and threshold (DrThr), all previously mentioned (All), and all previously mentioned with feature selection (FS). As a baseline, we use no construction (Base). The best results for each dataset and classifier are underlined, the results better or equal than the baseline are in bold.}
\label{tab:accArtDatasetsFinal}
  \centering
  \setlength{\tabcolsep}{11pt}
  \resizebox{\linewidth}{!}{
    \begin{tabular}{llrrrrrrr}
    \toprule
    \textbf{Dataset} & \textbf{classif.} & \multicolumn{1}{l}{\textbf{Base}} & \multicolumn{1}{l}{\textbf{Log}} & \multicolumn{1}{l}{\textbf{Rel}} & \multicolumn{1}{l}{\textbf{Cart}} & \multicolumn{1}{l}{\textbf{DrThr}} & \multicolumn{1}{l}{\textbf{All}} & \multicolumn{1}{l}{\textbf{FS}} \\
    \midrule
    \multirow{4}[0]{*}{LogicalConcB} & DT   & 100.00 & 100.00 & 100.00 & 100.00 & 100.00 & 100.00 & 100.00 \\
          & NB    & 83.10 & \textbf{94.40} & 83.10 & \textbf{90.75} & \underline{\textbf{96.60}} & \textbf{95.30} & \textbf{95.05} \\
          & FU    & 100.00 & 100.00 & 100.00 & 100.00 & 100.00 & 100.00 & 100.00 \\
          & RF    & 100.00 & 100.00 & 100.00 & 100.00 & 100.00 & 100.00 & 100.00 \\
    \midrule
    \multirow{4}[0]{*}{LogicalConcBNoisy} & DT   & 95.00 & 95.00 & 95.00 & 95.00 & 95.00 & 95.00 & 95.00 \\
          & NB    & 78.25 & \textbf{86.95} & 78.25 & \textbf{85.60} & \underline{\textbf{89.10}} & \textbf{88.70} & \textbf{88.80} \\
          & FU    & 95.00 & 95.00 & 95.00 & 95.00 & 95.00 & 95.00 & 95.00 \\
          & RF    & 95.00 & 95.00 & 95.00 & 95.00 & 95.00 & 95.00 & 95.00 \\
    \midrule
    \multirow{4}[0]{*}{BinClassDisAttr} & DT   & 100.00 & 100.00 & 100.00 & 100.00 & 100.00 & 100.00 & 100.00 \\
          & NB    & 88.60 & 88.60 & 88.60 & 88.60 & \underline{\textbf{95.25}} & 88.60 & 88.60 \\
          & FU    & 100.00 & 100.00 & 100.00 & 100.00 & 100.00 & 100.00 & 100.00 \\
          & RF    & 100.00 & 100.00 & 100.00 & 100.00 & 100.00 & 100.00 & 100.00 \\
    \midrule
    \multirow{4}[0]{*}{BinClassNumBinAttr} & DT   & 99.85 & 99.80 & 99.85 & 99.85 & 99.85 & 99.85 & 99.85 \\
          & NB    & 91.80 & \underline{\textbf{97.45}} & 91.80 & \textbf{94.70} & \textbf{93.90} & \textbf{95.20} & \textbf{97.30} \\
          & FU    & 99.85 & 99.80 & 99.85 & \underline{\textbf{99.90}} & \underline{\textbf{99.90}} & 99.85 & 99.80 \\
          & RF    & 99.85 & 99.80 & 99.85 & 99.80 & 99.80 & 99.85 & \underline{\textbf{99.90}} \\
    \midrule
    \multirow{4}[0]{*}{BinClassNumDisAttr} & DT   & 99.75 & 99.65 & 99.75 & 99.65 & \underline{\textbf{99.85}} & 99.75 & 99.70 \\
          & NB    & 83.95 & \textbf{90.30} & \textbf{85.55} & \textbf{90.95} & \underline{\textbf{97.80}} & \textbf{91.40} & \textbf{91.35} \\
          & FU    & 99.75 & 99.70 & 99.75 & 99.70 & \underline{\textbf{99.85}} & 99.75 & 99.75 \\
          & RF    & 99.75 & 99.75 & 99.70 & 99.75 & \underline{\textbf{99.85}} & 99.75 & 99.65 \\
    \midrule
    \multirow{4}[0]{*}{DisjunctN} & DT   & 100.00 & 100.00 & 100.00 & 100.00 & 100.00 & 100.00 & 100.00 \\
          & NB    & 94.55 & \underline{\textbf{100.00}} & 93.15 & \underline{\textbf{100.00}} & \underline{\textbf{100.00}} & \underline{\textbf{100.00}} & \underline{\textbf{100.00}} \\
          & FU    & 100.00 & 100.00 & 100.00 & 100.00 & 100.00 & 100.00 & 100.00 \\
          & RF    & 100.00 & 100.00 & 99.95 & 100.00 & 100.00 & 100.00 & 100.00 \\
    \midrule
    \multirow{4}[0]{*}{MultiVClassDisAttr} & DT   & 100.00 & 100.00 & 100.00 & 100.00 & 100.00 & 100.00 & 100.00 \\
          & NB    & 90.80 & 86.70 & 90.80 & \underline{\textbf{93.35}} & 89.90 & 86.65 & 90.60 \\
          & FU    & 100.00 & 100.00 & 100.00 & 100.00 & 100.00 & 100.00 & 100.00 \\
          & RF    & 100.00 & 100.00 & 100.00 & 100.00 & 100.00 & 100.00 & 100.00 \\
    \midrule
    \multirow{4}[0]{*}{Concept} & DT   & 100.00 & 100.00 & 100.00 & 100.00 & 100.00 & 100.00 & 100.00 \\
          & NB    & 68.25 & \textbf{86.30} & 68.25 & \textbf{80.45} & \textbf{80.90} & \underline{\textbf{92.35}} & \underline{\textbf{92.35}} \\
          & FU    & 100.00 & 100.00 & 100.00 & 100.00 & 100.00 & 100.00 & 100.00 \\
          & RF    & 100.00 & 100.00 & 100.00 & 100.00 & 100.00 & 100.00 & 100.00 \\
    \midrule
    \multirow{4}[0]{*}{ModGroups} & DT   & 33.85 & 33.85 & \textbf{88.40} & 33.85 & \textbf{88.20} & \textbf{87.55} & \underline{\textbf{88.45}} \\
          & NB    & 42.85 & 42.85 & 39.35 & 42.85 & \textbf{63.05} & \underline{\textbf{64.10}} & \textbf{63.10} \\
          & FU    & 88.70 & 88.70 & \underline{\textbf{88.90}} & 88.70 & 88.25 & 88.35 & 88.50 \\
          & RF    & 88.55 & 88.55 & \textbf{88.75} & 88.55 & \underline{\textbf{88.85}} & \textbf{88.80} & \textbf{88.80} \\
    \midrule
    \multirow{4}[0]{*}{CondInd} & DT   & 89.80 & 89.80 & 89.80 & 89.80 & 89.80 & 89.80 & \underline{\textbf{89.85}} \\
          & NB    & 91.05 & 91.05 & 91.05 & 90.70 & 90.60 & 90.55 & 90.55 \\
          & FU    & 89.85 & 89.80 & 89.85 & \textbf{90.05} & \underline{\textbf{90.35}} & \textbf{90.00} & 89.85 \\
          & RF    & 89.80 & 89.65 & 89.80 & 89.65 & 89.50 & 89.70 & 89.55 \\
    \bottomrule
    \end{tabular}
    }
\end{table}

For NB classifiers, at least one operator setting improves CA in 9 out of 10 datasets, which shows the usefulness of constructed features. For DT, FURIA, and RF, we cannot observe such benefits as the baseline methods are already very successful on tested synthetic problems, often achieving 100\% CA. Unsurprisingly, the results on synthetic data are also inconclusive concerning the  constructive operators as they are all useful in at least some of the datasets.

To summarise the findings on the synthetic data: \emph{EFC effectively identifies informative groups of attributes and ignores irrelevant attributes.}

\subsection{UCI datasets}
\label{sec:resultsUCI}
In this section, we analyse the EFC method on selected UCI datasets using different classifiers and various types of features. We first report predictive performance of various constructive operators, followed by the number of generated features and computational times used by different settings.

\subsubsection{Predictive performance on UCI datasets}
Following the settings used on synthetic data, we tested the classification accuracy of seven popular classifiers: DT, NB, FURIA, RF, k-nearest neighbours (kNN) with k=10, support vector machines with linear kernel (SVM-lin) and support vector machines with RBF kernel (SVM-RBF). In this section, we outline the results of these classifiers and present statistical significance of the differences between the approaches, while the complete tables with the results are available in \ref{sec:additionalTables}. We calculated CA using 10-fold cross-validation. 

Without using constructive induction as a baseline (Base), we compare the use of logical operators (Log), relational operators (Rel), Cartesian product (Cart), decision rule and threshold (DrThr) features, all types of features (All), and all types of features with feature selection (FS). In addition to these methods, we report results of feature construction with the method proposed by \cite{jakulin2005MLattrInteractionsPhD} (Jak) and with the exhaustive search (Exh). For Exh the '-' sign means that the result was not generated in feasible time (3 hours per fold); in the averages, we replace these missing scores with the baselines and mark them with *. We include results of XGBoost (XGB) ensemble as an estimate of the upper bound. The best CA for each dataset (excluding XGBoost) is marked with bold typeface. If none of the feature construction methods improves the baseline result, bold typeface is omitted (see the tables in \ref{sec:additionalTables}).

Results for DT (\Cref{tab:UCIDatasets-ACC-DTfinal}) show that construction with DrThr, Log, and FS settings outperforms the baseline (without feature construction) in 22 out of 30 datasets, the All method outperforms the base results in 20 datasets, and the Rel and Cart method outperforms the base results in 12 out of 30 datasets. In a few cases (8 out of 30), the DT classifier with one of feature construction methods is the best method overall, superior even to XGB. Statistical significance of the differences are reported later in this section.

Using the NB classifier (\Cref{tab:UCIDatasets-ACC-NBfinal}), the FS outperforms the baseline in 21 out of 30 datasets, the methods DrThr and All in 20 datasets, Log and Cart in 19, and Rel in 13 out of 30 datasets. In 4 out of 30 cases the most efficient approach is to use the NB classifier with one of the feature construction methods.

Using the SVM-lin classifier (\Cref{tab:UCIDatasets-ACC-SVMlin}), the DrThr setting outperforms the baseline in 19 out of 30 datasets, the FS setting outperforms the base results in 18 datasets, the All in 17, the Cart in 15, the Log in 14, and the Rel setting in 10 out of 30 datasets. In 11 out of 30 cases the most efficient approach is to use the SVM-lin classifier with one of the feature construction methods.

Using the SVM-RBF classifier (\Cref{tab:UCIDatasets-ACC-SVMRBF}), the FS setting outperforms the baseline in 26 out of 30 datasets, the DrThr method outperforms the base results in 25 datasets, the All method in 22, the Cart and Log in 19, and the Rel in 10 out of 30 datasets. In 7 out of 30 cases the most efficient approach is to use the SVM-RBF classifier with one of the feature construction methods.

Using the kNN classifier (\Cref{tab:UCIDatasets-ACC-kNN}), the DrThr setting outperforms the baseline in 19 out of 30 datasets, the FS method outperforms the base results in 18 datasets, the Cart method in 17, the Log in 15, the All method in 14, and the Rel in 11 out of 30 datasets. In 7 out of 30 cases the most efficient approach is to use the kNN classifier with one of the feature construction methods.

Using the FURIA classifier (\Cref{tab:UCIDatasets-ACC-FU}), FS outperforms the baseline in 15 out of 30 datasets, the All method outperforms the base results in 14 datasets, Log and Rel in 13, DrThr in 11, and Cart in 10 out of 30 datasets. In 10 out of 30 cases the most efficient approach is to use the FURIA classifier with one of the feature construction methods.

Using the RF classifier (\Cref{tab:UCIDatasets-ACC-RF}), the setting DrThr outperforms the baseline in 11 out of 30 datasets, the methods Cart and FS in 10, Log and All in 9, and Rel in 7 out of 30 datasets. In 15 out of 30 cases the most efficient approach is to use the RF classifier with one of the feature construction methods; this is the most of all classifiers tested.

As expected, we get the best average classification accuracy (86.75) with the "upper-bound" XGBoost algorithm. When comparing only the feature construction methods, the best average CA (84.04) for the DT classifier is obtained with all constructed features, and for NB the best score is obtained with the FS setting (79.93). With the FS setting, we also get the best CA for the SVM-RBF (81.00) classifier and the kNN (82.33) classifier. For the SVM-lin classifier, the best score is obtained with the setting All (84.62). For the FURIA classifier, the best CA (86.02) is obtained with Exh setting, and for RF the best score is obtained with the FS setting (86.88) which is even better than with the XGBoost algorithm (86.75). Note that the average CA of baseline for RF is 85.07.

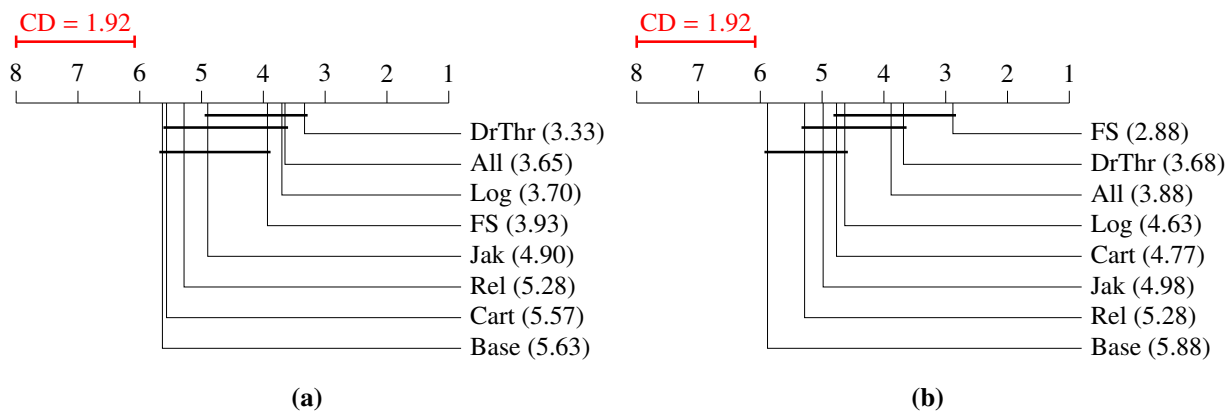
\begin{figure}[h!tb]
\caption{Statistical significance of differences between feature construction methods using the DT (a) and the NB (b) classifier. The critical distance diagram shows the results of the Nemenyi post-hoc test at $\alpha = 0.05$. }
\captionsetup{justification=centering}
\begin{subfigure}[b]{0.49\linewidth}
   \begin{adjustbox}{width=\linewidth}
        \begin{tikzpicture}
            \draw (7-2.333333, 0) -- (7-2.333333 ,-0.500000) -- (7+0.2, -0.500000) node[anchor=west] {DrThr (3.33)};
            \draw (7-2.650000, 0) -- (7-2.650000 ,-1.000000) -- (7+0.2, -1.000000) node[anchor=west] {All (3.65)};
            \draw (7-2.700000, 0) -- (7-2.700000 ,-1.500000) -- (7+0.2, -1.500000) node[anchor=west] {Log (3.70)};
            \draw (7-2.933333, 0) -- (7-2.933333 ,-2.000000) -- (7+0.2, -2.000000) node[anchor=west] {FS (3.93)};
            \draw (7-3.900000, 0) -- (7-3.900000 ,-2.500000) -- (7+0.2, -2.500000) node[anchor=west] {Jak (4.90)};
            \draw (7-4.283333, 0) -- (7-4.283333 ,-3.000000) -- (7+0.2, -3.000000) node[anchor=west] {Rel (5.28)};
            \draw (7-4.566667, 0) -- (7-4.566667 ,-3.500000) -- (7+0.2, -3.500000) node[anchor=west] {Cart (5.57)};
            \draw (7-4.633333, 0) -- (7-4.633333 ,-4.000000) -- (7+0.2, -4.000000) node[anchor=west] {Base (5.63)};
            \draw[very thick, red] (0,0.9) -- (0,1.1);
            \draw[very thick, red] (1.916973, 0.9) -- (1.916973, 1.1);
            \draw[very thick, red] (0,1) -- node[anchor=south] {CD = 1.92} (1.916973,1);
            \draw[very thick] (7 - 3.900000 - 0.05, -0.2*1) -- (7- 2.333333 + 0.05, -0.2*1);
            \draw[very thick] (7 - 4.566667 - 0.05, -0.2*2) -- (7- 2.650000 + 0.05, -0.2*2);
            \draw[very thick] (7 - 4.633333 - 0.05, -0.2*4) -- (7- 2.933333 + 0.05, -0.2*4);
            \draw (0, 0) -- (1, 0) -- (2, 0) -- (3, 0) -- (4, 0) -- (5, 0) -- (6, 0) -- (7, 0);
            \draw (0,0.2) node[anchor=south] {8} -- (0,0);
            \draw (1,0.2) node[anchor=south] {7} -- (1,0);
            \draw (2,0.2) node[anchor=south] {6} -- (2,0);
            \draw (3,0.2) node[anchor=south] {5} -- (3,0);
            \draw (4,0.2) node[anchor=south] {4} -- (4,0);
            \draw (5,0.2) node[anchor=south] {3} -- (5,0);
            \draw (6,0.2) node[anchor=south] {2} -- (6,0);
            \draw (7,0.2) node[anchor=south] {1} -- (7,0);
        \end{tikzpicture}
    \end{adjustbox}
    \caption{} \label{fig-left:CD-DT-plus-Jakulin}  
\end{subfigure}
    \begin{subfigure}[b]{0.49\linewidth}
        \begin{adjustbox}{width=\linewidth}
            \begin{tikzpicture}
            \draw (7-1.883333, 0) -- (7-1.883333 ,-0.500000) -- (7+0.2, -0.500000) node[anchor=west] {FS (2.88)};
            \draw (7-2.683333, 0) -- (7-2.683333 ,-1.000000) -- (7+0.2, -1.000000) node[anchor=west] {DrThr (3.68)};
            \draw (7-2.883333, 0) -- (7-2.883333 ,-1.500000) -- (7+0.2, -1.500000) node[anchor=west] {All (3.88)};
            \draw (7-3.633333, 0) -- (7-3.633333 ,-2.000000) -- (7+0.2, -2.000000) node[anchor=west] {Log (4.63)};
            \draw (7-3.766667, 0) -- (7-3.766667 ,-2.500000) -- (7+0.2, -2.500000) node[anchor=west] {Cart (4.77)};
            \draw (7-3.983333, 0) -- (7-3.983333 ,-3.000000) -- (7+0.2, -3.000000) node[anchor=west] {Jak (4.98)};
            \draw (7-4.283333, 0) -- (7-4.283333 ,-3.500000) -- (7+0.2, -3.500000) node[anchor=west] {Rel (5.28)};
            \draw (7-4.883333, 0) -- (7-4.883333 ,-4.000000) -- (7+0.2, -4.000000) node[anchor=west] {Base (5.88)};
            \draw[very thick, red] (0,0.9) -- (0,1.1);
            \draw[very thick, red] (1.916973, 0.9) -- (1.916973, 1.1);
            \draw[very thick, red] (0,1) -- node[anchor=south] {CD = 1.92} (1.916973,1);
            \draw[very thick] (7 - 3.766667 - 0.05, -0.2*1) -- (7- 1.883333 + 0.05, -0.2*1);
            \draw[very thick] (7 - 4.283333 - 0.05, -0.2*2) -- (7- 2.683333 + 0.05, -0.2*2);
            \draw[very thick] (7 - 4.883333 - 0.05, -0.2*4) -- (7- 3.633333 + 0.05, -0.2*4);
            \draw (0, 0) -- (1, 0) -- (2, 0) -- (3, 0) -- (4, 0) -- (5, 0) -- (6, 0) -- (7, 0);
            \draw (0,0.2) node[anchor=south] {8} -- (0,0);
            \draw (1,0.2) node[anchor=south] {7} -- (1,0);
            \draw (2,0.2) node[anchor=south] {6} -- (2,0);
            \draw (3,0.2) node[anchor=south] {5} -- (3,0);
            \draw (4,0.2) node[anchor=south] {4} -- (4,0);
            \draw (5,0.2) node[anchor=south] {3} -- (5,0);
            \draw (6,0.2) node[anchor=south] {2} -- (6,0);
            \draw (7,0.2) node[anchor=south] {1} -- (7,0);
            \end{tikzpicture}
        \end{adjustbox}
    \caption{} \label{fig-right:CD-NB-plus-Jakulin}  
    \end{subfigure}
\label{fig:CD-DT-NB-plus-Jakulin} 
\end{figure}

\begin{figure}[h!tb]  
\caption{Statistical significance of differences between feature construction methods using the SVM with the RBF kernel (a) and the kNN (b) classifier. The critical distance diagram shows the results of the Nemenyi post-hoc test at $\alpha = 0.05$.}
\captionsetup{justification=centering}
\begin{subfigure}[b]{0.49\linewidth}
   \begin{adjustbox}{width=\linewidth}
        \begin{tikzpicture}
		\draw (7-2.216667, 0) -- (7-2.216667 ,-0.500000) -- (7+0.2, -0.500000) node[anchor=west] {FS (3.22)};
		\draw (7-2.400000, 0) -- (7-2.400000 ,-1.000000) -- (7+0.2, -1.000000) node[anchor=west] {DrThr (3.40)};
		\draw (7-2.783333, 0) -- (7-2.783333 ,-1.500000) -- (7+0.2, -1.500000) node[anchor=west] {All (3.78)};
		\draw (7-3.416667, 0) -- (7-3.416667 ,-2.000000) -- (7+0.2, -2.000000) node[anchor=west] {Jak (4.42)};
		\draw (7-3.516667, 0) -- (7-3.516667 ,-2.500000) -- (7+0.2, -2.500000) node[anchor=west] {Cart (4.52)};
		\draw (7-3.616667, 0) -- (7-3.616667 ,-3.000000) -- (7+0.2, -3.000000) node[anchor=west] {Log (4.62)};
		\draw (7-4.833333, 0) -- (7-4.833333 ,-3.500000) -- (7+0.2, -3.500000) node[anchor=west] {Rel (5.83)};
		\draw (7-5.216667, 0) -- (7-5.216667 ,-4.000000) -- (7+0.2, -4.000000) node[anchor=west] {Base (6.22)};
		\draw[very thick, red] (0,0.9) -- (0,1.1);
		\draw[very thick, red] (1.916973, 0.9) -- (1.916973, 1.1);
		\draw[very thick, red] (0,1) -- node[anchor=south] {CD = 1.92} (1.916973,1);
		\draw[very thick] (7 - 3.616667 - 0.05, -0.2*1) -- (7- 2.216667 + 0.05, -0.2*1);
		\draw[very thick] (7 - 5.216667 - 0.05, -0.2*4) -- (7- 3.416667 + 0.05, -0.2*4);
		\draw (0, 0) -- (1, 0) -- (2, 0) -- (3, 0) -- (4, 0) -- (5, 0) -- (6, 0) -- (7, 0);
		\draw (0,0.2) node[anchor=south] {8} -- (0,0);
		\draw (1,0.2) node[anchor=south] {7} -- (1,0);
		\draw (2,0.2) node[anchor=south] {6} -- (2,0);
		\draw (3,0.2) node[anchor=south] {5} -- (3,0);
		\draw (4,0.2) node[anchor=south] {4} -- (4,0);
		\draw (5,0.2) node[anchor=south] {3} -- (5,0);
		\draw (6,0.2) node[anchor=south] {2} -- (6,0);
		\draw (7,0.2) node[anchor=south] {1} -- (7,0);
	    \end{tikzpicture}
    \end{adjustbox}
    \caption{} \label{fig-left:CD-SVMRBF}  
\end{subfigure}
    \begin{subfigure}[b]{0.49\linewidth}
        \begin{adjustbox}{width=\linewidth}
		\begin{tikzpicture}
		\draw (7-1.983333, 0) -- (7-1.983333 ,-0.500000) -- (7+0.2, -0.500000) node[anchor=west] {FS (2.98)};
		\draw (7-2.400000, 0) -- (7-2.400000 ,-1.000000) -- (7+0.2, -1.000000) node[anchor=west] {DrThr (3.40)};
		\draw (7-3.416667, 0) -- (7-3.416667 ,-1.500000) -- (7+0.2, -1.500000) node[anchor=west] {All (4.42)};
		\draw (7-3.433333, 0) -- (7-3.433333 ,-2.000000) -- (7+0.2, -2.000000) node[anchor=west] {Cart (4.43)};
		\draw (7-3.916667, 0) -- (7-3.916667 ,-2.500000) -- (7+0.2, -2.500000) node[anchor=west] {Base (4.92)};
		\draw (7-4.116667, 0) -- (7-4.116667 ,-3.000000) -- (7+0.2, -3.000000) node[anchor=west] {Rel (5.12)};
		\draw (7-4.316667, 0) -- (7-4.316667 ,-3.500000) -- (7+0.2, -3.500000) node[anchor=west] {Log (5.32)};
		\draw (7-4.416667, 0) -- (7-4.416667 ,-4.000000) -- (7+0.2, -4.000000) node[anchor=west] {Jak (5.42)};
		\draw[very thick, red] (0,0.9) -- (0,1.1);
		\draw[very thick, red] (1.916973, 0.9) -- (1.916973, 1.1);
		\draw[very thick, red] (0,1) -- node[anchor=south] {CD = 1.92} (1.916973,1);
		\draw[very thick] (7 - 3.433333 - 0.05, -0.2*1) -- (7- 1.983333 + 0.05, -0.2*1);
		\draw[very thick] (7 - 4.316667 - 0.05, -0.2*2) -- (7- 2.400000 + 0.05, -0.2*2);
		\draw[very thick] (7 - 4.416667 - 0.05, -0.2*3) -- (7- 3.416667 + 0.05, -0.2*3);
		\draw (0, 0) -- (1, 0) -- (2, 0) -- (3, 0) -- (4, 0) -- (5, 0) -- (6, 0) -- (7, 0);
		\draw (0,0.2) node[anchor=south] {8} -- (0,0);
		\draw (1,0.2) node[anchor=south] {7} -- (1,0);
		\draw (2,0.2) node[anchor=south] {6} -- (2,0);
		\draw (3,0.2) node[anchor=south] {5} -- (3,0);
		\draw (4,0.2) node[anchor=south] {4} -- (4,0);
		\draw (5,0.2) node[anchor=south] {3} -- (5,0);
		\draw (6,0.2) node[anchor=south] {2} -- (6,0);
		\draw (7,0.2) node[anchor=south] {1} -- (7,0);
		\end{tikzpicture}
        \end{adjustbox}
    \caption{} \label{fig-right:CD-kNN}  
    \end{subfigure}
\label{fig:CD-SVMRBF-kNN} 
\end{figure}
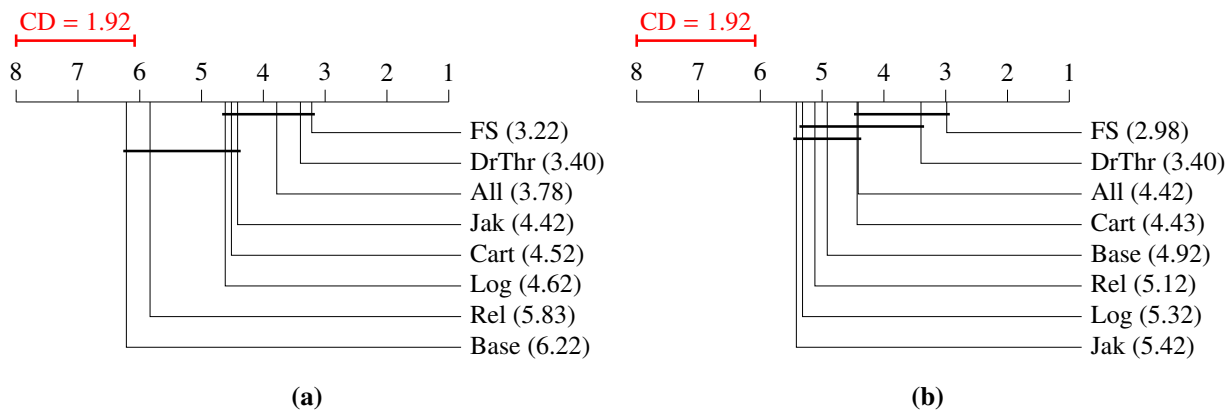

To assess the statistical significance of the differences between the tested feature construction methods, we perform a non-parametric Friedman test with the Nemenyi correction \citep{demvsar2006statistical} for all classifiers (DT, NB, FURIA, RF, kNN, SVM-lin, SVM-RBF). \Cref{fig:CD-DT-NB-plus-Jakulin} shows the critical difference (CD) diagrams for CA of DT and NB classifiers and \Cref{fig:CD-SVMRBF-kNN} for the CA of the SVM-RBF and kNN classifiers. The average rank of each method is marked on the horizontal axis (lower ranks mean better CA). Groups of methods that are not significantly different according to the Nemenyi test are connected with a thick line. The length of the critical difference (e.g., CD = 1.92 for the DT classifier) is shown above the axis.

Using the DT classifier, operators DrThr, All, and Log significantly improve CA compared to the baseline. The DrThr setting is superior also to constructs composed of only Rel and Cart features (see Figure \ref{fig-left:CD-DT-plus-Jakulin}). The NB classifier returns similar results, the difference is in the method with the lowest rank (see Figure \ref{fig-right:CD-NB-plus-Jakulin}), which is FS for NB. The FS setting is superior also to constructs composed of only Rel features. The SVM-RBF classifier returns similar  results as NB. The FS, DrThr, and All settings improve CA compared to the baseline (see Figure \ref{fig-left:CD-SVMRBF}); all these settings are superior also to constructs composed of only Rel features. Using the kNN classifier, the FS operator significantly improves CA compared to the baseline and is superior also to constructs composed of only Rel and Log features (see Figure \ref{fig-right:CD-kNN}). The results of the RF, FURIA, and SVM-lin classifiers show no statistical differences between the construction methods, so we skip the graphs.

We can conclude that the proposed feature construction is beneficial. Using the resulting constructs, the interpretable models, such as DT and NB as well as the kNN and the SVM models become significantly better compared to baselines without feature construction. Unsurprisingly, different datasets and learning methods require different feature construction operators. As there is no clear winner among the tested operators, we decided to use all operators (the All setting) in further analyses.

\subsubsection{Analysis of constructed features for UCI datasets}
\label{sec:numConstrFeat}
In this section, we analyse the properties of the obtained constructs. We start with the identified groups of co-occurring attributes, followed by the constructed features and features actually used in DT classifiers. We work with all types of features (the All setting).

\begin{figure}[!ht]
    \centering
    \caption{The summary of EFC's ability to reduce the search space shown for used UCI datasets. From left to right, the box-and-whiskers plots show the distribution of the number of groups, number of attributes in groups, and maximum length of groups. All numbers are small across all tested UCI datasets.}
    \includegraphics[keepaspectratio,width=0.8 \linewidth]{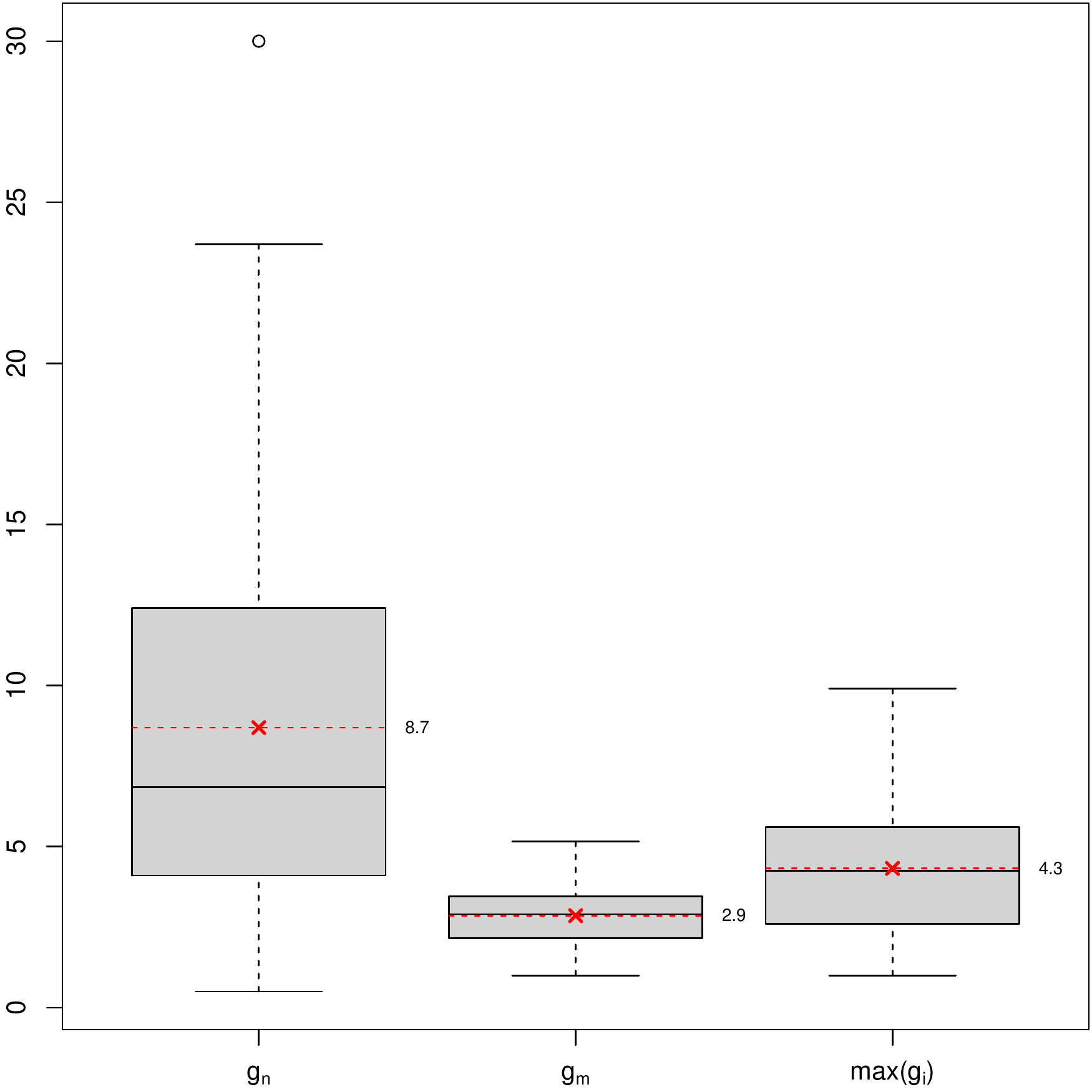}
    \label{fig:numOfGeneratedGroups}
\end{figure}

\Cref{fig:numOfGeneratedGroups} shows the inner workings of the EFC method, namely the distribution of the number of groups, number of attributes in groups and maximal length of groups, obtained on the UCI datasets with the DT classifier.

As the box-and-whiskers plots show, EFC generates an average of 8.7 groups of co-occurring attributes, which have the average size of 2.9 (number of attributes in the group), with the maximum size of 4.3 attributes per group. These numbers are relatively low and so is their variance. Comparison of the number and sizes of discovered groups with the overall number of attributes in \Cref{tab:UCIDatasets-summary} shows that EFC effectively reduces the search space. This finding is confirmed in \Cref{sec:time}, where we measure the running times of EFC.

\Cref{tab:numOfFeaturesV1} shows the average number  of the constructed features and features actually used in the DT classifier (using All setting). The number of generated features is proportional to the number of operators so EFC generates the largest number of logical features (5 operators), followed by decision rule features, threshold features, Cartesian product features, and  relational features.

The number of features actually used in the DT classifiers is moderately correlated with the number of generated features (Spearman's rank correlation coefficient $\rho = 0.487$ and $p = 0.006$)  so logical features are the most often used. An exception is in the relational features, which are used more often, and Cartesian product features, which are almost never used in All setting, where other types of features are also available.

\begin{table}[!htb]
  \centering
  \caption{Average numbers of constructed and actually used features for the All method and DT classifier. Columns show logical (Log), relational (Rel), Cartesian product (Cart), decision rule (Dr), and threshold features (Thr). } 
    \begin{tabular}{lrrrrr|rrrrr}
    \hline
    & \multicolumn{5}{c|}{constructed}       
    & \multicolumn{5}{c}{actually used} \\
    \textbf{Dataset}  
    & \multicolumn{1}{c}{\textbf{Log}} 
    & \multicolumn{1}{c}{\textbf{Rel}} 
    & \multicolumn{1}{c}{\textbf{Cart}} 
    & \multicolumn{1}{c}{\textbf{Dr}} 
    & \multicolumn{1}{c|}{\textbf{Thr}} 
    & \multicolumn{1}{c}{\textbf{Log}} 
    & \multicolumn{1}{c}{\textbf{Rel}} 
    & \multicolumn{1}{c}{\textbf{Cart}} 
    & \multicolumn{1}{c}{\textbf{Dr}} 
    & \multicolumn{1}{c}{\textbf{Thr}} \\
    \hline
    Aus. credit & 1191.4 & 13.3  & 18.8  & 10.2  & 9.0   & 11.8  & 1.8   & 0     & 2     & 0.2 \\
    Autos & 53.8  & 0.5   & 0.5   & 0.5   & 0.5   & 2.3   & 0     & 0     & 0.5   & 0.1 \\
    Balance scale & 46.4  & 12.0  & 6.0   & 35.0  & 32.9  & 2.4   & 12.5  & 0     & 10.7  & 2.8 \\
    Bank mark. & 197.0 & 0.2   & 1.1   & 1.2   & 0.4   & 13.6  & 0.2   & 0     & 1     & 0 \\
    Bio. resp. & 45.8  & 8.9   & 4.7   & 4.3   & 3.2   & 8.9   & 1.2   & 0.2   & 2.3   & 0.6 \\
    Car   & 1095.5 & 0.0   & 9.0   & 48.4  & 48.4  & 27.1  & 0     & 0     & 0     & 0 \\
    CNAE-9 & 12.9  & 4.6   & 2.3   & 2.4   & 0.0   & 0.9   & 0     & 0     & 0     & 0 \\
    Glass & 299.4 & 9.3   & 10.5  & 2.4   & 1.4   & 4.8   & 1.2   & 0.1   & 0.8   & 1.4 \\
    Heart & 145.1 & 20.7  & 20.7  & 17.1  & 13.5  & 3.8   & 0.7   & 0.1   & 3.6   & 0.2 \\
    Ionosphere & 1816.5 & 20.0  & 10.0  & 10.0  & 4.4   & 4.6   & 0.3   & 0     & 1.3   & 0 \\
    Jap. credit & 595.8 & 15.2  & 19.9  & 10.8  & 9.7   & 9.9   & 0.7   & 0     & 3     & 0.3 \\
    Leukemia & 70.8  & 12.3  & 10.2  & 2.6   & 2.3   & 0.7   & 0     & 0     & 0.3   & 0 \\
    Liver dis. & 17.9  & 16.0  & 3.9   & 17.4  & 15.0  & 0.3   & 2.5   & 0     & 9.6   & 2.1 \\
    LSVT voice & 354.7 & 14.0  & 26.4  & 8.3   & 7.4   & 1.3   & 0     & 0     & 2.4   & 0 \\
    Lung cancer & 89.6  & 0.0   & 1.5   & 1.1   & 0.9   & 1.7   & 0     & 0     & 0     & 0 \\
    Mamm. mass & 256.1 & 6.9   & 6.0   & 8.8   & 4.8   & 4.9   & 0.6   & 0.1   & 2.2   & 0.4 \\
    Molecular & 6612.3 & 0.0   & 25.7  & 27.4  & 25.8  & 3.8   & 0     & 0     & 0     & 0 \\
    Monks1 & 136.5 & 0.0   & 1.8   & 11.7  & 9.6   & 2.6   & 0     & 0     & 0.1   & 0 \\
    Monks2 & 866.1 & 0.0   & 14.3  & 0.8   & 0.8   & 38.2  & 0     & 0     & 0     & 0 \\
    Nursery & 1471.0 & 0.0   & 9.4   & 33.2  & 32.2  & 138.1 & 0     & 0     & 0.1   & 0 \\
    Parkinsons & 436.2 & 1.4   & 17.8  & 18.7  & 16.9  & 1     & 0     & 0     & 3.3   & 0 \\
    QSAR & 308.6 & 11.3  & 9.1   & 15.4  & 12.3  & 8.4   & 2.5   & 0.1   & 7     & 0.8 \\
    Semeion & 489.8 & 0.0   & 45.8  & 11.1  & 10.9  & 10.6  & 0     & 0.4   & 1.5   & 0.1 \\
    Sonar & 45.6  & 7.9   & 7.2   & 6.8   & 3.4   & 1.9   & 0.1   & 0.1   & 2.7   & 0.2 \\
    Spambase & 335.0 & 10.8  & 6.9   & 12.9  & 5.0   & 22.3  & 0.7   & 0.1   & 4.5   & 0.5 \\
    Spect heart & 225.1 & 0.0   & 25.1  & 4.0   & 4.0   & 0.4   & 0     & 0     & 0.9   & 0 \\
    Spectf heart & 26.5  & 4.0   & 2.0   & 2.8   & 2.6   & 1.7   & 0.7   & 0     & 1.9   & 0.2 \\
    Tic-tac-toe & 3180.4 & 0.0   & 27.4  & 83.5  & 82.5  & 12.9  & 0     & 0     & 1.4   & 0 \\
    Vehicle & 5923.1 & 6.9   & 19.7  & 43.4  & 41.9  & 25    & 2.3   & 0.1   & 8.5   & 10.3 \\
    Voting & 64.7  & 0.0   & 9.5   & 3.6   & 2.6   & 1.6   & 0     & 0     & 0     & 0 \\
    \hline
    \textbf{Average} & \textbf{880.3} & \textbf{6.5} & \textbf{12.4} & \textbf{15.2} & \textbf{13.5} & \textbf{12.25} & \textbf{0.93} & \textbf{0.04} & \textbf{2.39} & \textbf{0.67} \\
    \hline
    \end{tabular}
  \label{tab:numOfFeaturesV1}
\end{table}

We similarly analysed the learned decision rules and counted the number of rules in the ruleset and the number of attributes in each rule. Comparing rules produced on the originally selected UCI datasets and rules produced on enriched datasets, the average number of rules dropped from 29.33 to 20.02. However, the complexity of rules increased and the sum of attributes used in all rules showed almost the same complexity, i.e., 116.05 for rules on the original datasets and 114.10 for enriched dataset. The details are given in Appendix B, see \Cref{tab:featComplexity-DR}.

We can conclude that all types of features are useful in tested ML models but their use depends on the ML algorithm and dataset. While the size of the models is reduced  with introduction of constructed features, the overall complexity remains approximately the same.

\subsubsection{Running times on UCI datasets}
\label{sec:time}
Since feature construction typically operates in  very large search spaces, an important practical consideration is its running time. We compare running times of the EFC method in All setting (constructing all types of features) with the exhaustive method, which does not use the proposed search space reduction but is otherwise completely identical to EFC.

\begin{figure}[!ht]
    \centering
    \caption{Running times (in seconds) of the feature construction part of the EFC methodology compared to the exhaustive search on the UCI datasets. Times of exhaustive search which exceed 3 hours per fold are dotted. EFC times for many datasets are  below 1 sec and are not visible on the graph.}
    \includegraphics[keepaspectratio,width=\linewidth,height=\dimexpr\textheight-3\baselineskip]{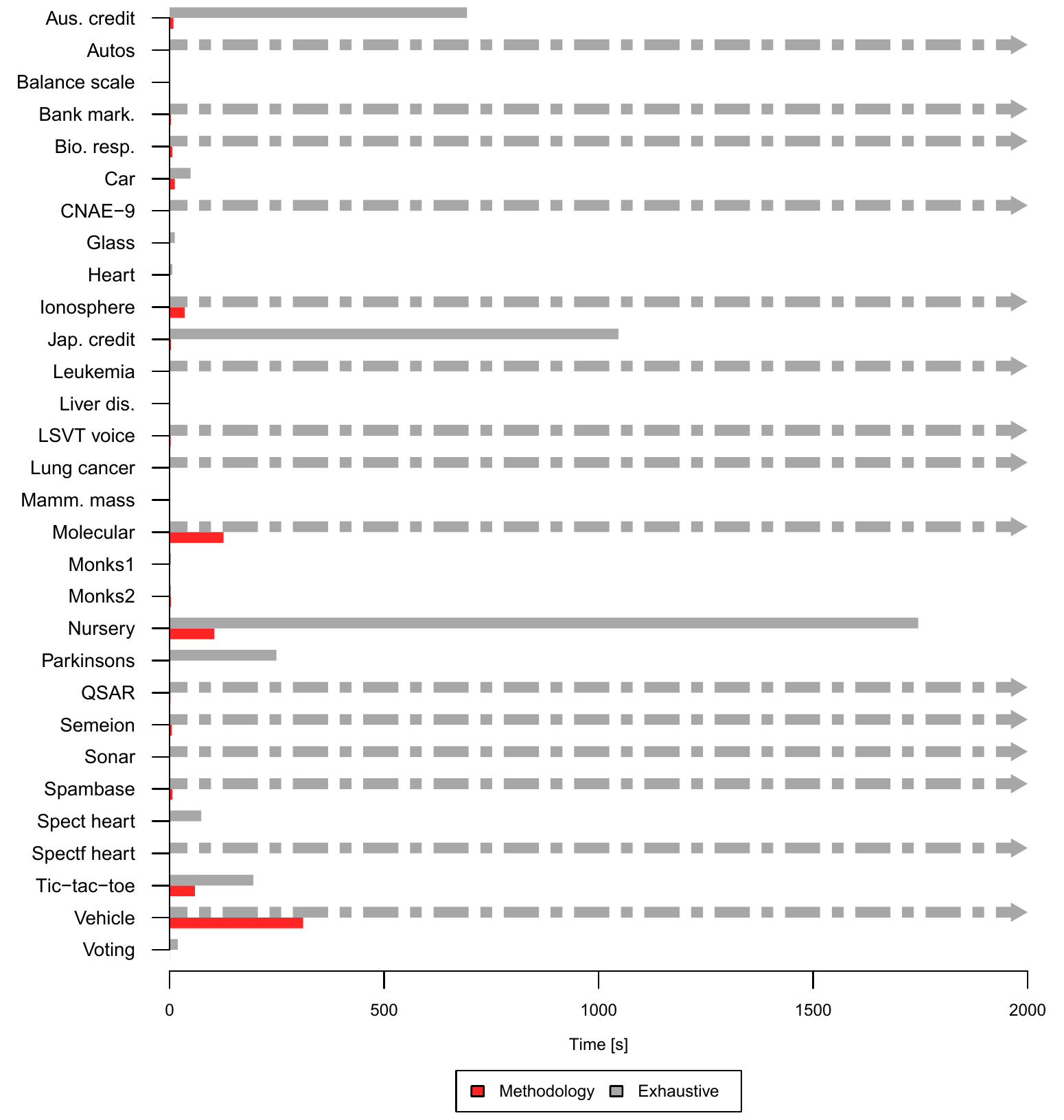}
    \label{fig:computTimeV1}
\end{figure}

\Cref{fig:computTimeV1} illustrates that running times of EFC are significantly lower than the times of the  exhaustive method. The exhaustive method was unable to terminate in the maximal allowed time (3 hours per one cross-validation fold) in half of the datasets (indicated with the dashed line with an arrow).  The detailed running times and their standard deviations are presented in Appendix C (see \Cref{tab:computTime}).

To summarise, \emph{EFC significantly improves the CA of DT, NB, SVM-RBF and kNN classifiers on UCI datasets, effectively reducing the search space and running times compared to the exhaustive approach and enabling the use of feature construction even for large datasets.}

\subsection{Use case in finance}
\label{sec:resultsFinance}
In this section, we present an analysis of a use case from the financial industry, namely the credit risk assessment. The use case demonstrates a practical utility of the EFC method to generate human-comprehensible features. The aim is not to produce the highest possible prediction accuracy but to help human expert to better understand the risks associated with different types of customers. We first use EFC to construct informative and practically relevant features from the baseline financial indicators. The constructed features are compared with known expert-designed (ED) features.

To construct informative features for credit risk assessment from raw financial data, we explain both class values (the ''good'' and the ''bad'' credit score). Besides the constructive operators presented previously, we also include numerical operators $O_{n}=\{+,-,/\}$ which frequently appear in the financial domain (thus, demonstrating the ease of adapting the EFC system to new domains). We set the noise threshold  $noiseThr=0$ to extract as much information as possible.

We validate the obtained features in two ways: i) by expert assessment of the constructed features in  \Cref{sec:experiment}, and ii) by assessment of their predictive performance in \Cref{sec:financialPerf}. We compare the predictive performance of models using several combinations of only baseline financial indicators, existing expert features, and constructed features. 

\subsubsection{Utility of constructed credit scoring features}
\label{sec:experiment}
In first evaluation of constructed features, we engaged a financial expert (professor of finance) to assess the features. In \Cref{tab:featuresInExperiment}, we compare the expert's numerical scores with the MDL scores. We show the strongest 25 constructed features (by the MDL score) obtained with EFC from the attributes in \Cref{tab:creditScoreAttrMDL} that use decision rules (13 rules) and quotients of financial indicators (12 rules). Since the threshold values in the rules do not have much meaning for the expert (they strongly depend on the examples in the dataset and not on general knowledge), we simplified the operators and values as follows: the decision rules with operator $<=$ was changed to "= low", and with operator $>=$ to "= high". The expert scored the features on a 4-point scale according to their impact on the credit scoring (0 - pointless, 1 - weak, 2 - strong, 3 - very strong). As a control group, we added 10 constructs that use sums, differences, or relations between attributes (these features are expected to be less important in this domain). Pearson’s statistical test showed that there is a significant linear correlation between the rankings of the MDL scores and expert scores ($\rho = 0.427$, $p=0.011$).

\begin{table}[!ht]
  \centering
    \centering
      \caption{Features (rules and ratios) constructed with EFC for the credit scoring use case. The column Exp shows the expert's opinion of the feature (0 - pointless, 1 - weak, 2 - strong, 3 - very strong). Column MDL shows the MDL score, and column EDf marks if the ED features are included (letter $y$) or excluded from EFC (letter $n$). Features from the control group are marked with an asterisk ($^\ast$).}
  \resizebox*{\linewidth}{!}{
    \begin{tabular}{l|c|c|c|}
    \multicolumn{1}{c|}{\textbf{EFC Feature}} & \multicolumn{1}{c|}{\textbf{Exp}} & \multicolumn{1}{c|}{\textbf{MDL}} & \multicolumn{1}{c|}{\textbf{EDf}} \\
    \hline
    ((Net Debt / EBITDA) = low) and (Equity Ratio = high) & 2     & 0.633 & y \\
    ((Net Debt / EBITDA) = low) and ((Total Operating Liabilities / Assets) = low) & 2     & 0.608 & y \\
    (TIE = high) and (Equity Ratio = high) and ((Net Debt / EBITDA) = low) & 3     & 0.575 & y \\
    ((Net Debt / EBITDA) = low) and ((Total Operating Liabilities / Assets) = low) and    & 3     & 0.568 & y \\
    \ \ \ \ \ \   (ROA = high) and (Current Ratio = high) & & & \\
    $^\ast$(Net Debt / EBITDA) $-$ Current Ratio & 0     & 0.553 & y \\
    $^\ast$(Net Debt / EBITDA) $-$ Equity Ratio & 1     & 0.548 & y \\
    ((Net Debt / EBITDA) = low) and (Current Ratio = high) and (TIE = high) and & 2     & 0.545 & y \\
    \ \ \ \ \ \ ((Total Operating Liabilities / Assets) = low) & & & \\
    $^\ast$(Net Debt / EBITDA) $+$ Equity Ratio & 0     & 0.540 & y \\
    $^\ast$TIE $-$ (Net Debt / EBITDA) & 0     & 0.535 & y \\
    ((Net Debt / EBITDA) = low) and (Current Ratio = high) and (FFO = high) & 3     & 0.529 & y \\
    (TIE = high) and (Equity Ratio = high) and ((Net Debt / EBITDA) = low) & 3     & 0.529 & y \\
    ((Net Debt / EBITDA) = low)) and (Current Ratio = high) and (ROA = high) & 2     & 0.529 & y \\
    ((Net Debt / EBITDA) = low) and (Current Ratio = high) and  & 3     & 0.524 & y \\
    \ \ \ \ \ \ (ROA = high) and (Depreciation = high) & & & \\
    EBITDA / Debt & 2     & 0.510 & n \\
    Net Income / Debt & 2     & 0.482 & n \\
    EBIT / Debt & 2     & 0.443 & n \\
    $^\ast$Interest $-$ Net Income & 1     & 0.405 & n \\
    FFO / Debt & 2     & 0.403 & n \\
    EBIT / Interest & 3     & 0.402 & n \\
    $^\ast$Financial Expenses $-$ Net Income & 1     & 0.394 & n \\
    Net Income / Interest & 2     & 0.391 & n \\
    Net Income / Financial Expenses & 2     & 0.382 & n \\
    Net Income / Short-Term Liabilities & 2     & 0.382 & n \\
    EBIT / Financial Expenses & 3     & 0.373 & n \\
    EBITDA / Interest & 3     & 0.370 & n \\
    (Net Income = high) and (Interest = low) and (Depreciation = high) & 3     & 0.366 & n \\
    Net Income / Short-Term Operating Liabilities & 2     & 0.365 & n \\
    (Net Income = high) and (Interest = low) & 3     & 0.364 & n \\
    (Net Income = high) and (Interest = low) and (Long-Term Assets = high) & 3     & 0.358 & n \\
    Net Income / Short-Term Assets & 1     & 0.346 & n \\
    (Net Income = high) and (Short-Term Liabilities = low) & 3     & 0.344 & n \\
    $^\ast$ROA $-$ Inventories & 0     & 0.004 & y \\
    $^\ast$Equity $+$ Debt & 0     & 0.002 & n \\
    $^\ast$Short-Term Assets $-$ EBIT & 0     & 0.002 & n \\
    $^\ast$Current Ratio $<$ Equity Ratio & 0     & 0.000 & y \\
    \hline
    \end{tabular}
    }
  \label{tab:featuresInExperiment}
\end{table}

On average the expert rated the rules and quotients significantly higher compared to the features in the control group. All conjunctive rules were rated by the expert as strongly predictive, in 9 out of 13 cases even as very strong. All quotients were marked as meaningful and for 11 of 12 quotients the expert indicated that they have a strong predictive power for the assessment of creditworthiness. On average, the obtained rules have higher predictive power than the quotients. For 9 of the 12 quotients, we found explanations in the financial literature \citep{calderon2017-EbitToFinancialExpenses, dang2019-NetIncomeToStAssets, gebauer2018-EbitdaToDebt, hayes2021-EbitdaToInterest, irfan2014-NetIncomeToDebt, jiangli2004-NetIncomeToInterestExpenses, kenton2020-FfoToDebt, koller2010-EbitToNetDebt, lee2016-EbitToInterest}, and for the remaining three quotients, we have obtained very good approximations to the standard financial ratios explained in the financial literature. We conclude that EFC constructs useful features. 

Expert-designed (ED) features show a strong relationship with credit score prediction, as indicated by their high MDL scores (see \Cref{tab:creditScoreAttrMDL}). The EFC system is able to generate identical features from the baseline attributes only (see features with $n$ in the EDf column, e.g., "EBIT / Interest" is actually the same as the ED feature TIE). If we include the ED features into the construction, EFC produces some complex features with MDL scores even higher than those of the existing ED features. Observing features with $y$ in the EDf column of \Cref{tab:featuresInExperiment}, we can see that these complex features are not difficult to comprehend when expressed as conjunctive rules. One of the common constructs in these rules is related to net debt divided by EBITDA, which indicates the number of years it takes a company to pay off its debt given its current cash flow.

However, EFC may produce complex features with a very high MDL score, which are more difficult to comprehend. An example of such a feature is "(EBIT = low) or (Net Debt / EBITDA = high) or (Equity Ratio = low)", with the MDL score of 0.684. Note that the operator OR is usually more difficult to interpret even for domain experts, so we did not use it this experiment.

\subsubsection{EFC for prediction in credit scoring} \label{sec:financialPerf}
In \Cref{tab:creditScore-ACC}, we present the classification accuracy of different models in the credit scoring dataset. We compare predictive performance of models using a) only baseline financial indicators, b) baseline indicators and existing expert features, c) baseline indicators and constructed features, and d) baseline indicators, constructed features, and expert-designed features.

CA of the models was calculated using 10-fold cross-validation. The bold value in the tables means that it is greater than the baseline result for the specific classifier. Due to the flexibility of EFC and specifics of the domain, we decided to include numeric features in the All setting. We use the same EFC settings as described in \Cref{sec:expSettings}, except for $noiseThr$ which is set to 0.

If we exclude the expert features (\Cref{tab:creditScore-ACC}a) and use the DT, NB, SVM-lin, SVM-RBF, kNN, or FU classifier, EFC improves CA over the baseline. If we use much stronger but incomprehensible RF, there is no improvement.
If we also include ED features in the evaluation, the classifiers NB, SVM-lin, SVM-RBF, kNN, FU and RF improve CA, while DT does not.

\begin{table}[h!tbp]
  \centering
  \caption{Classification accuracy on the credit scoring use case when expert designed (ED) features are excluded (a) and included (b).}
  \label{tab:creditScore-ACC}
    \begin{tabular}{l|ll|ll}
          & \multicolumn{2}{c|}{a} & \multicolumn{2}{c}{b} \\
    \textbf{Algorithm} & \multicolumn{1}{c}{\textbf{Base}} & \multicolumn{1}{c|}{\textbf{EFC}} & \multicolumn{1}{c}{\textbf{Base}} & \multicolumn{1}{c}{\textbf{EFC}} \\
    \midrule
    DT      & 83.38 \footnotesize{(\textpm12.48)} & \textbf{90.10 \footnotesize{(\textpm4.90)}} & 90.51 \footnotesize{(\textpm6.59)} & 88.30 \footnotesize{(\textpm6.19)} \\
    NB      & 50.63 \footnotesize{(\textpm6.48)}  & \textbf{82.43 \footnotesize{(\textpm8.01)}} & 68.99 \footnotesize{(\textpm6.15)} & \textbf{90.57 \footnotesize{(\textpm6.23)}} \\
    SVM-lin & 54.68 \footnotesize{(\textpm3.91)}  & \textbf{84.68 \footnotesize{(\textpm11.13)}} & 87.43 \footnotesize{(\textpm9.24)} & \textbf{92.35 \footnotesize{(\textpm6.40)}} \\
    SVM-RBF & 53.83 \footnotesize{(\textpm6.83)}  & \textbf{79.33 \footnotesize{(\textpm9.70)}} & 52.00 \footnotesize{(\textpm2.31)} & \textbf{86,88 \footnotesize{(\textpm10.31)}} \\
    kNN     & 81.60 \footnotesize{(\textpm7.92)}  & \textbf{82.49 \footnotesize{(\textpm8.44)}} & 84.74 \footnotesize{(\textpm6.50)} & \textbf{91.88 \footnotesize{(\textpm6.07)}}\\
    FU      & 81.94 \footnotesize{(\textpm13.06)} & \textbf{91.03 \footnotesize{(\textpm4.07)}} & 88.77 \footnotesize{(\textpm6.17)} & \textbf{91.42 \footnotesize{(\textpm5.18)}} \\
    RF      & 85.61 \footnotesize{(\textpm8.10)}  & 85.57 \footnotesize{(\textpm10.34)}         & 90.55 \footnotesize{(\textpm6.90)} & \textbf{91.90 \footnotesize{(\textpm6.05)}} \\
    \end{tabular}
\end{table}

To summarise, the findings in the financial use case show that \emph{EFC is able to construct useful human comprehensible features and significantly improves the CA of several classifiers.}

\section{Conclusion}
\label{sec:conclusions}
We presented a novel heuristic for reduction of search space in feature construction. The proposed EFC method is based on explanations of predictive models and reduces the search to groups of frequently co-occurring attributes in explained instances.  The method constructs meaningful, explainable features which improve the interpretability of ML models and their prediction performance. By limiting feature construction to these groups, we significantly reduce the search space and allow feature construction for many large datasets. The use case from the financial industry shows that EFC generates many features similar or equivalent to the ones constructed by human experts. Moreover, the domain expert assessed all the generated features (rules and quotients) from the use-case domain as meaningful. Using EFC with inherently interpretable ML models such as DT and NB improved their performance, often to the level of complex black-box models such as XGBoost and random forests. We have made the source code of EFC freely available\footnote{\url{https://github.com/bostjanv76/featConstr}}. 

\subsection{Limitations of the approach}
\label{subsec:limitations}
The main limitation of the proposed approach is that the method fails when there are no interactions between attributes in the dataset. The EFC method is also sensitive to the settings of parameters, such as $noiseThr$ and the number of discretisation intervals. If the $noiseThr$ parameter (the expected noise in explanations affecting groups of co-occuring attributes) is set too low, there is a risk of combinatorial explosion, especially with large datasets. When constructing logical features from numerical attributes, the potentially large number of discretisation intervals may increase the time and space complexity of the FC process.

\subsection{Future works}
\label{subsec:future-works}
There are ample opportunities for future work. It seems sensible to test an integration of feature complexity measures into the feature construction process. This could further automate the feature construction process and reduce the search space. Feature evaluation could be added during the FC process for additional control of time and space complexity. We plan to extensively test different parameters of EFC, including different explanation methods. Since EFC uses many parameters, hyperparameter optimisation might contribute to better performance. While we only use tabular data in classification setting, we see no obstacles in applying the EFC approach to regression problems and adapting it to the text classification. In the financial use case, we used feature construction for  both class values but a more systematic analysis on multi-class problems should investigate the effects of repeating the process for all classes. It would be interesting to test EFC in domains like bioinformatics where are many interactive attributes and fewer instances.

\section*{CRediT authorship contribution statement}
\textbf{Boštjan Vouk}: Conceptualisation, Methodology, Software, Validation, Formal analysis, Visualisation, Writing - Original Draft. \textbf{Matej Guid}: Conceptualisation, Validation, Writing - Review \& Editing. \textbf{Marko Robnik-Šikonja}: Conceptualisation, Methodology, Validation, Writing - Review \& Editing.

\section*{Declaration of competing interest}
The authors declare that they have no known competing financial interests or personal relationships that could have appeared to influence the work reported in this paper.

\section*{Acknowledgements}
The work was supported by the Slovenian Research Agency (ARRS) core research programme P6-0411. We thank the domain expert (professor of finance) for participating in the experiment, Sa\v{s}a Sirk for proofreading the paper, and anonymous reviewers for useful comments.

\clearpage
\appendix
\section{Additional results on the UCI datasets}
\label{sec:additionalTables}
\begin{table}[!htb]
\caption{CA of the DT classifier on the UCI datasets: without feature construction (Base), using only logical features (Log), relational features (Rel), Cartesian product (Cart), decision rule and threshold features (DrThr), all features (All), all with features selection (FS), exhaustive search (Exh), Jakulin's method (Jak), and "upper bound" classifier XGBoost (XGB).}
\label{tab:UCIDatasets-ACC-DTfinal}
 \resizebox{\linewidth}{!}{
    \begin{tabular}{l|cccccccccc}
    \textbf{Dataset} & 
    \multicolumn{1}{l}{\textbf{Base}} & 
    \multicolumn{1}{c}{\textbf{Log}} & 
    \multicolumn{1}{c}{\textbf{Rel}} & 
    \multicolumn{1}{c}{\textbf{Cart}} & 
    \multicolumn{1}{c}{\textbf{DrThr}} & 
    \multicolumn{1}{c}{\textbf{All}} & 
    \multicolumn{1}{c}{\textbf{FS}} & 
    \multicolumn{1}{c}{\textbf{Exh}} & 
    \multicolumn{1}{c}{\textbf{Jak}} & \multicolumn{1}{c}{\textbf{XGB}} \\
    \hline
    Aus. credit & 86.09 & 83.62 & 85.36 & 85.36 & \textbf{86.38} & 85.22 & 84.64 & 83.62 & 85.51 & 86.81 \\
    Autos & 81.40 & 80.43 & 81.40 & 78.45 & 77.00 & 78.95 & 79.45 & \multicolumn{1}{c}{-} & \textbf{82.76} & 80.93 \\
    Balance scale & 76.65 & 81.93 & \textbf{98.08} & 79.05 & 82.57 & 86.89 & 89.12 & 91.54 & 78.57 & 88.80 \\
    Bank mark. & 88.98 & 89.52 & 88.96 & 88.83 & 89.43 & 89.32 & 88.98 & \multicolumn{1}{c}{-} & \textbf{89.71} & 89.78 \\
    Bio. resp. & 73.77 & 74.17 & 73.93 & \textbf{74.81} & 73.71 & 72.94 & 73.93 & \multicolumn{1}{c}{-} & 73.98 & 78.81 \\
    Car   & 92.36 & 96.82 & 92.36 & 93.29 & 94.79 & 96.82 & 96.00 & \textbf{97.57} & 94.15 & 97.63 \\
    CNAE-9 & 88.80 & \textbf{89.44} & 88.98 & 88.80 & 88.89 & \textbf{89.44} & 88.89 & \multicolumn{1}{c}{-} & 89.07 & 91.85 \\
    Glass & 66.75 & 63.48 & 68.23 & 64.48 & 65.37 & 63.96 & 65.37 & 67.29 & \textbf{70.63} & 76.62 \\
    Heart & 76.67 & 80.00 & 77.04 & 79.63 & 80.37 & 78.52 & 80.00 & \textbf{81.85} & 79.26 & 80.74 \\
    Ionosphere & 91.46 & \textbf{92.32} & 87.17 & 91.46 & 91.17 & \textbf{92.32} & \textbf{92.32} & \multicolumn{1}{c}{-} & 89.75 & 93.75 \\
    Jap. credit & 85.94 & 82.75 & 86.23 & 84.64 & \textbf{87.39} & 85.51 & \textbf{87.39} & 84.06 & 86.09 & 87.54 \\
    Leukemia & 83.04 & \textbf{92.86} & 85.89 & 83.04 & 91.61 & 91.61 & 91.61 & \multicolumn{1}{c}{-} & 83.04 & 95.71 \\
    Liver dis. & 68.71 & 67.83 & 64.67 & 67.83 & \textbf{70.79} & 67.86 & 69.00 & 67.22 & 68.12 & 73.70 \\
    LSVT voice & 75.45 & 78.40 & 71.41 & 77.63 & 77.50 & \textbf{79.04} & 77.50 & \multicolumn{1}{c}{-} & 74.42 & 82.44 \\
    Lung cancer & 50.83 & 50.83 & 50.83 & 50.83 & 47.50 & 50.83 & 50.83 & \multicolumn{1}{c}{-} & 44.17 & 58.33 \\
    Mamm. mass & 82.10 & 82.21 & 82.62 & 81.48 & \textbf{83.66} & 82.00 & 82.52 & 82.21 & 81.79 & 83.46 \\
    Molecular  & 80.82 & 85.73 & 80.82 & 75.36 & \textbf{89.55} & 85.73 & 83.82 & \multicolumn{1}{c}{-} & 80.82 & 87.64 \\
    Monks1 & 96.52 & 99.07 & 96.52 & 99.07 & 99.07 & 99.07 & 99.07 & \textbf{100.00} & \textbf{100.00} & 85.67 \\
    Monks2 & 67.14 & \textbf{95.62} & 67.14 & 67.14 & 66.21 & \textbf{95.62} & 93.78 & 92.86 & 67.14 & 87.30 \\
    Nursery & 97.05 & 98.50 & 97.05 & 96.06 & 98.36 & 98.52 & 98.38 & \textbf{99.51} & 96.50 & 99.94 \\
    Parkinsons & 80.50 & 84.66 & 80.00 & 81.55 & \textbf{89.26} & 87.79 & 86.79 & 88.71 & 86.71 & 93.34 \\
    QSAR  & 82.37 & 83.31 & 83.51 & \textbf{84.36} & 83.79 & 83.70 & 82.18 & \multicolumn{1}{c}{-} & 82.57 & 86.64 \\
    Semeion & 75.32 & \textbf{77.03} & 75.32 & 75.51 & 76.33 & 76.40 & 76.02 & \multicolumn{1}{c}{-} & 75.51 & 91.15 \\
    Sonar & 71.17 & 74.00 & 72.14 & 73.10 & \textbf{74.07} & 72.17 & 72.19 & \multicolumn{1}{c}{-} & 68.71 & 83.60 \\
    Spambase & 92.98 & 92.81 & \textbf{93.33} & 92.81 & 92.87 & 92.94 & 92.85 & \multicolumn{1}{c}{-} & 92.52 & 95.35 \\
    Spect heart & 80.91 & 82.41 & 80.91 & 81.32 & 83.52 & 83.13 & 82.76 & 83.18 & \textbf{83.53} & 83.92 \\
    Spectf heart & 85.39 & 87.08 & 85.96 & 85.37 & \textbf{89.39} & 88.53 & 87.67 & \multicolumn{1}{c}{-} & 84.81 & 89.10 \\
    Tic-tac-toe & 84.55 & 97.60 & 84.55 & 74.22 & \textbf{97.91} & 97.39 & 97.18 & 96.87 & 78.60 & 98.33 \\
    Vehicle & 72.47 & 72.69 & 71.63 & \textbf{74.12} & 73.41 & 72.71 & 72.58 & \multicolumn{1}{c}{-} & 73.05 & 77.91 \\
    Voting & 96.33 & 96.33 & 96.33 & 95.87 & 96.33 & 96.33 & 96.33 & 94.96 & 96.33 & 95.63 \\
    \hline
    \textbf{Average} & 81.08 & 83.78 & 81.61 & 80.85 & 83.27 & 84.04 & 83.97 & 83.52* & 81.26 & 86.75
    \end{tabular}
}
\end{table}

\begin{table}[!htb]
\caption{CA of the NB classifier on the UCI datasets: without feature construction (Base), using only logical features (Log), relational features (Rel), Cartesian product (Cart), decision rule and threshold features (DrThr), all features (All), all with features selection (FS), exhaustive search (Exh), Jakulin's method (Jak), and "upper bound" classifier XGBoost (XGB).}
\label{tab:UCIDatasets-ACC-NBfinal}
 \resizebox{\linewidth}{!}{
    \begin{tabular}{l|cccccccccc}
    \textbf{Dataset} & 
    \multicolumn{1}{l}{\textbf{Base}} & 
    \multicolumn{1}{c}{\textbf{Log}} & 
    \multicolumn{1}{c}{\textbf{Rel}} & 
    \multicolumn{1}{c}{\textbf{Cart}} & 
    \multicolumn{1}{c}{\textbf{DrThr}} & 
    \multicolumn{1}{c}{\textbf{All}} & 
    \multicolumn{1}{c}{\textbf{FS}} & 
    \multicolumn{1}{c}{\textbf{Exh}} & 
    \multicolumn{1}{c}{\textbf{Jak}} & 
    \multicolumn{1}{c}{\textbf{XGB}} \\
    \hline
    Aus. credit & 77.54 & 85.51 & 78.26 & 84.78 & 86.09 & \textbf{86.67} & 85.80 & 85.07 & 82.61 & 86.81 \\
    Autos & 56.67 & 58.48 & 57.64 & 58.10 & 59.10 & 59.45 & 60.95 & \multicolumn{1}{c}{-} & \textbf{67.64} & 80.93 \\
    Balance scale & 90.39 & 76.17 & 88.97 & 82.89 & 78.26 & 78.26 & 90.39 & 79.37 & 82.73 & 88.80 \\
    Bank mark. & 86.88 & 78.04 & 86.93 & 86.88 & \textbf{87.28} & 77.88 & 86.88 & \multicolumn{1}{c}{-} & 86.07 & 89.78 \\
    Bio. resp. & 64.77 & 64.86 & 64.77 & 64.77 & 64.77 & \textbf{65.62} & 65.22 & \multicolumn{1}{c}{-} & 64.44 & 78.81 \\
    Car   & 85.53 & 82.81 & 85.53 & 87.04 & 72.80 & 82.81 & 85.13 & 81.89 & \textbf{88.72} & 97.63 \\
    CNAE-9 & 93.15 & 92.41 & 92.96 & 92.78 & 92.96 & 92.50 & 92.59 & \multicolumn{1}{c}{-} & 91.94 & 91.85 \\
    Glass & 48.59 & 51.86 & 51.86 & 58.35 & 48.59 & 52.34 & 55.11 & \textbf{67.73} & 61.73 & 76.62 \\
    Heart & 83.70 & 83.33 & 83.70 & 82.22 & 84.07 & 83.70 & \textbf{84.81} & 82.22 & 80.74 & 80.74 \\
    Ionosphere & 82.62 & 90.60 & 85.45 & 92.03 & 89.47 & 90.60 & \textbf{92.60} & \multicolumn{1}{c}{-} & 88.89 & 93.75 \\
    Jap. credit & 77.68 & 85.07 & 79.42 & 84.64 & \textbf{86.52} & 85.51 & 86.23 & 84.49 & 82.32 & 87.54 \\
    Leukemia & 98.57 & 98.57 & 98.57 & 98.57 & 98.57 & 98.57 & 98.57 & \multicolumn{1}{c}{-} & 98.57 & 95.71 \\
    Liver dis. & 55.39 & 54.80 & 64.66 & 56.53 & \textbf{68.43} & 62.66 & 62.66 & 61.17 & 56.23 & 73.70 \\
    LSVT voice & 56.41 & 61.22 & 57.18 & 57.18 & 57.18 & \textbf{62.76} & \textbf{62.76} & \multicolumn{1}{c}{-} & 56.41 & 82.44 \\
    Lung cancer & 53.33 & 64.17 & 53.33 & 60.83 & 63.33 & 54.17 & \textbf{74.17} & \multicolumn{1}{c}{-} & 60.00 & 58.33 \\
    Mamm. mass & 78.36 & 82.83 & 77.53 & 82.21 & \textbf{83.04} & 82.62 & 82.73 & 82.83 & 80.75 & 83.46 \\
    Molecular & 90.27 & \textbf{92.36} & 90.27 & 90.27 & 88.45 & \textbf{92.36} & 91.36 & \multicolumn{1}{c}{-} & 91.45 & 87.64 \\
    Monks1 & 75.02 & 87.74 & 75.02 & 87.74 & 87.74 & 87.74 & 87.74 & 94.92 & \textbf{100.00} & 85.67 \\
    Monks2 & 66.44 & 53.23 & 66.44 & 53.22 & 62.25 & 52.76 & 65.98 & 53.25 & 53.48 & 87.30 \\
    Nursery & 90.32 & 54.80 & 90.32 & 88.13 & 78.87 & 53.85 & 90.22 & 89.86 & 87.25 & 99.94 \\
    Parkinsons & 69.34 & 80.18 & 69.34 & 76.61 & 78.13 & 81.24 & \textbf{82.26} & 80.66 & 70.84 & 93.34 \\
    QSAR & 75.94 & 79.24 & 77.46 & 79.35 & \textbf{81.71} & 79.71 & 79.71 & \multicolumn{1}{c}{-} & 78.11 & 86.64 \\
    Semeion & 85.56 & 64.84 & 85.56 & 81.23 & 84.74 & 61.89 & 85.25 & \multicolumn{1}{c}{-} & \textbf{86.00} & 91.15 \\
    Sonar & 67.88 & 74.57 & 70.26 & 75.10 & 72.19 & \textbf{77.00} & 76.52 & \multicolumn{1}{c}{-} & 72.69 & 83.60 \\
    Spambase & 79.29 & 88.18 & 79.57 & 80.09 & 80.18 & \textbf{89.05} & 88.98 & \multicolumn{1}{c}{-} & 80.00 & 95.35 \\
    Spect heart & 78.96 & 78.23 & 78.96 & 76.37 & \textbf{81.60} & 78.23 & 77.86 & 77.11 & 77.48 & 83.92 \\
    Spectf heart & 70.78 & 73.91 & 72.21 & 71.92 & 73.06 & \textbf{74.22} & 73.93 & \multicolumn{1}{c}{-} & 72.21 & 89.10 \\
    Tic-tac-toe & 69.62 & 72.65 & 69.62 & 70.36 & \textbf{79.22} & 73.28 & 72.03 & 71.93 & 70.15 & 98.33 \\
    Vehicle & 44.80 & 61.46 & 47.64 & 64.42 & \textbf{67.15} & 61.35 & 64.66 & \multicolumn{1}{c}{-} & 66.44 & 77.91 \\
    Voting & 90.14 & 94.27 & 90.14 & 93.36 & 94.95 & \textbf{95.42} & 94.73 & 91.06 & 90.36 & 95.63 \\
    \hline
    \textbf{Average} & 74.80 & 75.55 & 75.65 & 77.27 & 77.69 & 75.81 & 79.93 & 76.35* & 77.54 & 86.75
    \end{tabular}
 }
\end{table}

\begin{table}[!htp]
\caption{CA of the SVM classifier with linear kernel on the UCI datasets: without feature construction (Base), using only logical features (Log), relational features (Rel), Cartesian product (Cart), decision rule and threshold features (DrThr), all features (All), all with features selection (FS), exhaustive search (Exh), Jakulin's method (Jak), and "upper bound" classifier XGBoost (XGB).}
\label{tab:UCIDatasets-ACC-SVMlin}
 \resizebox{\linewidth}{!}{
    \begin{tabular}{l|cccccccccc}
    \textbf{Dataset} & 
    \multicolumn{1}{l}{\textbf{Base}} & 
    \multicolumn{1}{l}{\textbf{Log}} & 
    \multicolumn{1}{l}{\textbf{Rel}} & 
    \multicolumn{1}{l}{\textbf{Cart}} & 
    \multicolumn{1}{l}{\textbf{DrThr}} & 
    \multicolumn{1}{l}{\textbf{All}} & 
    \multicolumn{1}{l}{\textbf{FS}} & 
    \multicolumn{1}{l}{\textbf{Exh}} & 
    \multicolumn{1}{l}{\textbf{Jak}} & 
    \multicolumn{1}{l}{\textbf{XGB}} \\
    \hline
    Aus. credit & 84.64 & 82.32 & 84.93 & 83.62 & \textbf{86.67} & 82.32 & 85.22 & 77.10 & 83.33 & 86.81 \\
    Autos & 71.19 & 77.55 & 72.17 & 78.55 & 73.64 & 78.05 & 77.55 & \multicolumn{1}{c}{-} & \textbf{80.90} & 80.93 \\
    Balance scale & 87.68 & 86.25 & 93.76 & 86.09 & 94.07 & 96.48 & 95.67 & \textbf{96.79} & 87.20 & 88.80 \\
    Bank mark. & 89.29 & 89.67 & 89.29 & 89.63 & 89.38 & 89.65 & 89.52 & \multicolumn{1}{c}{-} & \textbf{89.85} & 89.78 \\
    Bio. resp. & 75.58 & 75.31 & 75.26 & 75.23 & \textbf{76.03} & 75.37 & 75.63 & \multicolumn{1}{c}{-} & 74.83 & 78.81 \\
    Car   & 93.75 & 98.38 & 93.75 & 97.57 & 95.43 & 98.15 & 98.03 & \textbf{99.94} & 97.28 & 97.63 \\
    CNAE-9 & 94.17 & 93.80 & 93.89 & 93.89 & 94.17 & 93.80 & 93.70 & \multicolumn{1}{c}{-} & 92.41 & 91.85 \\
    Glass & 56.13 & 64.57 & 59.37 & 64.59 & 55.63 & 63.61 & 62.71 & \textbf{70.58} & 67.73 & 76.62 \\
    Heart & 84.07 & 81.11 & 82.22 & 81.48 & 81.85 & 80.00 & 82.59 & 77.04 & 84.07 & 80.74 \\
    Ionosphere & 88.60 & 92.32 & 90.60 & \textbf{92.89} & 92.60 & 91.74 & 91.46 & \multicolumn{1}{c}{-} & 91.48 & 93.75 \\
    Jap. credit & 84.93 & 83.04 & 84.49 & 84.20 & \textbf{87.10} & 84.06 & 84.93 & 76.81 & 82.90 & 87.54 \\
    Leukemia & 95.71 & 95.71 & 97.14 & \textbf{98.57} & 94.46 & 97.14 & 95.71 & \multicolumn{1}{c}{-} & 95.71 & 95.71 \\
    Liver dis. & 58.28 & 56.55 & 65.81 & 56.55 & 70.14 & \textbf{71.61} & 71.02 & 67.51 & 56.55 & 73.70 \\
    LSVT voice & 85.77 & 79.81 & 84.17 & 80.71 & 83.08 & 79.87 & 80.64 & \multicolumn{1}{c}{-} & 78.59 & 82.44 \\
    Lung cancer & 42.50 & 48.33 & 42.50 & 45.83 & \textbf{49.17} & 48.33 & 48.33 & \multicolumn{1}{c}{-} & 42.50 & 58.33 \\
    Mamm. mass & 79.29 & \textbf{83.14} & 81.68 & 82.42 & \textbf{83.14} & 82.52 & 81.48 & 82.31 & 81.79 & 83.46 \\
    Molecular & 93.27 & 87.45 & 93.27 & 90.36 & 91.36 & 87.45 & 89.27 & \multicolumn{1}{c}{-} & \textbf{94.27} & 87.64 \\
    Monks1 & 75.02 & 86.83 & 75.02 & 87.74 & 87.74 & 87.97 & 87.74 & \textbf{100.00} & \textbf{100.00} & 85.67 \\
    Monks2 & 67.14 & 95.39 & 67.14 & 95.40 & 67.14 & 95.39 & 88.45 & \textbf{100.00} & 68.30 & 87.30 \\
    Nursery & 93.13 & 97.30 & 93.13 & 97.15 & 93.01 & 97.30 & 97.23 & \textbf{99.97} & 95.32 & 99.94 \\
    Parkinsons & 87.18 & 80.03 & 87.68 & 82.13 & \textbf{89.32} & 87.29 & 86.79 & 87.18 & 83.13 & 93.34 \\
    QSAR  & 85.59 & 83.50 & 84.55 & 83.50 & 85.12 & 85.22 & 84.55 & \multicolumn{1}{c}{-} & 84.83 & 86.64 \\
    Semeion & 93.85 & 93.60 & 93.85 & 93.72 & \textbf{94.10} & 92.90 & 93.66 & \multicolumn{1}{c}{-} & 94.04 & 91.15 \\
    Sonar & 75.95 & 76.98 & 76.93 & 79.38 & 77.43 & 74.50 & 76.90 & \multicolumn{1}{c}{-} & \textbf{79.43} & 83.60 \\
    Spambase & 90.41 & 91.31 & 90.00 & 91.33 & 91.18 & \textbf{91.72} & 91.37 & \multicolumn{1}{c}{-} & 91.35 & 95.35 \\
    Spect heart & 81.28 & 80.53 & 81.28 & 79.79 & \textbf{83.13} & 80.53 & 79.77 & 81.70 & 80.53 & 83.92 \\
    Spectf heart & 81.08 & 79.65 & 78.51 & 79.08 & 78.22 & 78.78 & 79.66 & \multicolumn{1}{c}{-} & 80.22 & 89.10 \\
    Tic-tac-toe & 98.33 & 99.69 & 98.33 & 98.95 & 98.64 & 99.79 & \textbf{99.90} & 99.58 & 98.33 & 98.33 \\
    Vehicle & 74.36 & 71.51 & 74.13 & 70.80 & 72.59 & 70.81 & 71.87 & \multicolumn{1}{c}{-} & 71.39 & 77.91 \\
    Voting & 95.87 & 96.33 & 95.87 & \textbf{96.56} & 96.33 & 96.33 & 96.10 & 95.40 & 95.64 & 95.63 \\
    \hline
    \textbf{Average} & 82.13 & 83.60 & 82.69 & 83.92 & 83.73 & 84.62 & 84.58 & 84.97* & 83.46 & 86.75
    \end{tabular}
  }
\end{table}

\begin{table}[!htbp]
\caption{CA of the SVM classifier with RBF kernel on the UCI datasets: without feature construction (Base), using only logical features (Log), relational features (Rel), Cartesian product (Cart), decision rule and threshold features (DrThr), all features (All), all with features selection (FS), exhaustive search (Exh), Jakulin's method (Jak), and "upper bound" classifier XGBoost (XGB).}
\label{tab:UCIDatasets-ACC-SVMRBF}
 \resizebox{\linewidth}{!}{
    \begin{tabular}{l|cccccccccc}
    \textbf{Dataset} & 
	\multicolumn{1}{l}{\textbf{Base}} & 
	\multicolumn{1}{l}{\textbf{Log}} & 
	\multicolumn{1}{l}{\textbf{Rel}} & 
	\multicolumn{1}{l}{\textbf{Cart}} & 
	\multicolumn{1}{l}{\textbf{DrThr}} & 
	\multicolumn{1}{l}{\textbf{All}} & 
	\multicolumn{1}{l}{\textbf{FS}} & 
	\multicolumn{1}{l}{\textbf{Exh}} & 
	\multicolumn{1}{l}{\textbf{Jak}} & 
	\multicolumn{1}{l}{\textbf{XGB}} \\
    \hline
    Aus. credit & 85.51 & 85.22 & 85.51 & 85.51 & \textbf{85.94} & 85.65 & 85.80 & 58.55 & 85.80 & 86.81 \\
    Autos & 46.31 & 56.12 & 45.81 & 48.26 & 46.79 & \textbf{56.12} & \textbf{56.12} & \multicolumn{1}{c}{-} & 52.10 & 80.93 \\
    Balance scale & 86.87 & 77.44 & 86.40 & 84.01 & \textbf{89.76} & 88.80 & 89.44 & 88.96 & 85.28 & 88.80 \\
    Bank mark. & 88.48 & 88.87 & 88.48 & 88.61 & 88.56 & \textbf{88.90} & 88.83 & \multicolumn{1}{c}{-} & 88.59 & 89.78 \\
    Bio. resp. & 78.22 & 78.54 & 78.35 & 78.43 & 78.54 & 78.62 & 78.27 & \multicolumn{1}{c}{-} & \textbf{78.65} & 78.81 \\
    Car   & 84.66 & 93.40 & 84.66 & 90.34 & 87.10 & 93.35 & \textbf{93.87} & 75.75 & 88.31 & 97.63 \\
    CNAE-9 & 78.61 & 79.17 & 79.07 & 78.89 & \textbf{79.44} & 78.24 & 78.70 & \multicolumn{1}{c}{-} & 62.50 & 91.85 \\
    Glass & 35.52 & 58.90 & 41.10 & 59.39 & 40.63 & 58.90 & 58.03 & \textbf{70.13} & 61.80 & 76.62 \\
    Heart & 82.59 & 80.37 & 81.85 & 81.48 & 81.11 & 81.48 & 82.22 & \textbf{85.19} & 81.85 & 80.74 \\
    Ionosphere & 76.37 & 91.77 & 86.61 & \textbf{92.89} & 90.60 & 91.77 & 91.75 & \multicolumn{1}{c}{-} & 89.78 & 93.75 \\
    Jap. credit & 85.51 & 85.51 & 85.51 & 84.93 & \textbf{86.38} & 85.80 & 86.09 & 59.13 & 85.22 & 87.54 \\
    Leukemia & 65.36 & 65.36 & 65.36 & 65.36 & 65.36 & 65.36 & 65.36 & \multicolumn{1}{c}{-} & 65.36 & 95.71 \\
    Liver dis. & 57.98 & 56.25 & 57.98 & 55.96 & 67.58 & 67.87 & 67.29 & \textbf{68.12} & 56.55 & 73.70 \\
    LSVT voice & 68.21 & 83.08 & 77.69 & 83.14 & \textbf{83.97} & 82.31 & 81.41 & \multicolumn{1}{c}{-} & 76.73 & 82.44 \\
    Lung cancer & 40.00 & 56.67 & 40.00 & 40.00 & 43.33 & \textbf{56.67} & \textbf{56.67} & \multicolumn{1}{c}{-} & 43.33 & 58.33 \\
    Mamm. mass & 77.94 & 82.52 & 79.61 & 82.62 & \textbf{83.56} & 82.31 & 81.90 & 82.62 & 83.56 & 83.46 \\
    Molecular & 92.36 & 58.73 & 92.36 & 91.27 & 91.27 & 57.82 & 90.36 & \multicolumn{1}{c}{-} & 92.36 & 87.64 \\
    Monks1 & 75.02 & 89.60 & 75.02 & 75.72 & 87.74 & 89.60 & 87.74 & \textbf{100.00} & 75.95 & 85.67 \\
    Monks2 & 67.14 & 66.68 & 67.14 & 67.14 & 67.14 & 66.45 & \textbf{67.36} & 66.90 & 67.14 & 87.30 \\
    Nursery & 92.27 & 81.20 & 92.27 & 95.60 & 93.69 & 73.82 & 93.52 & \textbf{95.88} & 94.05 & 99.94 \\
    Parkinsons & 75.39 & 84.68 & 75.39 & 84.63 & \textbf{90.37} & 86.21 & 87.24 & 83.58 & 81.00 & 93.34 \\
    QSAR  & 66.35 & 78.48 & 74.80 & 78.49 & 81.62 & 82.37 & \textbf{82.46} & \multicolumn{1}{c}{-} & 78.77 & 86.64 \\
    Semeion & 94.92 & 91.15 & 94.92 & 93.97 & 94.73 & 87.51 & 94.85 & \multicolumn{1}{c}{-} & \textbf{94.98} & 91.15 \\
    Sonar & 68.83 & 76.48 & 74.14 & 76.90 & 76.40 & 76.95 & 76.00 & \multicolumn{1}{c}{-} & \textbf{77.45} & 83.60 \\
    Spambase & 74.61 & 89.00 & 84.98 & 88.61 & 88.42 & 89.33 & \textbf{89.50} & \multicolumn{1}{c}{-} & 84.39 & 95.35 \\
    Spect heart & 79.42 & 81.64 & 79.42 & 82.04 & 82.41 & 82.39 & 82.39 & \textbf{82.81} & 79.42 & 83.92 \\
    Spectf heart & 72.78 & 73.64 & 72.78 & 72.21 & \textbf{78.50} & 77.93 & 77.93 & \multicolumn{1}{c}{-} & 73.92 & 89.10 \\
    Tic-tac-toe & 76.41 & 73.93 & 76.41 & 91.33 & \textbf{97.18} & 69.22 & 94.99 & 65.34 & 75.68 & 98.33 \\
    Vehicle & 39.48 & 65.24 & 48.10 & 67.38 & 66.56 & 65.95 & \textbf{68.32} & \multicolumn{1}{c}{-} & 65.72 & 77.91 \\
    Voting & 94.50 & 95.87 & 94.50 & 95.64 & 95.64 & \textbf{96.10} & 95.64 & 91.73 & 94.73 & 95.63 \\
    \hline
    \textbf{Average} & 73.59 & 77.52 & 75.54 & 78.69 & 79.68 & 78.13 & 81.00 & 74.19* & 77.37 & 86.75
    \end{tabular}
  }
\end{table}
\begin{table}[!htbp]
\caption{CA of the kNN classifier on the UCI datasets: without feature construction (Base), using only logical features (Log), relational features (Rel), Cartesian product (Cart), decision rule and threshold features (DrThr), all features (All), all with features selection (FS), exhaustive search (Exh), Jakulin's method (Jak), and "upper bound" classifier XGBoost (XGB).}
\label{tab:UCIDatasets-ACC-kNN}
 \resizebox{\linewidth}{!}{
    \begin{tabular}{l|cccccccccc}
    \textbf{Dataset} & 
	\multicolumn{1}{l}{\textbf{Base}} & 
	\multicolumn{1}{l}{\textbf{Log}} & 
	\multicolumn{1}{l}{\textbf{Rel}} & 
	\multicolumn{1}{l}{\textbf{Cart}} & 
	\multicolumn{1}{l}{\textbf{DrThr}} & 
	\multicolumn{1}{l}{\textbf{All}} & 
	\multicolumn{1}{l}{\textbf{FS}} & 
	\multicolumn{1}{l}{\textbf{Exh}} & 
	\multicolumn{1}{l}{\textbf{Jak}} & \multicolumn{1}{l}{\textbf{XGB}} \\
    \hline
    Aus. credit & 86.52 & 85.07 & 85.36 & 85.65 & 86.38 & 84.78 & 86.09 & 85.36 & 83.91 & 86.81 \\
    Autos & 59.55 & 61.95 & 61.00 & 61.05 & 61.50 & 61.95 & 61.93 & \multicolumn{1}{c}{-} & \textbf{61.98} & 80.93 \\
    Balance scale & 90.08 & 85.77 & \textbf{96.16} & 85.77 & 86.55 & 84.97 & 88.95 & 86.56 & 85.77 & 88.80 \\
    Bank mark. & 88.56 & 89.03 & 88.59 & 88.94 & 88.63 & \textbf{89.10} & 89.07 & \multicolumn{1}{c}{-} & 88.92 & 89.78 \\
    Bio. resp. & 73.58 & 76.49 & 74.75 & 74.38 & 74.70 & \textbf{76.97} & 76.94 & \multicolumn{1}{c}{-} & 73.90 & 78.81 \\
    Car   & 93.52 & 82.58 & 93.52 & 90.57 & 85.13 & 84.26 & 93.52 & 93.29 & 92.88 & 97.63 \\
    CNAE-9 & 82.96 & 82.59 & 82.87 & 82.96 & \textbf{83.33} & 82.59 & 82.78 & \multicolumn{1}{c}{-} & 75.65 & 91.85 \\
    Glass & 66.39 & 59.37 & 66.41 & 61.71 & 65.91 & 61.71 & 61.21 & \textbf{70.58} & 64.98 & 76.62 \\
    Heart & 81.48 & 81.11 & 81.85 & 81.48 & 82.59 & 81.48 & \textbf{83.33} & 81.48 & 81.48 & 80.74 \\
    Ionosphere & 84.89 & 90.60 & 83.48 & 90.89 & 88.03 & 90.60 & \textbf{92.03} & \multicolumn{1}{c}{-} & 85.20 & 93.75 \\
    Jap. credit & 86.09 & 84.78 & 83.91 & 85.65 & \textbf{87.83} & 85.36 & 86.09 & 84.78 & 84.35 & 87.54 \\
    Leukemia & 76.43 & 80.71 & 77.86 & 77.86 & 77.86 & 83.21 & \textbf{83.21} & \multicolumn{1}{c}{-} & 77.86 & 95.71 \\
    Liver dis. & 59.38 & 59.40 & 63.73 & 59.40 & 67.55 & 67.54 & \textbf{68.98} & 66.95 & 60.26 & 73.70 \\
    LSVT voice & 73.01 & 82.31 & 76.15 & \textbf{83.85} & 82.31 & 83.01 & 82.31 & \multicolumn{1}{c}{-} & 79.17 & 82.44 \\
    Lung cancer & 46.67 & 55.00 & 46.67 & 48.33 & \textbf{55.00} & 51.67 & 51.67 & \multicolumn{1}{c}{-} & 45.83 & 58.33 \\
    Mamm. mass & 81.06 & 81.90 & 79.39 & 81.48 & \textbf{83.25} & 82.52 & 83.14 & 82.10 & 81.69 & 83.46 \\
    Molecular & 78.27 & 84.91 & 78.27 & 85.82 & \textbf{90.45} & 86.73 & 87.55 & \multicolumn{1}{c}{-} & 77.18 & 87.64 \\
    Monks1 & 95.61 & 97.02 & 95.61 & 96.55 & 93.32 & 97.02 & 98.85 & \textbf{99.54} & 97.91 & 85.67 \\
    Monks2 & 62.97 & 62.07 & 62.97 & 62.97 & 62.97 & 62.07 & \textbf{69.20} & 62.52 & 59.72 & 87.30 \\
    Nursery & 98.38 & 62.37 & 98.38 & 62.40 & 91.06 & 62.37 & 98.38 & 96.34 & 95.35 & 99.94 \\
    Parkinsons & 89.79 & 85.68 & 86.18 & 85.68 & \textbf{90.34} & 87.21 & 86.18 & 86.74 & 85.24 & 93.34 \\
    QSAR  & 84.93 & 83.42 & 82.46 & 83.51 & 82.47 & 83.23 & 84.64 & \multicolumn{1}{c}{-} & 82.46 & 86.64 \\
    Semeion & 90.39 & 75.33 & 90.39 & 88.89 & 90.27 & 73.88 & 90.52 & \multicolumn{1}{c}{-} & \textbf{90.65} & 91.15 \\
    Sonar & 75.98 & 74.55 & 74.07 & 78.88 & 77.86 & 75.02 & 73.17 & \multicolumn{1}{c}{-} & \textbf{81.29} & 83.60 \\
    Spambase & 89.22 & 89.33 & 89.70 & 89.89 & \textbf{90.37} & 89.05 & 89.89 & \multicolumn{1}{c}{-} & 88.81 & 95.35 \\
    Spect heart & 80.11 & 81.27 & 80.11 & \textbf{83.13} & 82.41 & 81.28 & 81.28 & 81.65 & 81.61 & 83.92 \\
    Spectf heart & 75.08 & 76.20 & 77.08 & 76.20 & \textbf{81.08} & 77.66 & 77.09 & \multicolumn{1}{c}{-} & 76.47 & 89.10 \\
    Tic-tac-toe & 98.85 & 98.33 & 98.85 & \textbf{98.96} & 91.33 & 98.22 & 96.97 & 97.08 & 92.80 & 98.33 \\
    Vehicle & 70.22 & 67.38 & 68.57 & 68.32 & 70.11 & 69.04 & 68.92 & \multicolumn{1}{c}{-} & 70.10 & 77.91 \\
    Voting & 92.90 & 94.73 & 92.90 & 94.73 & \textbf{95.88} & 95.19 & \textbf{95.88} & 93.13 & 93.57 & 95.63 \\
    \hline
    \textbf{Average} & 80.43 & 79.04 & 80.57 & 79.86 & 81.55 & 79.66 & 82.33 & 80.59* & 79.90 & 86.75
    \end{tabular}
    }
\end{table}
\begin{table}[!htb]
\caption{CA of the FURIA classifier on the UCI datasets: without feature construction (Base), using only logical features (Log), relational features (Rel), Cartesian product (Cart), decision rule and threshold features (DrThr), all features (All), all with features selection (FS), exhaustive search (Exh), Jakulin's method (Jak), and "upper bound" classifier XGBoost (XGB).}
\label{tab:UCIDatasets-ACC-FU}
 \resizebox{\linewidth}{!}{
    \begin{tabular}{l|cccccccccc}
    \textbf{Dataset} & 
    \multicolumn{1}{c}{\textbf{Base}} & 
    \multicolumn{1}{c}{\textbf{Log}} & 
    \multicolumn{1}{c}{\textbf{Rel}} & 
    \multicolumn{1}{c}{\textbf{Cart}} & 
    \multicolumn{1}{c}{\textbf{DrThr}} & 
    \multicolumn{1}{c}{\textbf{All}} & 
    \multicolumn{1}{c}{\textbf{FS}} & 
    \multicolumn{1}{c}{\textbf{Exh}} & 
    \multicolumn{1}{c}{\textbf{Jak}} & 
    \multicolumn{1}{c}{\textbf{XGB}} \\
    \hline
    Aus. credit & 86.67 & 84.78 & 85.22 & 85.65 & 85.07 & 85.22 & 85.94 & 85.65 & 86.23 & 86.81 \\
    Autos & 80.98 & 78.45 & 80.95 & 80.98 & 80.48 & 78.95 & 80.50  &   -    & 80.52 & 80.93 \\
    Balance scale & 80.65 & 83.21 & \textbf{97.44} & 83.04 & 89.11 & 94.40  & 95.04 & 95.20  & 81.28 & 88.80 \\
    Bank mark. & 88.96 & 89.45 & 89.16 & 88.85 & 89.18 & \textbf{89.49} & 88.98 &    -   & 89.16 & 89.78 \\
    Bio. resp. & 77.37 & 76.91 & 76.35 & 76.91 & 77.10  & 76.73 & 77.31 &   -    & 76.89 & 78.81 \\
    Car   & 93.40  & 96.76 & 93.40  & 92.59 & 93.29 & 96.82 & 97.22 & \textbf{97.97} & 92.88 & 97.63 \\
    CNAE-9 & 89.35 & 89.44 & \textbf{89.72} & 89.35 & 89.44 & 89.54 & 89.44 &   -    & 89.26 & 91.85 \\
    Glass & 70.50  & 71.47 & 72.92 & \textbf{74.26} & 70.02 & 69.13 & 67.75 & 70.58 & 69.52 & 76.62 \\
    Heart & 81.11 & \textbf{82.22} & 81.48 & 78.15 & 78.52 & 78.89 & 80.74 & 78.15 & 81.48 & 80.74 \\
    Ionosphere & 91.17 & 90.32 & 90.61 & \textbf{91.48} & 90.05 & 89.18 & 89.75 &   -    & 91.46 & 93.75 \\
    Jap. credit & 86.23 & 85.07 & 86.81 & 85.65 & \textbf{87.97} & 86.67 & 85.94 & 86.38 & 86.52 & 87.54 \\
    Leukemia & 87.32 & 92.86 & 90.18 & 90.18 & \textbf{93.04} & 91.61 & 91.61 &    -   & 83.04 & 95.71 \\
    Liver dis. & 67.83 & 68.73 & 68.11 & 68.71 & 68.40  & 70.13 & 69.86 & \textbf{72.20}  & 65.23 & 73.70 \\
    LSVT voice & 82.37 & 79.17 & 80.77 & 80.83 & \textbf{84.04} & 83.21 & 80.71 &   -    & 81.67 & 82.44 \\
    Lung cancer & 60.83 & 60.83 & 60.83 & 57.50  & 50.83 & 57.50  & 47.50  &   -    & 50.83 & 58.33 \\
    Mamm. mass & 82.52 & 82.52 & 83.14 & 81.68 & 82.41 & 82.21 & 82.21 & 82.73 & \textbf{83.25} & 83.46 \\
    Molecular & 86.73 & 83.73 & 86.73 & 83.64 & \textbf{88.64} & 84.64 & 82.91 &    -   & 85.82 & 87.64 \\
    Monks1 & 100.00   & 100.00   & 100.00   & 100.00   & 100.00   & 100.00   & 100.00   & 100.00   & 100.00   & 85.67 \\
    Monks2 & 66.21 & 98.61 & 66.21 & 67.14 & 65.74 & 98.84 & 95.39 & \textbf{99.53} & 66.93 & 87.30 \\
    Nursery & 97.04 & 99.44 & 97.04 & 96.67 & 99.00    & 99.4  & 99.43 & \textbf{99.85} & 97.08 & 99.94 \\
    Parkinsons & 88.79 & 87.26 & 88.29 & 87.74 & 89.84 & 89.79 & 88.82 & \textbf{91.32} & 88.82 & 93.34 \\
    QSAR & 84.37 & \textbf{85.89} & 84.55 & 83.99 & 83.98 & 85.69 & 83.42 &   -    & 84.55 & 86.64 \\
    Semeion & 82.93 & 82.05 & 82.93 & 82.48 & 82.74 & 81.67 & 81.61 &  -     & \textbf{83.93} & 91.15 \\
    Sonar & 79.76 & 79.29 & 79.29 & 78.31 & 77.40  & 78.81 & 82.12 &    -   & \textbf{83.10}  & 83.60 \\
    Spambase & 93.65 & 93.15 & 93.78 & 93.52 & 93.39 & \textbf{93.89} & 93.78 &   -    & 93.44 & 95.35 \\
    Spect heart & 79.05 & 82.02 & 79.05 & 79.43 & 82.76 & 81.65 & 83.13 & \textbf{83.55} & 80.91 & 83.92 \\
    Spectf heart & 87.97 & 87.97 & 88.55 & 86.55 & 87.66 & 87.12 & 88.55 &  -     & \textbf{89.38} & 89.10 \\
    Tic-tac-toe & 98.02 & 98.54 & 98.02 & \textbf{98.64} & 97.80  & 98.02 & 98.33 & 97.60  & 98.43 & 98.33 \\
    Vehicle & 70.57 & 69.97 & 72.22 & 72.69 & 70.33 & 69.63 & 71.52 &  -     & \textbf{73.53} & 77.91 \\
    Voting & 95.87 & 95.17 & 95.87 & 96.09 & 95.64 & 95.41 & 95.87 & 95.42 & \textbf{96.33} & 95.63 \\
    \hline
    \textbf{Average} & 83.94 & 85.18 & 84.65 & 83.76 & 84.13 & 85.47 & 85.18 & 86.02* & 83.72 & 86.75
    \end{tabular}
}
\end{table}

\begin{table}[!htb]
\caption{CA of the RF classifier on the UCI datasets: without feature construction (Base), using only logical features (Log), relational features (Rel), Cartesian product (Cart), decision rule and threshold features (DrThr), all features (All), all with features selection (FS), exhaustive search (Exh), Jakulin's method (Jak), and "upper bound" classifier XGBoost (XGB).}
\label{tab:UCIDatasets-ACC-RF}
 \resizebox{\linewidth}{!}{
    \begin{tabular}{l|cccccccccc}
    \textbf{Dataset} & 
    \multicolumn{1}{c}{\textbf{Base}} & 
    \multicolumn{1}{c}{\textbf{Log}} & 
    \multicolumn{1}{c}{\textbf{Rel}} & 
    \multicolumn{1}{c}{\textbf{Cart}} & 
    \multicolumn{1}{c}{\textbf{DrThr}} & 
    \multicolumn{1}{c}{\textbf{All}} & 
    \multicolumn{1}{c}{\textbf{FS}} & 
    \multicolumn{1}{c}{\textbf{Exh}} & 
    \multicolumn{1}{c}{\textbf{Jak}} & 
    \multicolumn{1}{c}{\textbf{XGB}} \\
    \hline
    Aus. credit & 85.65 & 84.64 & \textbf{87.25} & 85.36 & 85.80 & 84.49 & 85.07 & 85.22 & 84.06 & 86.81 \\
    Autos & 84.79 & 86.26 & 84.79 & 85.79 & 84.79 & \textbf{86.26} & 84.76 &   -    & 82.76 & 80.93 \\
    Balance scale & 81.45 & 82.57 & \textbf{97.12} & 82.72 & 85.28 & 85.45 & 86.41 & 87.05 & 82.08 & 88.80 \\
    Bank mark. & 89.67 & 89.43 & 89.58 & \textbf{89.94} & 89.63 & 89.25 & 89.60 &   -    & 88.85 & 89.78 \\
    Bio. resp. & 79.98 & 80.32 & 79.93 & 80.48 & \textbf{80.64} & 80.14 & 80.41 &   -    & 79.53 & 78.81 \\
    Car   & 94.50 & 93.92 & 94.50 & 94.27 & 94.10 & 93.98 & 95.72 & \textbf{97.40} & 94.73 & 97.63 \\
    CNAE-9 & 93.61 & 93.24 & 93.15 & 92.96 & \textbf{93.70} & 93.70 & 93.15 &   -    & 91.94 & 91.85 \\
    Glass & 79.89 & 73.38 & \textbf{81.26} & 74.26 & 77.99 & 73.38 & 74.78 & 76.60 & 75.58 & 76.62 \\
    Heart & 81.11 & 81.85 & \textbf{83.70} & 81.85 & 81.48 & 78.52 & 82.59 & 82.96 & 80.00 & 80.74 \\
    Ionosphere & 92.89 & 92.60 & 92.89 & 92.88 & \textbf{94.32} & 92.04 & 92.32 &   -    & 93.17 & 93.75 \\
    Jap. credit & 87.39 & 83.48 & 86.96 & 84.64 & 86.96 & 84.35 & 86.38 & 85.65 & 87.25 & 87.54 \\
    Leukemia & 88.75 & 90.18 & 91.61 & 92.86 & 91.61 & \textbf{94.46} & 92.86 &   -    & 87.32 & 95.71 \\
    Liver dis. & 73.11 & 71.64 & 73.35 & 73.08 & 72.20 & 70.75 & \textbf{73.39} & 72.47 & 72.49 & 73.70 \\
    LSVT voice & 85.64 & 83.14 & 83.14 & 83.21 & 80.83 & 83.97 & 82.44 &    -   & 80.71 & 82.44 \\
    Lung cancer & 50.83 & 56.67 & 50.83 & \textbf{63.33} & 53.33 & 60.00 & 60.00 &    -   & 42.50 & 58.33 \\
    Mamm. mass & 79.40 & 79.40 & 79.08 & 79.40 & 79.29 & 79.19 & 79.29 & 79.08 & \textbf{80.13} & 83.46 \\
    Molecular & 93.36 & 89.45 & 93.36 & 93.18 & 90.55 & 91.36 & 89.64 &   -    & 85.64 & 87.64 \\
    Monks1 & 100.00 & 100.00 & 100.00 & 100.00 & 100.00 & 100.00 & 100.00 & 100.00 & 100.00 & 85.67 \\
    Monks2 & 38.64 & 76.62 & 38.64 & 40.51 & 41.20 & 68.27 & \textbf{96.08} & 71.06 & 60.18 & 87.30 \\
    Nursery & 99.10 & 92.24 & 99.10 & 96.40 & 97.15 & 92.00 & 99.10 & \textbf{99.66} & 98.09 & 99.94 \\
    Parkinsons & 92.87 & 86.74 & 91.84 & 87.16 & 90.34 & 88.84 & 89.32 & 87.74 & 91.32 & 93.34 \\
    QSAR & 87.01 & 85.40 & 86.45 & 85.50 & 86.26 & 85.02 & 86.06 &    -   & 86.82 & 86.64 \\
    Semeion & 94.10 & 88.95 & 94.10 & 90.08 & 92.53 & 85.62 & 91.77 &   -    & 93.03 & 91.15 \\
    Sonar & 81.24 & 82.69 & 81.71 & 84.62 & 80.29 & 82.24 & 82.21 &   -    & \textbf{85.12} & 83.60 \\
    Spambase & 95.39 & 93.96 & 95.11 & 94.57 & 95.11 & 93.70 & 95.39 &  -     & 94.78 & 95.35 \\
    Spect heart & 81.31 & 78.68 & 81.31 & 79.80 & 80.93 & 78.68 & 80.20 & \textbf{81.67} & 79.80 & 83.92 \\
    Spectf heart & 91.97 & 90.55 & 90.82 & 91.40 & \textbf{92.27} & 90.55 & 90.25 &  -     & 91.68 & 89.10 \\
    Tic-tac-toe & 96.35 & \textbf{98.64} & 96.35 & 97.91 & 97.08 & 98.02 & 97.81 & 98.23 & 86.01 & 98.33 \\
    Vehicle & 76.01 & 72.11 & 75.06 & 74.36 & 74.60 & 72.58 & 73.76 &   -    & \textbf{76.25} & 77.91 \\
    Voting & 96.10 & 95.18 & 96.10 & 95.41 & 96.10 & 95.41 & 95.64 & \textbf{96.33} & 96.10 & 95.63 \\
    \hline
    \textbf{Average} & 85.07 & 85.13 & 85.64 & 84.93 & 84.88 & 85.07 & 86.88 & 86.21* & 84.26 & 86.75
    \end{tabular}
}
\end{table}

\clearpage
\section{Complexity of decision rules}
\begin{table}[hbt!]
\caption{Complexity of decision rules trained using the original UCI datasets and the enriched datasets. We present $\#rules$ - average number of rules, $\#attr$ - average number of attributes in rules, $\#attr/rule$ - average number of attributes per rule, $\#all$ - number of attributes in all rules. }
\label{tab:featComplexity-DR}
  \centering
    \begin{tabular}{lrrrrrr} & 
    \multicolumn{3}{c}{Original dataset} & 
    \multicolumn{3}{c}{Enriched dataset} \\
    \midrule
    \textbf{Dataset} & 
    \multicolumn{1}{l}{$\#rules$} & 
    \multicolumn{1}{l}{$\#attr$} & 
    \multicolumn{1}{l}{$\#attr/rule$} & 
    \multicolumn{1}{l}{$\#rules$} & 
    \multicolumn{1}{c}{$\#all$} & 
    \multicolumn{1}{l}{$\#attr/rule$} \\
    \midrule
    Aus. credit         & 7.4   & 19.3  & 2.6   & 8.6   & 48.3  & 5.6 \\
    Autos               & 20.2  & 50.6  & 2.5   & 19.1  & 49.8  & 2.6 \\
    Balance scale       & 23.8  & 73.1  & 3.1   & 20.1  & 121.0 & 6.0 \\
    Bank mark.          & 14.5  & 45.7  & 3.2   & 12.7  & 49.8  & 3.9 \\
    Bio. resp.          & 38.6  & 209.0 & 5.4   & 34.2  & 225.5 & 6.6 \\
    Car                 & 80.6  & 345.8 & 4.3   & 28.4  & 210.9 & 7.4 \\
    CNAE-9              & 61.6  & 95.4  & 1.5   & 60.0  & 95.1  & 1.6 \\
    Glass               & 14.2  & 42.5  & 3.0   & 12.8  & 43.2  & 3.4 \\
    Heart               & 7.9   & 21.1  & 2.7   & 7.8   & 36.6  & 4.7 \\
    Ionosphere          & 11.6  & 28.7  & 2.5   & 9.3   & 41.7  & 4.5 \\
    Jap. credit         & 6.7   & 17.8  & 2.7   & 11.0  & 60.3  & 5.5 \\
    Leukemia            & 3.0   & 3.6   & 1.2   & 2.0   & 4.0   & 2.0 \\
    Liver dis.          & 10.2  & 29.5  & 2.9   & 7.8   & 31.1  & 4.0 \\
    LSVT voice          & 6.5   & 13.2  & 2.0   & 5.7   & 15.8  & 2.8 \\
    Lung cancer         & 5.3   & 8.8   & 1.7   & 4.5   & 10.6  & 2.4 \\
    Mamm. mass          & 3.5   & 5.7   & 1.6   & 2.9   & 7.4   & 2.6 \\
    Molecular           & 12.0  & 23.9  & 2.0   & 4.8   & 21.2  & 4.4 \\
    Monks1              & 22.9  & 64.6  & 2.8   & 10.7  & 31.8  & 3.0 \\
    Monks2              & 17.1  & 49.7  & 2.9   & 15.0  & 94.2  & 6.3 \\
    Nursery             & 273.3 & 1367.8 & 5.0   & 103.5 & 1063.2 & 10.3 \\
    Parkinsons          & 8.2   & 19.9  & 2.4   & 7.3   & 27.6  & 3.8 \\
    QSAR                & 20.8  & 71.2  & 3.4   & 20.4  & 111.1 & 5.4 \\
    Semeion             & 88.3  & 455.5 & 5.2   & 82.3  & 491.7 & 6.0 \\
    Sonar               & 11.5  & 30.8  & 2.7   & 10.0  & 31.3  & 3.1 \\
    Spambase            & 34.6  & 148.9 & 4.3   & 34.3  & 203.3 & 5.9 \\
    Spect heart         & 7.1   & 14.5  & 2.0   & 3.5   & 15.5  & 4.4 \\
    Spectf heart        & 16.3  & 50.1  & 3.1   & 16.9  & 56.4  & 3.3 \\
    Tic-tac-toe         & 21.5  & 75.7  & 3.5   & 20.3  & 101.6 & 5.0 \\
    Vehicle             & 24.4  & 83.4  & 3.4   & 18.9  & 104.9 & 5.6 \\
    Voting              & 6.3   & 15.7  & 2.5   & 5.9   & 18.2  & 3.1 \\
    \midrule
    \textbf{Average} & \textbf{29.33} & \textbf{116.05} & \textbf{3.96} & \textbf{20.02} & \textbf{114.10} & \textbf{5.70} \\
    \end{tabular}
\end{table}

\newpage
\section{Running times of feature construction}
\begin{table}[!htbp]
  \centering
\caption{Running times [in sec]. The columns report $t_{EFC}$ - running time of the proposed EFC methodology when using all types of features, $t_{EXH}$ - time of exhaustive search, $sd$ - standard deviation. The symbol '-' indicates that feature construction did not terminate in the maximal allowed time of 3 hours (per fold).}
    \label{tab:computTime}
    \begin{tabular}{lrrrr}
    \textbf{Dataset} & 
    \multicolumn{1}{c}{$t_{EFC}$} & 
    \multicolumn{1}{l}{$sd_{EFC}$} & 
    \multicolumn{1}{c}{$t_{EXH}$} &
    \multicolumn{1}{l}{$sd_{EXH}$} \\
    \midrule
    Aus. credit     & 8.9   & 6.9   & 693.3 & 172.5 \\
    Autos           & 0.0   & 0.1   & \multicolumn{1}{c}{-} & \multicolumn{1}{c}{-} \\
    Balance scale   & 1.0   & 0.2   & 0.5   & 0.1 \\
    Bank mark.      & 2.7   & 2.7   & \multicolumn{1}{c}{-} & \multicolumn{1}{c}{-} \\
    Bio. resp.      & 5.9   & 1.8   & \multicolumn{1}{c}{-} & \multicolumn{1}{c}{-} \\
    Car             & 11.9  & 3.1   & 49.0  & 0.7 \\
    CNAE-9          & 0.4   & 0.2   & \multicolumn{1}{c}{-} & \multicolumn{1}{c}{-} \\
    Glass           & 0.8   & 0.5   & 12.3  & 2.1 \\
    Heart           & 0.7   & 0.2   & 5.9   & 0.1 \\
    Ionosphere      & 34.8  & 86.9  & \multicolumn{1}{c}{-} & \multicolumn{1}{c}{-} \\
    Jap. credit     & 3.0   & 1.4   & 1046.1 & 148.5 \\
    Leukemia        & 1.1   & 0.3   & \multicolumn{1}{c}{-} & \multicolumn{1}{c}{-} \\
    Liver dis.      & 0.6   & 0.1   & 0.3   & 0.0 \\
    LSVT voice      & 2.4   & 4.7   & \multicolumn{1}{c}{-} & \multicolumn{1}{c}{-} \\
    Lung cancer     & 0.1   & 0.1   & \multicolumn{1}{c}{-} & \multicolumn{1}{c}{-} \\
    Mamm. mass      & 0.9   & 0.2   & 1.3   & 0.3 \\
    Molecular       & 125.7 & 81.1  & \multicolumn{1}{c}{-} & \multicolumn{1}{c}{-} \\
    Monks1          & 0.2   & 0.2   & 3.4   & 0.1 \\
    Monks2          & 2.7   & 0.7   & 3.3   & 0.0 \\
    Nursery         & 104.4 & 25.8  & 1744.8 & 70.7 \\
    Parkinsons      & 1.5   & 1.4   & 249.0 & 85.8 \\
    QSAR            & 1.9   & 1.0   & \multicolumn{1}{c}{-} & \multicolumn{1}{c}{-} \\
    Semeion         & 5.2   & 2.6   & \multicolumn{1}{c}{-} & \multicolumn{1}{c}{-} \\
    Sonar           & 0.2   & 0.1   & \multicolumn{1}{c}{-} & \multicolumn{1}{c}{-} \\
    Spambase        & 6.5   & 2.9   & \multicolumn{1}{c}{-} & \multicolumn{1}{c}{-} \\
    Spect heart     & 0.4   & 0.2   & 73.6  & 0.9 \\
    Spectf heart    & 0.1   & 0.1   & \multicolumn{1}{c}{-} & \multicolumn{1}{c}{-} \\
    Tic-tac-toe     & 58.7  & 17.6  & 194.8 & 2.1 \\
    Vehicle         & 311.1 & 272.1 & \multicolumn{1}{c}{-} & \multicolumn{1}{c}{-} \\
    Voting          & 0.1   & 0.0   & 19.0  & 0.7 \\ \hline
    \end{tabular}
\end{table}

\clearpage
\bibliographystyle{elsarticle-harv}
\bibliography{ms}

\providecommand{\NOOPSORT}[1]{}
\begin{thebibliography}{97}
\expandafter\ifx\csname natexlab\endcsname\relax\def\natexlab#1{#1}\fi
\providecommand{\url}[1]{\texttt{#1}}
\providecommand{\href}[2]{#2}
\providecommand{\path}[1]{#1}
\providecommand{\DOIprefix}{doi:}
\providecommand{\ArXivprefix}{arXiv:}
\providecommand{\URLprefix}{URL: }
\providecommand{\Pubmedprefix}{pmid:}
\providecommand{\doi}[1]{\href{http://dx.doi.org/#1}{\path{#1}}}
\providecommand{\Pubmed}[1]{\href{pmid:#1}{\path{#1}}}
\providecommand{\bibinfo}[2]{#2}
\ifx\xfnm\relax \def\xfnm[#1]{\unskip,\space#1}\fi
\bibitem[{Aha(1991)}]{aha1991incremental}
\bibinfo{author}{Aha, D.W.}, \bibinfo{year}{1991}.
\newblock \bibinfo{title}{Incremental constructive induction: An instance-based
  approach}, in: \bibinfo{editor}{Birnbaum, L.A.}, \bibinfo{editor}{Collins,
  G.C.} (Eds.), \bibinfo{booktitle}{{Machine Learning Proceedings 1991}}.
  \bibinfo{publisher}{Elsevier}, pp. \bibinfo{pages}{117--121}.
\newblock \bibinfo{note}{\url{https://doi.org/g8pq}}.
\bibitem[{Albashrawi(2016)}]{albashrawi2016detecting}
\bibinfo{author}{Albashrawi, M.}, \bibinfo{year}{2016}.
\newblock \bibinfo{title}{Detecting financial fraud using data mining
  techniques: A decade review from 2004 to 2015}.
\newblock \bibinfo{journal}{Journal of Data Science} \bibinfo{volume}{14},
  \bibinfo{pages}{553--569}.
\newblock \bibinfo{note}{\url{https://doi.org/gkkjph}}.
\bibitem[{Arjona-Medina et~al.(2019)Arjona-Medina, Gillhofer, Widrich,
  Unterthiner, Brandstetter and Hochreiter}]{arjonamedina2018rudder}
\bibinfo{author}{Arjona-Medina, J.A.}, \bibinfo{author}{Gillhofer, M.},
  \bibinfo{author}{Widrich, M.}, \bibinfo{author}{Unterthiner, T.},
  \bibinfo{author}{Brandstetter, J.}, \bibinfo{author}{Hochreiter, S.},
  \bibinfo{year}{2019}.
\newblock \bibinfo{title}{{RUDDER}: Return decomposition for delayed rewards},
  in: \bibinfo{editor}{Wallach, H.}, \bibinfo{editor}{Larochelle, H.},
  \bibinfo{editor}{Beygelzimer, A.}, \bibinfo{editor}{d\textquotesingle
  Alch\'{e}-Buc, F.}, \bibinfo{editor}{Fox, E.}, \bibinfo{editor}{Garnett, R.}
  (Eds.), \bibinfo{booktitle}{{Advances in Neural Information Processing
  Systems 32 (NIPS 2019)}}. \bibinfo{publisher}{Curran Associates, Inc.}, pp.
  \bibinfo{pages}{13544--13555}.
\newblock \URLprefix \url{https://bit.ly/3phuH6S}.
\bibitem[{Arras et~al.(2017)Arras, Horn, Montavon, Müller and
  Samek}]{Arras_2017}
\bibinfo{author}{Arras, L.}, \bibinfo{author}{Horn, F.},
  \bibinfo{author}{Montavon, G.}, \bibinfo{author}{Müller, K.R.},
  \bibinfo{author}{Samek, W.}, \bibinfo{year}{2017}.
\newblock \bibinfo{title}{"{W}hat is relevant in a text document?": An
  interpretable machine learning approach}.
\newblock \bibinfo{journal}{PLOS ONE} \bibinfo{volume}{12},
  \bibinfo{pages}{e0181142}.
\newblock \bibinfo{note}{\url{https://doi.org/gbrnrz}}.
\bibitem[{Arrieta et~al.(2020)Arrieta, D{\'i}az-Rodr{\'i}guez, Del~Ser,
  Bennetot, Tabik, Barbado, Garc{\'i}a, Gil-L{\'o}pez, Molina, Benjamins,
  Chatila and Herrera}]{arrieta2020XAI}
\bibinfo{author}{Arrieta, A.B.}, \bibinfo{author}{D{\'i}az-Rodr{\'i}guez, N.},
  \bibinfo{author}{Del~Ser, J.}, \bibinfo{author}{Bennetot, A.},
  \bibinfo{author}{Tabik, S.}, \bibinfo{author}{Barbado, A.},
  \bibinfo{author}{Garc{\'i}a, S.}, \bibinfo{author}{Gil-L{\'o}pez, S.},
  \bibinfo{author}{Molina, D.}, \bibinfo{author}{Benjamins, R.},
  \bibinfo{author}{Chatila, R.}, \bibinfo{author}{Herrera, F.},
  \bibinfo{year}{2020}.
\newblock \bibinfo{title}{Explainable artificial intelligence ({XAI}):
  Concepts, taxonomies, opportunities and challenges toward responsible {AI}}.
\newblock \bibinfo{journal}{Information Fusion} \bibinfo{volume}{58},
  \bibinfo{pages}{82--115}.
\newblock \bibinfo{note}{\url{https://doi.org/ggqs5w}}.
\bibitem[{Azadifar et~al.(2022)Azadifar, Rostami, Berahmand, Moradi and
  Oussalah}]{azadifar2022graph}
\bibinfo{author}{Azadifar, S.}, \bibinfo{author}{Rostami, M.},
  \bibinfo{author}{Berahmand, K.}, \bibinfo{author}{Moradi, P.},
  \bibinfo{author}{Oussalah, M.}, \bibinfo{year}{2022}.
\newblock \bibinfo{title}{Graph-based relevancy-redundancy gene selection
  method for cancer diagnosis}.
\newblock \bibinfo{journal}{Computers in Biology and Medicine}
  \bibinfo{volume}{147}, \bibinfo{pages}{105766}.
\newblock \bibinfo{note}{\url{https://doi.org/jcfg}}.
\bibitem[{Bach et~al.(2015)Bach, Binder, Montavon, Klauschen, M{\"u}ller and
  Samek}]{bach2015RelevancePropagation}
\bibinfo{author}{Bach, S.}, \bibinfo{author}{Binder, A.},
  \bibinfo{author}{Montavon, G.}, \bibinfo{author}{Klauschen, F.},
  \bibinfo{author}{M{\"u}ller, K.R.}, \bibinfo{author}{Samek, W.},
  \bibinfo{year}{2015}.
\newblock \bibinfo{title}{On pixel-wise explanations for non-linear classifier
  decisions by layer-wise relevance propagation}.
\newblock \bibinfo{journal}{PLOS ONE} \bibinfo{volume}{10},
  \bibinfo{pages}{e0130140}.
\newblock \bibinfo{note}{\url{https://doi.org/gcgmcp}}.
\bibitem[{Bagallo and Haussler(1990)}]{bagallo1990boolean}
\bibinfo{author}{Bagallo, G.}, \bibinfo{author}{Haussler, D.},
  \bibinfo{year}{1990}.
\newblock \bibinfo{title}{Boolean feature discovery in empirical learning}.
\newblock \bibinfo{journal}{Machine Learning} \bibinfo{volume}{5},
  \bibinfo{pages}{71--99}.
\newblock \bibinfo{note}{\url{https://doi.org/cxtn2p}}.
\bibitem[{Bodria et~al.(2021)Bodria, Giannotti, Guidotti, Naretto, Pedreschi
  and Rinzivillo}]{bodria2021benchmarking}
\bibinfo{author}{Bodria, F.}, \bibinfo{author}{Giannotti, F.},
  \bibinfo{author}{Guidotti, R.}, \bibinfo{author}{Naretto, F.},
  \bibinfo{author}{Pedreschi, D.}, \bibinfo{author}{Rinzivillo, S.},
  \bibinfo{year}{2021}.
\newblock \bibinfo{title}{{Benchmarking and survey of explanation methods for
  black box models}}.
\newblock \bibinfo{journal}{arXiv preprint}
  \href{http://arxiv.org/abs/2102.13076}{{\tt arXiv:2102.13076}}.
\bibitem[{Bohanec et~al.(2017)Bohanec, {Borštnar} and
  {Robnik-Šikonja}}]{Bohanec2017SalesPredictions}
\bibinfo{author}{Bohanec, M.}, \bibinfo{author}{{Borštnar}, M.},
  \bibinfo{author}{{Robnik-Šikonja}, M.}, \bibinfo{year}{2017}.
\newblock \bibinfo{title}{Explaining machine learning models in sales
  predictions}.
\newblock \bibinfo{journal}{Expert Systems with Applications}
  \bibinfo{volume}{71}, \bibinfo{pages}{416–428}.
\newblock \bibinfo{note}{\url{https://doi.org/f9ph9c}}.
\bibitem[{Calderon-Monge and
  Pastor-Sanz(2017)}]{calderon2017-EbitToFinancialExpenses}
\bibinfo{author}{Calderon-Monge, E.}, \bibinfo{author}{Pastor-Sanz, I.},
  \bibinfo{year}{2017}.
\newblock \bibinfo{title}{Effects of contract and trust on franchisor
  performance}.
\newblock \bibinfo{journal}{Contemporary Economics} \bibinfo{volume}{11},
  \bibinfo{pages}{383--401}.
\newblock \URLprefix \url{https://bit.ly/3eerdLP}.
\bibitem[{Chapelle(2007)}]{chapelle2007training}
\bibinfo{author}{Chapelle, O.}, \bibinfo{year}{2007}.
\newblock \bibinfo{title}{Training a support vector machine in the primal}.
\newblock \bibinfo{journal}{Neural Computation} \bibinfo{volume}{19},
  \bibinfo{pages}{1155--1178}.
\newblock \bibinfo{note}{\url{https://doi.org/dhk8ws}}.
\bibitem[{Chen and Guestrin(2016)}]{chen2016xgboost}
\bibinfo{author}{Chen, T.}, \bibinfo{author}{Guestrin, C.},
  \bibinfo{year}{2016}.
\newblock \bibinfo{title}{{XGBoost: A scalable tree boosting system}}, in:
  \bibinfo{booktitle}{{Proceedings of the 22nd ACM SIGKDD International
  Conference on Knowledge Discovery and Data Mining}}.
  \bibinfo{publisher}{ACM}, pp. \bibinfo{pages}{785--794}.
\newblock \bibinfo{note}{\url{https://doi.org/gdp84q}}.
\bibitem[{Cohen(1995)}]{cohen1995fast}
\bibinfo{author}{Cohen, W.W.}, \bibinfo{year}{1995}.
\newblock \bibinfo{title}{Fast effective rule induction}, in:
  \bibinfo{editor}{Prieditis, A.}, \bibinfo{editor}{Russell, S.} (Eds.),
  \bibinfo{booktitle}{{Machine Learning Proceedings 1995}}.
  \bibinfo{publisher}{Elsevier}, pp. \bibinfo{pages}{115--123}.
\newblock \bibinfo{note}{\url{https://doi.org/ghk75s}}.
\bibitem[{Dang and Chuc(2019)}]{dang2019-NetIncomeToStAssets}
\bibinfo{author}{Dang, L.N.}, \bibinfo{author}{Chuc, A.T.},
  \bibinfo{year}{2019}.
\newblock \bibinfo{title}{{Challenges in implementing the credit guarantee
  scheme for small and medium-sized enterprises: The case of Viet Nam}}.
\newblock \bibinfo{type}{Working Paper} \bibinfo{number}{941}. Asian
  Development Bank Institute.
\newblock \bibinfo{note}{\url{https://doi.org/g8p5}}.
\bibitem[{Datta et~al.(2016)Datta, Sen and Zick}]{datta2016algorithmic}
\bibinfo{author}{Datta, A.}, \bibinfo{author}{Sen, S.}, \bibinfo{author}{Zick,
  Y.}, \bibinfo{year}{2016}.
\newblock \bibinfo{title}{Algorithmic transparency via quantitative input
  influence: Theory and experiments with learning systems}, in:
  \bibinfo{booktitle}{{2016 IEEE Symposium on Security and Privacy}}.
  \bibinfo{publisher}{IEEE}, pp. \bibinfo{pages}{598--617}.
\newblock \bibinfo{note}{\url{https://doi.org/gft8t7}}.
\bibitem[{Dem{\v{s}}ar(2006)}]{demvsar2006statistical}
\bibinfo{author}{Dem{\v{s}}ar, J.}, \bibinfo{year}{2006}.
\newblock \bibinfo{title}{Statistical comparisons of classifiers over multiple
  data sets}.
\newblock \bibinfo{journal}{Journal of Machine Learning Research}
  \bibinfo{volume}{7}, \bibinfo{pages}{1--30}.
\newblock \URLprefix \url{https://bit.ly/3EfEQ8m}.
\bibitem[{Dem{\v{s}}ar and Bosni{\'c}(2018)}]{demvsar2018detecting}
\bibinfo{author}{Dem{\v{s}}ar, J.}, \bibinfo{author}{Bosni{\'c}, Z.},
  \bibinfo{year}{2018}.
\newblock \bibinfo{title}{Detecting concept drift in data streams using model
  explanation}.
\newblock \bibinfo{journal}{Expert Systems with Applications}
  \bibinfo{volume}{92}, \bibinfo{pages}{546--559}.
\newblock \bibinfo{note}{\url{https://doi.org/gf9djp}}.
\bibitem[{Dem\v{s}ar et~al.(2013)Dem\v{s}ar, Curk, Erjavec, \v{C}rt Gorup,
  Ho\v{c}evar, Milutinovi\v{c}, Mo\v{z}ina, Polajnar, Toplak, Stari\v{c},
  \v{S}tajdohar, Umek, \v{Z}agar, \v{Z}bontar, \v{Z}itnik and
  Zupan}]{JMLR:demsar13a}
\bibinfo{author}{Dem\v{s}ar, J.}, \bibinfo{author}{Curk, T.},
  \bibinfo{author}{Erjavec, A.}, \bibinfo{author}{\v{C}rt Gorup},
  \bibinfo{author}{Ho\v{c}evar, T.}, \bibinfo{author}{Milutinovi\v{c}, M.},
  \bibinfo{author}{Mo\v{z}ina, M.}, \bibinfo{author}{Polajnar, M.},
  \bibinfo{author}{Toplak, M.}, \bibinfo{author}{Stari\v{c}, A.},
  \bibinfo{author}{\v{S}tajdohar, M.}, \bibinfo{author}{Umek, L.},
  \bibinfo{author}{\v{Z}agar, L.}, \bibinfo{author}{\v{Z}bontar, J.},
  \bibinfo{author}{\v{Z}itnik, M.}, \bibinfo{author}{Zupan, B.},
  \bibinfo{year}{2013}.
\newblock \bibinfo{title}{Orange: Data mining toolbox in python}.
\newblock \bibinfo{journal}{Journal of Machine Learning Research}
  \bibinfo{volume}{14}, \bibinfo{pages}{2349--2353}.
\newblock \URLprefix \url{https://bit.ly/3Fye4tn}.
\bibitem[{Dong and Liu(2018)}]{dong2018feature}
\bibinfo{author}{Dong, G.}, \bibinfo{author}{Liu, H.}, \bibinfo{year}{2018}.
\newblock \bibinfo{title}{Feature engineering for machine learning and data
  analytics}.
\newblock \bibinfo{publisher}{CRC Press}.
\newblock \bibinfo{note}{\url{https://doi.org/hbg5}}.
\bibitem[{Dor and Reich(2012)}]{dor2012strengthening}
\bibinfo{author}{Dor, O.}, \bibinfo{author}{Reich, Y.}, \bibinfo{year}{2012}.
\newblock \bibinfo{title}{Strengthening learning algorithms by feature
  discovery}.
\newblock \bibinfo{journal}{Information Sciences} \bibinfo{volume}{189},
  \bibinfo{pages}{176--190}.
\newblock \bibinfo{note}{\url{https://doi.org/d882mw}}.
\bibitem[{Duan et~al.(2018)Duan, Zeng, Oprea and Vasudevan}]{duan2018automated}
\bibinfo{author}{Duan, J.}, \bibinfo{author}{Zeng, Z.}, \bibinfo{author}{Oprea,
  A.}, \bibinfo{author}{Vasudevan, S.}, \bibinfo{year}{2018}.
\newblock \bibinfo{title}{Automated generation and selection of interpretable
  features for enterprise security}, in: \bibinfo{booktitle}{{2018 IEEE
  International Conference on Big Data (Big Data)}}. \bibinfo{publisher}{IEEE},
  pp. \bibinfo{pages}{1258--1265}.
\newblock \bibinfo{note}{\url{https://doi.org/g8rj}}.
\bibitem[{Eibe et~al.(2016)Eibe, Hall and Witten}]{weka}
\bibinfo{author}{Eibe, F.}, \bibinfo{author}{Hall, M.A.},
  \bibinfo{author}{Witten, I.H.}, \bibinfo{year}{2016}.
\newblock \bibinfo{title}{{The WEKA workbench. Online appendix for data mining:
  Practical machine learning tools and techniques}}.
\newblock \bibinfo{edition}{4} ed., \bibinfo{publisher}{Morgan Kaufmann}.
\bibitem[{Fan et~al.(2010)Fan, Zhong, Peng, Verscheure, Zhang, Ren, Yan and
  Yang}]{fan2010generalized}
\bibinfo{author}{Fan, W.}, \bibinfo{author}{Zhong, E.}, \bibinfo{author}{Peng,
  J.}, \bibinfo{author}{Verscheure, O.}, \bibinfo{author}{Zhang, K.},
  \bibinfo{author}{Ren, J.}, \bibinfo{author}{Yan, R.}, \bibinfo{author}{Yang,
  Q.}, \bibinfo{year}{2010}.
\newblock \bibinfo{title}{Generalized and heuristic-free feature construction
  for improved accuracy}, in: \bibinfo{editor}{Parthasarathy, S.},
  \bibinfo{editor}{Liu, B.}, \bibinfo{editor}{Goethals, B.},
  \bibinfo{editor}{Pei, J.}, \bibinfo{editor}{Kamath, C.} (Eds.),
  \bibinfo{booktitle}{{Proceedings of the 2010 SIAM International Conference on
  Data Mining}}. \bibinfo{publisher}{SIAM}, pp. \bibinfo{pages}{629--640}.
\newblock \bibinfo{note}{\url{https://doi.org/g8rk}}.
\bibitem[{Fang et~al.(2020)Fang, Xia, Lin, Xia, Liu and
  Jiang}]{fang2019automatic}
\bibinfo{author}{Fang, J.}, \bibinfo{author}{Xia, S.}, \bibinfo{author}{Lin,
  J.}, \bibinfo{author}{Xia, Z.}, \bibinfo{author}{Liu, X.},
  \bibinfo{author}{Jiang, Y.}, \bibinfo{year}{2020}.
\newblock \bibinfo{title}{{Alpha Discovery Neural Network, the Special Fountain
  of Financial Trading Signals}}.
\newblock \bibinfo{journal}{arXiv preprint}
  \href{http://arxiv.org/abs/1912.11761v5}{{\tt arXiv:1912.11761v5}}.
\bibitem[{Feurer et~al.(2015)Feurer, Klein, Eggensperger, Springenberg, Blum
  and Hutter}]{feurer2015auto-sklearn}
\bibinfo{author}{Feurer, M.}, \bibinfo{author}{Klein, A.},
  \bibinfo{author}{Eggensperger, K.}, \bibinfo{author}{Springenberg, J.},
  \bibinfo{author}{Blum, M.}, \bibinfo{author}{Hutter, F.},
  \bibinfo{year}{2015}.
\newblock \bibinfo{title}{Efficient and robust automated machine learning}, in:
  \bibinfo{editor}{Cortes, C.}, \bibinfo{editor}{Lawrence, N.},
  \bibinfo{editor}{Lee, D.}, \bibinfo{editor}{Sugiyama, M.},
  \bibinfo{editor}{Garnett, R.} (Eds.), \bibinfo{booktitle}{Advances in Neural
  Information Processing Systems 28 {(NIPS 2015)}}, \bibinfo{publisher}{Curran
  Associates, Inc.}. pp. \bibinfo{pages}{2962--2970}.
\newblock \URLprefix \url{https://bit.ly/3rPhK5B}.
\bibitem[{Gama(1999)}]{gama1999discriminant}
\bibinfo{author}{Gama, J.}, \bibinfo{year}{1999}.
\newblock \bibinfo{title}{Discriminant trees}, in: \bibinfo{editor}{Ivan, B.},
  \bibinfo{editor}{Sa\v{s}o, D.} (Eds.), \bibinfo{booktitle}{{Proceedings of
  the 16th International Conference on Machine Learning (ICML 1999)}}.
  \bibinfo{publisher}{Elsevier}. volume~\bibinfo{volume}{1}, pp.
  \bibinfo{pages}{134--142}.
\bibitem[{Ganguin and Bilardello(2004)}]{ganguin2004standard}
\bibinfo{author}{Ganguin, B.}, \bibinfo{author}{Bilardello, J.},
  \bibinfo{year}{2004}.
\newblock \bibinfo{title}{Standard \& poor's fundamentals of corporate credit
  analysis}.
\newblock \bibinfo{publisher}{McGraw Hill Professional}.
\bibitem[{Gebauer et~al.(2018)Gebauer, Setzer and
  Westphal}]{gebauer2018-EbitdaToDebt}
\bibinfo{author}{Gebauer, S.}, \bibinfo{author}{Setzer, R.},
  \bibinfo{author}{Westphal, A.}, \bibinfo{year}{2018}.
\newblock \bibinfo{title}{Corporate debt and investment: A firm-level analysis
  for stressed euro area countries}.
\newblock \bibinfo{journal}{Journal of International Money and Finance}
  \bibinfo{volume}{86}, \bibinfo{pages}{112--130}.
\newblock \bibinfo{note}{\url{https://doi.org/gf23rs}}.
\bibitem[{Guid et~al.(2012)Guid, Mo{\v{z}}ina, Groznik, Georgiev, Sadikov,
  Pirto{\v{s}}ek and Bratko}]{guid2012abml}
\bibinfo{author}{Guid, M.}, \bibinfo{author}{Mo{\v{z}}ina, M.},
  \bibinfo{author}{Groznik, V.}, \bibinfo{author}{Georgiev, D.},
  \bibinfo{author}{Sadikov, A.}, \bibinfo{author}{Pirto{\v{s}}ek, Z.},
  \bibinfo{author}{Bratko, I.}, \bibinfo{year}{2012}.
\newblock \bibinfo{title}{{ABML} knowledge refinement loop: A case study}, in:
  \bibinfo{editor}{Chen, L.}, \bibinfo{editor}{Felfernig, A.},
  \bibinfo{editor}{Liu, J.}, \bibinfo{editor}{W.~Raś, Z.} (Eds.),
  \bibinfo{booktitle}{{International Symposium on Methodologies for Intelligent
  Systems}}, \bibinfo{publisher}{Springer}. pp. \bibinfo{pages}{41--50}.
\newblock \bibinfo{note}{\url{https://doi.org/g8rm}}.
\bibitem[{Guid et~al.(2019)Guid, Mo{\v{z}}ina, Pavli{\v{c}} and
  Tur{\v{s}}i{\v{c}}}]{guid2019learning}
\bibinfo{author}{Guid, M.}, \bibinfo{author}{Mo{\v{z}}ina, M.},
  \bibinfo{author}{Pavli{\v{c}}, M.}, \bibinfo{author}{Tur{\v{s}}i{\v{c}}, K.},
  \bibinfo{year}{2019}.
\newblock \bibinfo{title}{Learning by arguing in argument-based machine
  learning framework}, in: \bibinfo{booktitle}{International Conference on
  Intelligent Tutoring Systems}, pp. \bibinfo{pages}{112--122}.
\bibitem[{Hall et~al.(2021)Hall, Gill, Kurka and Phan}]{H2ODriverlessAI}
\bibinfo{author}{Hall, P.}, \bibinfo{author}{Gill, N.}, \bibinfo{author}{Kurka,
  M.}, \bibinfo{author}{Phan, W.}, \bibinfo{year}{2021}.
\newblock \bibinfo{title}{{H2O driverless AI (Version 1.10.1.2) [Computer
  software]}}.
\newblock \bibinfo{note}{H2O.ai. URL: \url{https://docs.h2o.ai/}}.
\bibitem[{Hammami et~al.(2020)Hammami, Bechikh, Louati, Makhlouf and
  Said}]{hammami2020feature}
\bibinfo{author}{Hammami, M.}, \bibinfo{author}{Bechikh, S.},
  \bibinfo{author}{Louati, A.}, \bibinfo{author}{Makhlouf, M.},
  \bibinfo{author}{Said, L.B.}, \bibinfo{year}{2020}.
\newblock \bibinfo{title}{Feature construction as a bi-level optimization
  problem}.
\newblock \bibinfo{journal}{Neural Computing and Applications}
  \bibinfo{volume}{32}, \bibinfo{pages}{13783--13804}.
\newblock \bibinfo{note}{\url{https://doi.org/g8rp}}.
\bibitem[{Hassine et~al.(2019)Hassine, Erbad and Hamila}]{hassine2019important}
\bibinfo{author}{Hassine, K.}, \bibinfo{author}{Erbad, A.},
  \bibinfo{author}{Hamila, R.}, \bibinfo{year}{2019}.
\newblock \bibinfo{title}{Important complexity reduction of random forest in
  multi-classification problem}, in: \bibinfo{booktitle}{{2019 15th
  International Wireless Communications \& Mobile Computing Conference
  (IWCMC)}}. \bibinfo{publisher}{IEEE}, pp. \bibinfo{pages}{226--231}.
\newblock \bibinfo{note}{\url{https://doi.org/g8rq}}.
\bibitem[{Hayes(2021)}]{hayes2021-EbitdaToInterest}
\bibinfo{author}{Hayes, A.}, \bibinfo{year}{2021}.
\newblock \bibinfo{title}{{EBITDA-To-Interest Coverage Ratio}. [{Online}]}.
\newblock \bibinfo{howpublished}{Available at: \url{https://bit.ly/3y8RrIL}}.
\newblock \bibinfo{note}{[Accessed: 24 November 2021]}.
\bibitem[{He et~al.(2021)He, Zhao and Chu}]{he2019automl}
\bibinfo{author}{He, X.}, \bibinfo{author}{Zhao, K.}, \bibinfo{author}{Chu,
  X.}, \bibinfo{year}{2021}.
\newblock \bibinfo{title}{{AutoML: A survey of the state-of-the-art}}.
\newblock \bibinfo{journal}{Knowledge-Based Systems} \bibinfo{volume}{212},
  \bibinfo{pages}{106622}.
\newblock \bibinfo{note}{\url{https://doi.org/gh69f6}}.
\bibitem[{Henelius et~al.(2017)Henelius, Puolam{\"a}ki and
  Ukkonen}]{henelius2017interpreting}
\bibinfo{author}{Henelius, A.}, \bibinfo{author}{Puolam{\"a}ki, K.},
  \bibinfo{author}{Ukkonen, A.}, \bibinfo{year}{2017}.
\newblock \bibinfo{title}{Interpreting classifiers through attribute
  interactions in datasets}, in: \bibinfo{editor}{Kim, B.},
  \bibinfo{editor}{M.~Malioutov, D.}, \bibinfo{editor}{R.~Varshney, K.},
  \bibinfo{editor}{Weller, A.} (Eds.), \bibinfo{booktitle}{{Proceedings of the
  2017 ICML. Workshop on Human Interpretability in Machine Learning (WHI
  2017)}}, pp. \bibinfo{pages}{8--13}.
\bibitem[{Holt(2001)}]{holt2001financial}
\bibinfo{author}{Holt, R.N.}, \bibinfo{year}{2001}.
\newblock \bibinfo{title}{Financial accounting: A management perspective}.
\newblock \bibinfo{publisher}{Ivy Learning Systems}.
\bibitem[{H{\"u}hn and H{\"u}llermeier(2009)}]{Huhn-2009-Furia}
\bibinfo{author}{H{\"u}hn, J.}, \bibinfo{author}{H{\"u}llermeier, E.},
  \bibinfo{year}{2009}.
\newblock \bibinfo{title}{{FURIA}: An algorithm for unordered fuzzy rule
  induction}.
\newblock \bibinfo{journal}{Data Mining and Knowledge Discovery}
  \bibinfo{volume}{19}, \bibinfo{pages}{293--319}.
\newblock \bibinfo{note}{\url{https://doi.org/cb4gb7}}.
\bibitem[{Ibrahim et~al.(2019)Ibrahim, Louie, Modarres and
  Paisley}]{ibrahim2019global}
\bibinfo{author}{Ibrahim, M.}, \bibinfo{author}{Louie, M.},
  \bibinfo{author}{Modarres, C.}, \bibinfo{author}{Paisley, J.},
  \bibinfo{year}{2019}.
\newblock \bibinfo{title}{Global explanations of neural networks: Mapping the
  landscape of predictions}, in: \bibinfo{booktitle}{{Proceedings of the 2019
  AAAI/ACM Conference on AI, Ethics, and Society}}. \bibinfo{publisher}{ACM}.
  AIES '19, pp. \bibinfo{pages}{279--287}.
\newblock \bibinfo{note}{\url{https://doi.org/ghj6ft}}.
\bibitem[{Irfan et~al.(2014)Irfan, Majeed and
  Zaman}]{irfan2014-NetIncomeToDebt}
\bibinfo{author}{Irfan, M.}, \bibinfo{author}{Majeed, Y.},
  \bibinfo{author}{Zaman, K.}, \bibinfo{year}{2014}.
\newblock \bibinfo{title}{The performance and efficiency of islamic banking in
  {South Asian} countries}.
\newblock \bibinfo{journal}{Economia. Seria Management} \bibinfo{volume}{17},
  \bibinfo{pages}{223--237}.
\bibitem[{Jakulin(2005)}]{jakulin2005MLattrInteractionsPhD}
\bibinfo{author}{Jakulin, A.}, \bibinfo{year}{2005}.
\newblock \bibinfo{title}{{Machine learning based on attribute interactions}}.
\newblock Ph.D. thesis. University of Ljubljana.
\newblock \URLprefix \url{https://bit.ly/3eiJ18x}.
\bibitem[{Jakulin and Bratko(2003a)}]{jakulin2003analyzing}
\bibinfo{author}{Jakulin, A.}, \bibinfo{author}{Bratko, I.},
  \bibinfo{year}{2003}a.
\newblock \bibinfo{title}{Analyzing attribute dependencies}, in:
  \bibinfo{editor}{Lavra\v{c}, N.}, \bibinfo{editor}{Gamberger, D.},
  \bibinfo{editor}{Todorovski, L.}, \bibinfo{editor}{Hendrik, B.} (Eds.),
  \bibinfo{booktitle}{{European Conference on Principles of Data Mining and
  Knowledge Discovery}}. \bibinfo{publisher}{Springer}, pp.
  \bibinfo{pages}{229--240}.
\newblock \bibinfo{note}{\url{https://doi.org/cnswcq}}.
\bibitem[{Jakulin and Bratko(2003b)}]{jakulin2003quantifying}
\bibinfo{author}{Jakulin, A.}, \bibinfo{author}{Bratko, I.},
  \bibinfo{year}{2003}b.
\newblock \bibinfo{title}{{Quantifying and visualizing attribute
  interactions}}.
\newblock \bibinfo{journal}{arXiv preprint}
  \href{http://arxiv.org/abs/cs/0308002v3}{{\tt arXiv:cs/0308002v3}}.
\bibitem[{Jiangli et~al.(2004)Jiangli, Unal and
  Yom}]{jiangli2004-NetIncomeToInterestExpenses}
\bibinfo{author}{Jiangli, W.}, \bibinfo{author}{Unal, H.},
  \bibinfo{author}{Yom, C.}, \bibinfo{year}{2004}.
\newblock \bibinfo{title}{{Relationship lending, accounting disclosure, and
  credit availability during the Asian financial crisis}}.
\newblock \bibinfo{journal}{Journal of Money, Credit, and Banking}
  \bibinfo{volume}{40}, \bibinfo{pages}{25--55}.
\newblock \bibinfo{note}{\url{https://doi.org/d29w37}}.
\bibitem[{Kanter and Veeramachaneni(2015)}]{kanter2015deep}
\bibinfo{author}{Kanter, J.M.}, \bibinfo{author}{Veeramachaneni, K.},
  \bibinfo{year}{2015}.
\newblock \bibinfo{title}{Deep feature synthesis: Towards automating data
  science endeavors}, in: \bibinfo{booktitle}{{2015 IEEE International
  Conference on Data Science and Advanced Analytics (DSAA)}}.
  \bibinfo{publisher}{IEEE}, pp. \bibinfo{pages}{1--10}.
\newblock \bibinfo{note}{\url{https://doi.org/gf2kxh}}.
\bibitem[{Katz et~al.(2016)Katz, Shin and Song}]{katz2016explorekit}
\bibinfo{author}{Katz, G.}, \bibinfo{author}{Shin, E.C.R.},
  \bibinfo{author}{Song, D.}, \bibinfo{year}{2016}.
\newblock \bibinfo{title}{{ExploreKit}: Automatic feature generation and
  selection}, in: \bibinfo{booktitle}{{2016 IEEE 16th International Conference
  on Data Mining (ICDM}}. \bibinfo{publisher}{IEEE}, pp.
  \bibinfo{pages}{979--984}.
\newblock \bibinfo{note}{\url{https://doi.org/g8zc}}.
\bibitem[{Kenton(2020)}]{kenton2020-FfoToDebt}
\bibinfo{author}{Kenton, W.}, \bibinfo{year}{2020}.
\newblock \bibinfo{title}{{Funds from operations (FFO) to total debt ratio}.
  [{Online}]}.
\newblock \bibinfo{howpublished}{Available at: \url{https://bit.ly/3rKdnb4}}.
\newblock \bibinfo{note}{[Accessed: 5 November 2021]}.
\bibitem[{Koller et~al.(2010)Koller, Goedhart, Wessels
  et~al.}]{koller2010-EbitToNetDebt}
\bibinfo{author}{Koller, T.}, \bibinfo{author}{Goedhart, M.},
  \bibinfo{author}{Wessels, D.}, et~al., \bibinfo{year}{2010}.
\newblock \bibinfo{title}{Valuation: Measuring and managing the value of
  companies}.
\newblock \bibinfo{edition}{5} ed., \bibinfo{publisher}{John Wiley \& Sons}.
\bibitem[{Kononenko(1991)}]{kononenko1991semiNB}
\bibinfo{author}{Kononenko, I.}, \bibinfo{year}{1991}.
\newblock \bibinfo{title}{Semi-naive {Bayesian} classifier}, in:
  \bibinfo{editor}{Kodratoff, Y.} (Ed.), \bibinfo{booktitle}{{Machine Learning
  -- EWSL-91. Lecture Notes in Computer Science (Lecture Notes in Artificial
  Intelligence)}}, \bibinfo{publisher}{Springer}. pp.
  \bibinfo{pages}{206--219}.
\newblock \bibinfo{note}{\url{https://doi.org/bt4q4w}}.
\bibitem[{Kononenko(1995)}]{kononenko-1995-biases}
\bibinfo{author}{Kononenko, I.}, \bibinfo{year}{1995}.
\newblock \bibinfo{title}{On biases in estimating multi-valued attributes}, in:
  \bibinfo{booktitle}{{Proceedings of the Fourteenth International Joint
  Conference on Artificial Intelligence (I)}}, \bibinfo{publisher}{Morgan
  Kaufmann}. pp. \bibinfo{pages}{1034--1040}.
\newblock \URLprefix \url{https://bit.ly/3HkEGhT}.
\bibitem[{Lachiche(2010)}]{Lachiche2010}
\bibinfo{author}{Lachiche, N.}, \bibinfo{year}{2010}.
\newblock \bibinfo{title}{Propositionalization}, in: \bibinfo{editor}{Sammut,
  C.}, \bibinfo{editor}{Webb, G.I.} (Eds.), \bibinfo{booktitle}{Encyclopedia of
  Machine Learning}. \bibinfo{publisher}{Springer US},
  \bibinfo{address}{Boston, MA}, pp. \bibinfo{pages}{812--817}.
\newblock \bibinfo{note}{\url{https://doi.org/cbkgvp}}.
\bibitem[{Lam et~al.(2019)Lam, Minh, Sinn, Buesser and Wistuba}]{lam2018RNN}
\bibinfo{author}{Lam, H.T.}, \bibinfo{author}{Minh, T.N.},
  \bibinfo{author}{Sinn, M.}, \bibinfo{author}{Buesser, B.},
  \bibinfo{author}{Wistuba, M.}, \bibinfo{year}{2019}.
\newblock \bibinfo{title}{{Neural feature learning from relational databases}}.
\newblock \bibinfo{journal}{arXiv preprint}
  \href{http://arxiv.org/abs/1801.05372v4}{{\tt arXiv:1801.05372v4}}.
\bibitem[{Lam et~al.(2017)Lam, Thiebaut, Sinn, Chen, Mai and
  Alkan}]{lam2017one}
\bibinfo{author}{Lam, H.T.}, \bibinfo{author}{Thiebaut, J.},
  \bibinfo{author}{Sinn, M.}, \bibinfo{author}{Chen, B.}, \bibinfo{author}{Mai,
  T.}, \bibinfo{author}{Alkan, O.}, \bibinfo{year}{2017}.
\newblock \bibinfo{title}{{One button machine for automating feature
  engineering in relational databases}}.
\newblock \bibinfo{journal}{arXiv preprint}
  \href{http://arxiv.org/abs/1706.00327}{{\tt arXiv:1706.00327}}.
\bibitem[{Lee and Lee(2016)}]{lee2016-EbitToInterest}
\bibinfo{author}{Lee, J.C.}, \bibinfo{author}{Lee, C.F.}, \bibinfo{year}{2016}.
\newblock \bibinfo{title}{{Accounting Information and Regression Analysis}},
  in: \bibinfo{booktitle}{{Financial Analysis, Planning \& Forecasting: Theory
  and Application}}. \bibinfo{edition}{3} ed.. \bibinfo{publisher}{World
  Scientific}, pp. \bibinfo{pages}{13--73}.
\newblock \bibinfo{note}{\url{https://doi.org/g8zf}}.
\bibitem[{Lemaire et~al.(2008)Lemaire, F{\'e}raud and
  Voisine}]{lemaire2008contact}
\bibinfo{author}{Lemaire, V.}, \bibinfo{author}{F{\'e}raud, R.},
  \bibinfo{author}{Voisine, N.}, \bibinfo{year}{2008}.
\newblock \bibinfo{title}{Contact personalization using a score understanding
  method}, in: \bibinfo{booktitle}{{2008 IEEE International Joint Conference on
  Neural Networks}}. \bibinfo{publisher}{IEEE}, pp. \bibinfo{pages}{649--654}.
\newblock \bibinfo{note}{\url{https://doi.org/c6ngtm}}.
\bibitem[{Lipovetsky and Conklin(2001)}]{lipovetsky2001analysis}
\bibinfo{author}{Lipovetsky, S.}, \bibinfo{author}{Conklin, M.},
  \bibinfo{year}{2001}.
\newblock \bibinfo{title}{Analysis of regression in game theory approach}.
\newblock \bibinfo{journal}{Applied Stochastic Models in Business and Industry}
  \bibinfo{volume}{17}, \bibinfo{pages}{319--330}.
\newblock \bibinfo{note}{\url{https://doi.org/cdc72h}}.
\bibitem[{Lundberg et~al.(2020)Lundberg, Erion, Chen, DeGrave, Prutkin, Nair,
  Katz, Himmelfarb, Bansal and Lee}]{lundberg2020local}
\bibinfo{author}{Lundberg, S.M.}, \bibinfo{author}{Erion, G.},
  \bibinfo{author}{Chen, H.}, \bibinfo{author}{DeGrave, A.},
  \bibinfo{author}{Prutkin, J.M.}, \bibinfo{author}{Nair, B.},
  \bibinfo{author}{Katz, R.}, \bibinfo{author}{Himmelfarb, J.},
  \bibinfo{author}{Bansal, N.}, \bibinfo{author}{Lee, S.I.},
  \bibinfo{year}{2020}.
\newblock \bibinfo{title}{From local explanations to global understanding with
  explainable {AI} for trees}.
\newblock \bibinfo{journal}{Nature Machine Intelligence} \bibinfo{volume}{2},
  \bibinfo{pages}{56--67}.
\newblock \bibinfo{note}{\url{https://doi.org/ggjtp4}}.
\bibitem[{Lundberg and Lee(2017a)}]{lundberg2017consistent}
\bibinfo{author}{Lundberg, S.M.}, \bibinfo{author}{Lee, S.},
  \bibinfo{year}{2017}a.
\newblock \bibinfo{title}{Consistent feature attribution for tree ensembles},
  in: \bibinfo{editor}{Kim, B.}, \bibinfo{editor}{M.~Malioutov, D.},
  \bibinfo{editor}{R.~Varshney, K.}, \bibinfo{editor}{Weller, A.} (Eds.),
  \bibinfo{booktitle}{{Proceedings of the 2017 ICML. Workshop on Human
  Interpretability in Machine Learning (WHI 2017)}}, pp.
  \bibinfo{pages}{15--21}.
\newblock \bibinfo{note}{\url{https://doi.org/jghb}}.
\bibitem[{Lundberg and Lee(2017b)}]{Lundberg-2017-SHAP-modelInterpreting}
\bibinfo{author}{Lundberg, S.M.}, \bibinfo{author}{Lee, S.I.},
  \bibinfo{year}{2017}b.
\newblock \bibinfo{title}{A unified approach to interpreting model
  predictions}, in: \bibinfo{editor}{Guyon, I.}, \bibinfo{editor}{Luxburg,
  U.V.}, \bibinfo{editor}{Bengio, S.}, \bibinfo{editor}{Wallach, H.},
  \bibinfo{editor}{Fergus, R.}, \bibinfo{editor}{Vishwanathan, S.},
  \bibinfo{editor}{Garnett, R.} (Eds.), \bibinfo{booktitle}{Advances in Neural
  Information Processing Systems 30 {(NIPS 2017)}}, \bibinfo{publisher}{Curran
  Associates, Inc.}. pp. \bibinfo{pages}{4765--4774}.
\newblock \URLprefix \url{https://bit.ly/3zhk5Is}.
\bibitem[{Lundberg et~al.(2018)Lundberg, Nair, Vavilala, Horibe, Eisses, Adams,
  Liston, Low, Newman, Kim and Lee}]{Lundberg-2018-SHAP-hypoxaemia}
\bibinfo{author}{Lundberg, S.M.}, \bibinfo{author}{Nair, B.},
  \bibinfo{author}{Vavilala, M.S.}, \bibinfo{author}{Horibe, M.},
  \bibinfo{author}{Eisses, M.J.}, \bibinfo{author}{Adams, T.},
  \bibinfo{author}{Liston, D.E.}, \bibinfo{author}{Low, D.K.W.},
  \bibinfo{author}{Newman, S.F.}, \bibinfo{author}{Kim, J.},
  \bibinfo{author}{Lee, S.I.}, \bibinfo{year}{2018}.
\newblock \bibinfo{title}{Explainable machine-learning predictions for the
  prevention of hypoxaemia during surgery}.
\newblock \bibinfo{journal}{Nature Biomedical Engineering} \bibinfo{volume}{2},
  \bibinfo{pages}{749--760}.
\newblock \bibinfo{note}{\url{https://doi.org/ggbrzv}}.
\bibitem[{Markovitch and Rosenstein(2002)}]{markovitch2002feature}
\bibinfo{author}{Markovitch, S.}, \bibinfo{author}{Rosenstein, D.},
  \bibinfo{year}{2002}.
\newblock \bibinfo{title}{Feature generation using general constructor
  functions}.
\newblock \bibinfo{journal}{Machine Learning} \bibinfo{volume}{49},
  \bibinfo{pages}{59--98}.
\newblock \bibinfo{note}{\url{https://doi.org/dqkwj5}}.
\bibitem[{Matheus and Rendell(1989)}]{matheus1989constructive}
\bibinfo{author}{Matheus, C.J.}, \bibinfo{author}{Rendell, L.A.},
  \bibinfo{year}{1989}.
\newblock \bibinfo{title}{Constructive induction on decision trees}, in:
  \bibinfo{booktitle}{{Proceedings of the Eleventh International Joint
  Conference on Artificial Intelligence (I)}}. \bibinfo{publisher}{Morgan
  Kaufmann}, pp. \bibinfo{pages}{645--650}.
\newblock \URLprefix \url{https://bit.ly/3EZT2mk}.
\bibitem[{Molnar(2021)}]{molnar2019-Interpretable-ML}
\bibinfo{author}{Molnar, C.}, \bibinfo{year}{2021}.
\newblock \bibinfo{title}{Interpretable machine learning: {A} guide for making
  black box models explainable}.
\newblock \bibinfo{publisher}{Leanpub}.
\newblock \URLprefix \url{https://bit.ly/3A2dUXV}.
\bibitem[{Možina(2009)}]{mozina-phd-thesis}
\bibinfo{author}{Možina, M.}, \bibinfo{year}{2009}.
\newblock \bibinfo{title}{Argument based machine learning}.
\newblock Ph.D. thesis. University of Ljubljana.
\newblock \URLprefix \url{https://bit.ly/3Ts1dPA}.
\bibitem[{Muharram and Smith(2005)}]{muharram2005evolutionary}
\bibinfo{author}{Muharram, M.}, \bibinfo{author}{Smith, G.D.},
  \bibinfo{year}{2005}.
\newblock \bibinfo{title}{Evolutionary constructive induction}.
\newblock \bibinfo{journal}{IEEE Transactions on Knowledge and Data
  Engineering} \bibinfo{volume}{17}, \bibinfo{pages}{1518--1528}.
\newblock \bibinfo{note}{\url{https://doi.org/cpvc65}}.
\bibitem[{Murthy et~al.(2018)Murthy, Chanda et~al.}]{murthy2018generation}
\bibinfo{author}{Murthy, C.}, \bibinfo{author}{Chanda, B.}, et~al.,
  \bibinfo{year}{2018}.
\newblock \bibinfo{title}{Generation of compound features based on feature
  interaction for classification}.
\newblock \bibinfo{journal}{Expert Systems with Applications}
  \bibinfo{volume}{108}, \bibinfo{pages}{61--73}.
\newblock \bibinfo{note}{\url{https://doi.org/g82q}}.
\bibitem[{Nargesian et~al.(2017)Nargesian, Samulowitz, Khurana, Khalil and
  Turaga}]{nargesian2017learning}
\bibinfo{author}{Nargesian, F.}, \bibinfo{author}{Samulowitz, H.},
  \bibinfo{author}{Khurana, U.}, \bibinfo{author}{Khalil, E.B.},
  \bibinfo{author}{Turaga, D.S.}, \bibinfo{year}{2017}.
\newblock \bibinfo{title}{Learning feature engineering for classification.},
  in: \bibinfo{editor}{Sierra, C.} (Ed.), \bibinfo{booktitle}{{Proceedings of
  the Twenty-Sixth International Joint Conference on Artificial Intelligence}}.
  \bibinfo{publisher}{IJCAI}, pp. \bibinfo{pages}{2529--2535}.
\newblock \bibinfo{note}{\url{https://doi.org/gj9xqm}}.
\bibitem[{Ozdemir and Susarla(2018)}]{ozdemir2018feature}
\bibinfo{author}{Ozdemir, S.}, \bibinfo{author}{Susarla, D.},
  \bibinfo{year}{2018}.
\newblock \bibinfo{title}{Feature engineering made easy: Identify unique
  features from your dataset in order to build powerful machine learning
  systems}.
\newblock \bibinfo{publisher}{Packt}.
\bibitem[{Pazzani(1996)}]{pazzani1996searching}
\bibinfo{author}{Pazzani, M.J.}, \bibinfo{year}{1996}.
\newblock \bibinfo{title}{Searching for dependencies in {Bayesian}
  classifiers}, in: \bibinfo{editor}{Fisher, D.}, \bibinfo{editor}{Lenz, H.J.}
  (Eds.), \bibinfo{booktitle}{Learning from data: {A}rtificial intelligence and
  statistics {V}}. \bibinfo{publisher}{Springer}. volume \bibinfo{volume}{112},
  pp. \bibinfo{pages}{239--248}.
\newblock \bibinfo{note}{\url{https://doi.org/drgmkr}}.
\bibitem[{Pechenizkiy(2005)}]{pechenizkiy2005impact}
\bibinfo{author}{Pechenizkiy, M.}, \bibinfo{year}{2005}.
\newblock \bibinfo{title}{The impact of feature extraction on the performance
  of a classifier: {kNN}, {Na{\"i}ve Bayes} and {C4.5}}, in:
  \bibinfo{editor}{K{\'e}gl, B.}, \bibinfo{editor}{Lapalm, G.} (Eds.),
  \bibinfo{booktitle}{{Conference of the Canadian Society for Computational
  Studies of Intelligence}}. \bibinfo{publisher}{Springer}, pp.
  \bibinfo{pages}{268--279}.
\newblock \bibinfo{note}{\url{https://doi.org/bzj499}}.
\bibitem[{Perez and Rendell(1995)}]{perez1995using}
\bibinfo{author}{Perez, E.}, \bibinfo{author}{Rendell, L.A.},
  \bibinfo{year}{1995}.
\newblock \bibinfo{title}{Using multidimensional projection to find relations},
  in: \bibinfo{editor}{Prieditis, A.}, \bibinfo{editor}{Russell, S.} (Eds.),
  \bibinfo{booktitle}{{Machine Learning Proceedings 1995}}.
  \bibinfo{publisher}{Elsevier}, pp. \bibinfo{pages}{447--455}.
\newblock \bibinfo{note}{\url{https://doi.org/g82s}}.
\bibitem[{Quinlan(1986)}]{quinlan1986induction}
\bibinfo{author}{Quinlan, J.R.}, \bibinfo{year}{1986}.
\newblock \bibinfo{title}{Induction of decision trees}.
\newblock \bibinfo{journal}{Machine learning} \bibinfo{volume}{1},
  \bibinfo{pages}{81--106}.
\newblock \bibinfo{note}{\url{https://doi.org/ctd6mv}}.
\bibitem[{Ragavan et~al.(1993)Ragavan, Rendell, Shaw and
  Tessmer}]{ragavan1993complex}
\bibinfo{author}{Ragavan, H.}, \bibinfo{author}{Rendell, L.},
  \bibinfo{author}{Shaw, M.}, \bibinfo{author}{Tessmer, A.},
  \bibinfo{year}{1993}.
\newblock \bibinfo{title}{Complex concept acquisition through directed search
  and feature caching}, in: \bibinfo{booktitle}{{Proceedings of the Thirteenth
  International Joint Conference on Artificial Intelligence (II)}}.
  \bibinfo{publisher}{Morgan Kaufmann}, pp. \bibinfo{pages}{946--951}.
\newblock \URLprefix \url{https://bit.ly/34lRwy8}.
\bibitem[{Ribeiro et~al.(2016)Ribeiro, Singh and Guestrin}]{Ribeiro-2016-LIME}
\bibinfo{author}{Ribeiro, M.T.}, \bibinfo{author}{Singh, S.},
  \bibinfo{author}{Guestrin, C.}, \bibinfo{year}{2016}.
\newblock \bibinfo{title}{"{Why should I trust you}?": Explaining the
  predictions of any classifier}, in: \bibinfo{booktitle}{{Proceedings of the
  22nd ACM SIGKDD International Conference on Knowledge Discovery and Data
  Mining}}. \bibinfo{publisher}{ACM}. KDD '16, pp. \bibinfo{pages}{1135--1144}.
\newblock \bibinfo{note}{\url{https://doi.org/gfgrbd}}.
\bibitem[{Robnik-{\v{S}}ikonja(2003)}]{robnik2003experiments}
\bibinfo{author}{Robnik-{\v{S}}ikonja, M.}, \bibinfo{year}{2003}.
\newblock \bibinfo{title}{Experiments with cost-sensitive feature evaluation},
  in: \bibinfo{editor}{Lavra\v{c}, N.}, \bibinfo{editor}{Gamberger, D.},
  \bibinfo{editor}{Blockeel, H.}, \bibinfo{editor}{Todorovski, L.} (Eds.),
  \bibinfo{booktitle}{{Machine Learning: ECML 2003}},
  \bibinfo{organization}{Springer}. pp. \bibinfo{pages}{325--336}.
\newblock \bibinfo{note}{\url{https://doi.org/fj768j}}.
\bibitem[{Robnik-{\v{S}}ikonja and Kononenko(2008)}]{robnik2008explain}
\bibinfo{author}{Robnik-{\v{S}}ikonja, M.}, \bibinfo{author}{Kononenko, I.},
  \bibinfo{year}{2008}.
\newblock \bibinfo{title}{Explaining classifications for individual instances}.
\newblock \bibinfo{journal}{IEEE Transactions on Knowledge and Data
  Engineering} \bibinfo{volume}{20}, \bibinfo{pages}{589--600}.
\newblock \bibinfo{note}{\url{https://doi.org/bkcczp}}.
\bibitem[{Rostami et~al.(2021)Rostami, Berahmand, Nasiri and
  Forouzandeh}]{rostami2021-review}
\bibinfo{author}{Rostami, M.}, \bibinfo{author}{Berahmand, K.},
  \bibinfo{author}{Nasiri, E.}, \bibinfo{author}{Forouzandeh, S.},
  \bibinfo{year}{2021}.
\newblock \bibinfo{title}{Review of swarm intelligence-based feature selection
  methods}.
\newblock \bibinfo{journal}{Engineering Applications of Artificial
  Intelligence} \bibinfo{volume}{100}, \bibinfo{pages}{104210}.
\newblock \bibinfo{note}{\url{https://doi.org/gppvr9}}.
\bibitem[{Rostami et~al.(2022)Rostami, Forouzandeh, Berahmand, Soltani,
  Shahsavari and Oussalah}]{rostami2022-gene}
\bibinfo{author}{Rostami, M.}, \bibinfo{author}{Forouzandeh, S.},
  \bibinfo{author}{Berahmand, K.}, \bibinfo{author}{Soltani, M.},
  \bibinfo{author}{Shahsavari, M.}, \bibinfo{author}{Oussalah, M.},
  \bibinfo{year}{2022}.
\newblock \bibinfo{title}{Gene selection for microarray data classification via
  multi-objective graph theoretic-based method}.
\newblock \bibinfo{journal}{Artificial Intelligence in Medicine}
  \bibinfo{volume}{123}, \bibinfo{pages}{102228}.
\newblock \bibinfo{note}{\url{https://doi.org/gpwpbg}}.
\bibitem[{Saabas(2014)}]{Saabas2014}
\bibinfo{author}{Saabas, A.}, \bibinfo{year}{2014}.
\newblock \bibinfo{title}{Diving into data. [{Online}]}.
\newblock \bibinfo{howpublished}{Available at: \url{https://bit.ly/3lryDjG}}.
\newblock \bibinfo{note}{[Accessed: 24 July 2020]}.
\bibitem[{Shrikumar et~al.(2017)Shrikumar, Greenside and
  Kundaje}]{DeepLIFT2017}
\bibinfo{author}{Shrikumar, A.}, \bibinfo{author}{Greenside, P.},
  \bibinfo{author}{Kundaje, A.}, \bibinfo{year}{2017}.
\newblock \bibinfo{title}{Learning important features through propagating
  activation differences}, in: \bibinfo{editor}{Precup, D.},
  \bibinfo{editor}{Whye~Teh, Y.} (Eds.), \bibinfo{booktitle}{{Proceedings of
  the 34th International Conference on Machine Learning}}, pp.
  \bibinfo{pages}{3145--3153}.
\newblock \URLprefix \url{https://bit.ly/3J1Mtme}.
\bibitem[{St.~Amand and Huan(2017)}]{st2017sparse}
\bibinfo{author}{St.~Amand, J.}, \bibinfo{author}{Huan, J.},
  \bibinfo{year}{2017}.
\newblock \bibinfo{title}{Sparse compositional local metric learning}, in:
  \bibinfo{booktitle}{{Proceedings of the 23rd ACM SIGKDD International
  Conference on Knowledge Discovery and Data Mining}}.
  \bibinfo{publisher}{ACM}, pp. \bibinfo{pages}{1097--1104}.
\newblock \bibinfo{note}{\url{https://doi.org/g82v}}.
\bibitem[{Tang et~al.(2019)Tang, Dai and Xiang}]{tang2019feature}
\bibinfo{author}{Tang, X.}, \bibinfo{author}{Dai, Y.}, \bibinfo{author}{Xiang,
  Y.}, \bibinfo{year}{2019}.
\newblock \bibinfo{title}{Feature selection based on feature interactions with
  application to text categorization}.
\newblock \bibinfo{journal}{Expert Systems with Applications}
  \bibinfo{volume}{120}, \bibinfo{pages}{207--216}.
\newblock \bibinfo{note}{\url{https://doi.org/gj49sg}}.
\bibitem[{Tjoa and Guan(2021)}]{tjoa2019survey}
\bibinfo{author}{Tjoa, E.}, \bibinfo{author}{Guan, C.}, \bibinfo{year}{2021}.
\newblock \bibinfo{title}{A survey on explainable artificial intelligence
  ({XAI}): Towards medical {XAI}}.
\newblock \bibinfo{journal}{IEEE Transactions on Neural Networks and Learning
  Systems} \bibinfo{volume}{32}, \bibinfo{pages}{4793--4813}.
\newblock \bibinfo{note}{\url{https://doi.org/ghknrh}}.
\bibitem[{\v{S}trumbelj et~al.(2010)\v{S}trumbelj, Bosni{\'c}, Kononenko,
  Zakotnik and Kuhar-Gra{\v{s}}i{\v{c}}}]{vstrumbelj2010explanation}
\bibinfo{author}{\v{S}trumbelj, E.}, \bibinfo{author}{Bosni{\'c}, Z.},
  \bibinfo{author}{Kononenko, I.}, \bibinfo{author}{Zakotnik, B.},
  \bibinfo{author}{Kuhar-Gra{\v{s}}i{\v{c}}, C.}, \bibinfo{year}{2010}.
\newblock \bibinfo{title}{Explanation and reliability of prediction models: The
  case of breast cancer recurrence}.
\newblock \bibinfo{journal}{Knowledge and Information Systems}
  \bibinfo{volume}{24}, \bibinfo{pages}{305--324}.
\newblock \bibinfo{note}{\url{https://doi.org/dqtvqq}}.
\bibitem[{\v{S}trumbelj and Kononenko(2010)}]{Strumbelj-2010}
\bibinfo{author}{\v{S}trumbelj, E.}, \bibinfo{author}{Kononenko, I.},
  \bibinfo{year}{2010}.
\newblock \bibinfo{title}{An efficient explanation of individual
  classifications using game theory}.
\newblock \bibinfo{journal}{Journal of Machine Learning Research}
  \bibinfo{volume}{11}, \bibinfo{pages}{1--18}.
\bibitem[{\v{S}trumbelj and Kononenko(2014)}]{Strumbelj-2014}
\bibinfo{author}{\v{S}trumbelj, E.}, \bibinfo{author}{Kononenko, I.},
  \bibinfo{year}{2014}.
\newblock \bibinfo{title}{Explaining prediction models and individual
  predictions with feature contributions}.
\newblock \bibinfo{journal}{Knowledge and Information Systems}
  \bibinfo{volume}{41}, \bibinfo{pages}{647--665}.
\newblock \bibinfo{note}{\url{https://doi.org/f6pnsr}}.
\bibitem[{\v{S}trumbelj et~al.(2009)\v{S}trumbelj, Kononenko and
  {Robnik-Šikonja}}]{Strumbelj-2009}
\bibinfo{author}{\v{S}trumbelj, E.}, \bibinfo{author}{Kononenko, I.},
  \bibinfo{author}{{Robnik-Šikonja}, M.}, \bibinfo{year}{2009}.
\newblock \bibinfo{title}{Explaining instance classifications with interactions
  of subsets of feature values}.
\newblock \bibinfo{journal}{Data \& Knowledge Engineering}
  \bibinfo{volume}{68}, \bibinfo{pages}{886--904}.
\newblock \bibinfo{note}{\url{https://doi.org/fnkvnd}}.
\bibitem[{Woolf(2010)}]{woolf2010building}
\bibinfo{author}{Woolf, B.P.}, \bibinfo{year}{2010}.
\newblock \bibinfo{title}{Building intelligent interactive tutors:
  Student-centered strategies for revolutionizing e-learning}.
\newblock \bibinfo{publisher}{Morgan Kaufmann}.
\bibitem[{Xing et~al.(2022a)Xing, Xiao, Qu, Zhu and Zhao}]{xing2022-EFDLS}
\bibinfo{author}{Xing, H.}, \bibinfo{author}{Xiao, Z.}, \bibinfo{author}{Qu,
  R.}, \bibinfo{author}{Zhu, Z.}, \bibinfo{author}{Zhao, B.},
  \bibinfo{year}{2022}a.
\newblock \bibinfo{title}{An efficient federated distillation learning system
  for multitask time series classification}.
\newblock \bibinfo{journal}{IEEE Transactions on Instrumentation and
  Measurement} \bibinfo{volume}{71}, \bibinfo{pages}{1--12}.
\newblock \bibinfo{note}{\url{https://doi.org/jb85}}.
\bibitem[{Xing et~al.(2022b)Xing, Xiao, Zhan, Luo, Dai and
  Li}]{xing2022-selfmatch}
\bibinfo{author}{Xing, H.}, \bibinfo{author}{Xiao, Z.}, \bibinfo{author}{Zhan,
  D.}, \bibinfo{author}{Luo, S.}, \bibinfo{author}{Dai, P.},
  \bibinfo{author}{Li, K.}, \bibinfo{year}{2022}b.
\newblock \bibinfo{title}{Selfmatch: Robust semisupervised time-series
  classification with self-distillation}.
\newblock \bibinfo{journal}{International Journal of Intelligent Systems}
  \bibinfo{volume}{37}, \bibinfo{pages}{8583--8610}.
\newblock \bibinfo{note}{\url{https://doi.org/jb87}}.
\bibitem[{Yazdani et~al.(2017)Yazdani, Shanbehzadeh and
  Hadavandi}]{yazdani2017mbcgp}
\bibinfo{author}{Yazdani, S.}, \bibinfo{author}{Shanbehzadeh, J.},
  \bibinfo{author}{Hadavandi, E.}, \bibinfo{year}{2017}.
\newblock \bibinfo{title}{{MBCGP-FE: A modified balanced cartesian genetic
  programming feature extractor}}.
\newblock \bibinfo{journal}{Knowledge-Based Systems} \bibinfo{volume}{135},
  \bibinfo{pages}{89--98}.
\newblock \bibinfo{note}{\url{https://doi.org/gcf9vp}}.
\bibitem[{Zeng et~al.(2015)Zeng, Zhang, Zhang and Zhang}]{zeng2015mixed}
\bibinfo{author}{Zeng, Z.}, \bibinfo{author}{Zhang, H.},
  \bibinfo{author}{Zhang, R.}, \bibinfo{author}{Zhang, Y.},
  \bibinfo{year}{2015}.
\newblock \bibinfo{title}{A mixed feature selection method considering
  interaction}.
\newblock \bibinfo{journal}{Mathematical Problems in Engineering}
  \bibinfo{volume}{2015}, \bibinfo{pages}{989067}.
\newblock \bibinfo{note}{\url{https://doi.org/f67swf}}.
\bibitem[{Zhao et~al.(2009)Zhao, Sinha and Ge}]{zhao2009effects}
\bibinfo{author}{Zhao, H.}, \bibinfo{author}{Sinha, A.P.}, \bibinfo{author}{Ge,
  W.}, \bibinfo{year}{2009}.
\newblock \bibinfo{title}{Effects of feature construction on classification
  performance: An empirical study in bank failure prediction}.
\newblock \bibinfo{journal}{Expert Systems with Applications}
  \bibinfo{volume}{36}, \bibinfo{pages}{2633--2644}.
\newblock \bibinfo{note}{\url{https://doi.org/fnk5mk}}.
\bibitem[{Zheng(2000)}]{zheng2000-XofN}
\bibinfo{author}{Zheng, Z.}, \bibinfo{year}{2000}.
\newblock \bibinfo{title}{Constructing {X-of-N} attributes for decision tree
  learning}.
\newblock \bibinfo{journal}{Machine Learning} \bibinfo{volume}{40},
  \bibinfo{pages}{35--75}.
\newblock \bibinfo{note}{\url{https://doi.org/b9vmp9}}.
\bibitem[{Zupan et~al.(1998)Zupan, Bohanec, Dem{\v{s}}ar and
  Bratko}]{zupan1998feature}
\bibinfo{author}{Zupan, B.}, \bibinfo{author}{Bohanec, M.},
  \bibinfo{author}{Dem{\v{s}}ar, J.}, \bibinfo{author}{Bratko, I.},
  \bibinfo{year}{1998}.
\newblock \bibinfo{title}{Feature transformation by function decomposition},
  in: \bibinfo{editor}{Liu, H.}, \bibinfo{editor}{Motoda, H.} (Eds.),
  \bibinfo{booktitle}{{Feature extraction, construction and selection: A data
  mining perspective}}, \bibinfo{publisher}{Springer}. pp.
  \bibinfo{pages}{325--340}.
\newblock \bibinfo{note}{\url{https://doi.org/dkd2r7}}.
\bibitem[{Zupan et~al.(2001)Zupan, Bratko, Bohanec and
  Dem{\v{s}}ar}]{zupan1999function}
\bibinfo{author}{Zupan, B.}, \bibinfo{author}{Bratko, I.},
  \bibinfo{author}{Bohanec, M.}, \bibinfo{author}{Dem{\v{s}}ar, J.},
  \bibinfo{year}{2001}.
\newblock \bibinfo{title}{Function decomposition in machine learning}, in:
  \bibinfo{editor}{Paliouras, G.}, \bibinfo{editor}{Karkaletsis, V.},
  \bibinfo{editor}{Spyropoulos, C.D.} (Eds.), \bibinfo{booktitle}{{Machine
  learning and its applications: Advanced lectures}}.
  \bibinfo{publisher}{Springer}. volume \bibinfo{volume}{2049}, pp.
  \bibinfo{pages}{71--101}.
\newblock \bibinfo{note}{\url{https://doi.org/b6mz5s}}.

\end{thebibliography}
\end{document}